\ifdefined\pdfminorversion\pdfminorversion=7\fi
\ifdefined\pdfobjcompresslevel\pdfobjcompresslevel=0\fi
\PassOptionsToPackage{table}{xcolor}
\documentclass[11pt,letterpaper]{applemlr}
\usepackage{amsmath,amssymb,xspace,enumitem,url,pifont}
\usepackage{longtable,array,tabularx,alltt,float,needspace,wrapfig}
\usepackage{fix-cm}
\newcolumntype{Y}{>{\raggedright\arraybackslash}X}
\newcolumntype{R}{>{\raggedleft\arraybackslash}X}
\definecolor{kblue}{HTML}{2369A8}
\definecolor{korange}{HTML}{D87525}
\definecolor{kpale}{HTML}{EEF5FA}
\definecolor{titlegray}{HTML}{F3EEFA}   
\hypersetup{colorlinks=true,linkcolor=kblue,citecolor=kblue,urlcolor=kblue}
\newcommand{\tool}[1]{\texttt{#1}}

\titlespacing*{\section}{0pt}{3ex plus .5ex minus .3ex}{1.2ex}
\titlespacing*{\subsection}{0pt}{2.3ex plus .4ex minus .2ex}{.9ex}
\titlespacing*{\subsubsection}{0pt}{1.8ex plus .3ex minus .2ex}{.7ex}
\renewcommand\authorformat[2][]{\mbox{\sffamily\bfseries #2\textsuperscript{#1}}}
\newcommand{\projectlabel}{\textbf{\raisebox{.2em}{\begin{tikzpicture}[x=1em,y=1em,line width=.65pt,baseline]
  \draw (0,0) circle (.4);\draw (0,0) ellipse (.18 and .4);\draw (-.4,0)--(.4,0);
\end{tikzpicture}}\; Project Page}}
\newcommand{\rwemptyurl}{}
\newcommand{\projectlinks}{%
  \ifx\projecturl\rwemptyurl
    {\color{kblue}\projectlabel}\,{\color{red}\scriptsize TBD pending}%
  \else\href{\projecturl}{\projectlabel}\fi
  \hspace{1.8em}\href{\codeurl}{\textbf{\texttt{</>}\; Code}}}

\renewcommand{\mymaketitle}{%
  \begin{tcolorbox}[enhanced,frame hidden,colback=titlegray,arc=12pt,
    left=12pt,right=12pt,top=7pt,bottom=7pt,before skip=0pt,after skip=4pt]
    \setlength{\parindent}{0pt}\setlength{\parskip}{0pt}
    {\centering\sffamily\bfseries\fontsize{17.4}{20}\selectfont
      \paperdisplaytitle\par}
    \vspace{5pt}
    \begin{minipage}[c]{0.79\linewidth}
      {\centering\fontsize{8.6}{10.4}\selectfont\authorlist\par}
    \end{minipage}\hfill
    \begin{minipage}[c]{0.18\linewidth}
      \centering\papermascot
    \end{minipage}\par
      \vspace{4pt}
      {\centering\sffamily\color{fgcolor}\fontsize{7.6}{9.2}\selectfont
        \paperaffiliations\hspace{2em}\projectlinks\par}
      \vspace{4pt}
      {\centering\institutionlogos\par}
    \vspace{4pt}
    {\fontsize{9.5}{11.2}\selectfont\abstractlist\par}
  \end{tcolorbox}
  \FloatBarrier
}

\definecolor{rwink}{HTML}{302938}
\definecolor{rwblue}{HTML}{F3EEFA}
\definecolor{rwgray}{HTML}{F5F6F7}
\definecolor{rwsand}{HTML}{F5F3F8}
\definecolor{rwline}{HTML}{8B8494}
\definecolor{rwmuted}{HTML}{625B6B}
\definecolor{rwpositive}{HTML}{10B981}
\definecolor{rwnegative}{HTML}{EF4444}
\DeclareRobustCommand{\rwcheck}{\textcolor{rwpositive}{\ding{51}}}
\DeclareRobustCommand{\rwcross}{\textcolor{rwnegative}{\ding{55}}}
\newcommand{\rwhead}[1]{{\leavevmode\sffamily\bfseries\color{black}#1}}

\newcommand{\rwtable}{\setlength{\parskip}{0pt}\renewcommand{\arraystretch}{1.28}\setlength{\tabcolsep}{5pt}\arrayrulecolor{black}}
\makeatletter
\def\rwpath@arm{\draw (.08,.08)--(.70,.08);\draw (.22,.10)--(.22,.27)--(.49,.53)--(.35,.79)--(.70,.92);\draw (.70,.92)--(.86,.79)--(.79,.66);\draw (.70,.92)--(.61,.75);\fill (.22,.27) circle(.055);\fill (.49,.53) circle(.055);\fill (.35,.79) circle(.055);}
\def\rwpath@hand{\draw (.28,.13)--(.17,.43)--(.09,.60)..controls(.06,.71)and(.17,.73)..(.26,.60)--(.32,.51)--(.32,.87)..controls(.32,1.02)and(.44,1.02)..(.44,.87)--(.44,.61);\draw (.44,.83)..controls(.44,1.03)and(.56,1.03)..(.56,.84)--(.56,.61);\draw (.56,.78)..controls(.56,.97)and(.68,.97)..(.68,.77)--(.68,.57);\draw (.68,.67)..controls(.68,.86)and(.80,.86)..(.80,.65)--(.80,.40)--(.66,.13)--cycle;}
\def\rwpath@mobile{\draw (.10,.20) rectangle(.73,.40);\draw (.27,.40)--(.27,.63)--(.55,.86)--(.75,.65);\draw (.75,.65)--(.89,.67);\draw (.18,.14) circle(.07);\draw (.65,.14) circle(.07);\fill (.27,.63) circle(.05);\fill (.55,.86) circle(.05);}
\def\rwpath@biped{\draw (.49,.87) circle(.105);\draw (.49,.76)--(.49,.40);\draw (.23,.69)--(.49,.65)--(.76,.70);\draw (.49,.40)--(.29,.10)--(.14,.10);\draw (.49,.40)--(.68,.10)--(.84,.10);}
\def\rwpath@dog{\draw (.18,.58)--(.72,.58)--(.83,.70)--(.95,.58)--(.84,.48);\draw (.22,.55)--(.13,.31)--(.26,.12);\draw (.39,.55)--(.31,.31)--(.42,.12);\draw (.67,.55)--(.58,.31)--(.71,.12);\draw (.79,.49)--(.75,.28)--(.86,.12);\draw (.18,.58)--(.06,.74);}
\def\rwpath@car{\draw (.07,.27)--(.07,.51)--(.23,.57)--(.36,.77)--(.66,.77)--(.80,.56)--(.94,.49)--(.94,.27)--(.82,.27);\draw (.18,.27) circle(.10);\draw (.74,.27) circle(.10);\draw (.29,.27)--(.63,.27);\draw (.31,.57)--(.69,.57);}
\def\rwpath@drone{\draw (.24,.24)--(.76,.76);\draw (.24,.76)--(.76,.24);\draw (.39,.39) rectangle(.61,.61);\foreach \x/\y in {.2/.2,.2/.8,.8/.2,.8/.8}{\draw (\x,\y) ellipse(.18 and .10);}}
\def\rwpath@model{\draw (.12,.18) rectangle(.87,.78);\draw (.26,.36)--(.37,.49)--(.26,.62);\draw (.52,.35)--(.71,.35);\draw (.35,.08)--(.65,.08);\draw (.5,.08)--(.5,.18);}
\def\rwpath@eye{\draw (.04,.5)..controls(.28,.86)and(.72,.86)..(.96,.5)..controls(.72,.14)and(.28,.14)..(.04,.5);\draw (.5,.5) circle(.15);}
\def\rwpath@check{\draw (.15,.12) rectangle(.82,.91);\draw (.29,.50)--(.44,.34)--(.70,.70);}
\def\rwpath@clock{\draw (.50,.49) circle(.39);\draw (.50,.49)--(.50,.77);\draw (.50,.49)--(.71,.37);}
\newcommand{\rwicon}[1]{\tikz[x=1.15em,y=1.15em,baseline=-.13em,line width=.65pt,line cap=round,line join=round,rounded corners=0pt,draw=rwink]{\path[use as bounding box](0,0)rectangle(1,1);\csname rwpath@#1\endcsname}}
\makeatother

\newcommand{\rwtablemono}{\rwtable\arrayrulecolor{black}\renewcommand{\rwhead}[1]{{\leavevmode\sffamily\bfseries\color{black}##1}}}

\usepackage{hyperref}
\newcommand{\mail}[1]{\href{mailto:#1}{\texttt{#1}}}
\newcommand{\papertitle}{RobotWorld: Benchmarking Multimodal Agents for Robot Use Across Diverse Tasks and Embodiments}
\newcommand{\paperdisplaytitle}{\mbox{\raisebox{-.20em}{\paperlogo}:} Benchmarking Multimodal Agents for Robot Use Across Diverse Tasks and Embodiments}
\title{\papertitle}
\author[*]{Zhiqin Yang}
\author[*,$\dag$]{Chenxin Li}
\author[*,$\dag$]{Xiaomeng Hu}
\author{Yibin Liu}
\author{Weidong Huang}
\author{Jiankai Sun}
\author{Haitao Li}
\author{Zijian Wu}
\author{Yuzhi Huang}
\author{Fanding Huang}
\author{Hanwen Sun}
\author{Jiashun Liu}
\author{Jingqi Tong}
\author{Mingxin Huang}
\author{Shaoli Hu}
\author{Shijue Huang}
\author{Tianyi Bai}
\author{Xinyuan Wang}
\author{Yunlong Lin}
\author{Zhengyang Tang}
\author{Zhexin Zhang}
\author{Zhuo Chen}
\author{Xierui Song}
\author{Juntao Dai}
\author{Boyuan Chen}
\author{Jiaming Ji}
\author{Fangneng Zhan}
\author{Mengkang Hu}
\author{Wei Xue}
\author{Yonggang Zhang}
\author{Han Hu}
\author{Tsung-Yi Ho}
\author{Yike Guo}

\newcommand{\paperaffiliations}{%
  $^{*}$Equal contribution\qquad $^{\dag}$Project leads}
\newcommand{\institutionlogos}{%
  \noindent%
  \begin{minipage}[c][.85cm][c]{.087\linewidth}
    \centering\includegraphics[width=.94\linewidth,height=.50cm,keepaspectratio]{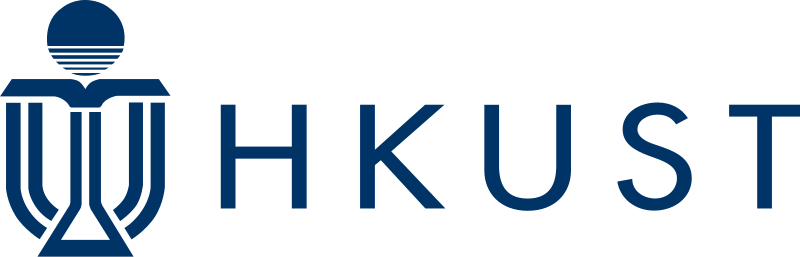}%
  \end{minipage}\hfill%
  \begin{minipage}[c][.85cm][c]{.087\linewidth}
    \centering\includegraphics[width=.94\linewidth,height=.78cm,keepaspectratio]{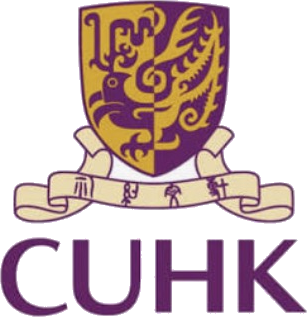}%
  \end{minipage}\hfill%
  \begin{minipage}[c][.85cm][c]{.087\linewidth}
    \centering\includegraphics[width=.94\linewidth,height=.50cm,keepaspectratio]{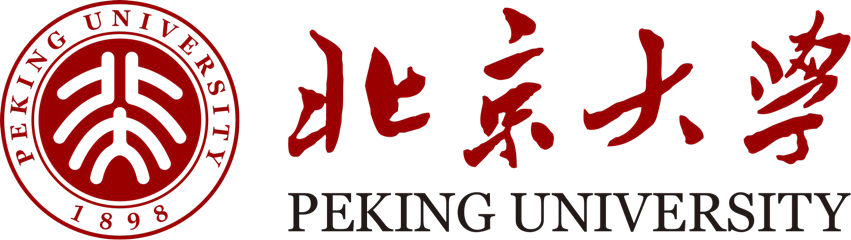}%
  \end{minipage}\hfill%
  \begin{minipage}[c][.85cm][c]{.087\linewidth}
    \centering\includegraphics[width=.94\linewidth,height=.50cm,keepaspectratio]{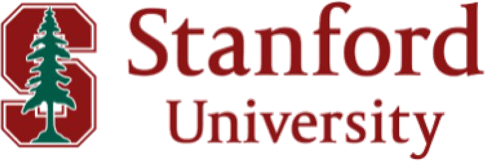}%
  \end{minipage}\hfill%
  \begin{minipage}[c][.85cm][c]{.087\linewidth}
    \centering\includegraphics[width=.94\linewidth,height=.50cm,keepaspectratio]{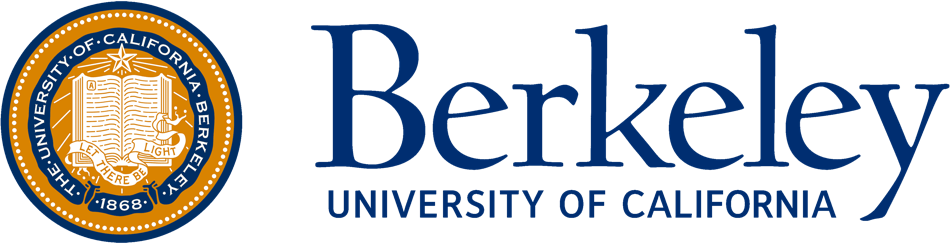}%
  \end{minipage}\hfill%
  \begin{minipage}[c][.85cm][c]{.087\linewidth}
    \centering\includegraphics[width=.94\linewidth,height=.50cm,keepaspectratio]{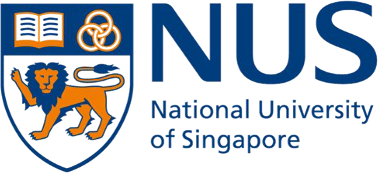}%
  \end{minipage}\hfill%
  \begin{minipage}[c][.85cm][c]{.087\linewidth}
    \centering\includegraphics[width=.94\linewidth,height=.50cm,keepaspectratio]{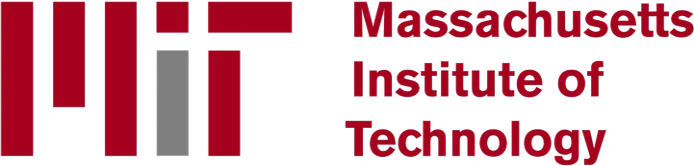}%
  \end{minipage}\hfill%
  \begin{minipage}[c][.85cm][c]{.087\linewidth}
    \centering\includegraphics[width=.94\linewidth,height=.50cm,keepaspectratio]{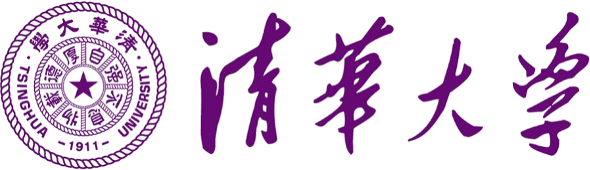}%
  \end{minipage}\hfill%
  \begin{minipage}[c][.85cm][c]{.087\linewidth}
    \centering\includegraphics[width=.94\linewidth,height=.50cm,keepaspectratio]{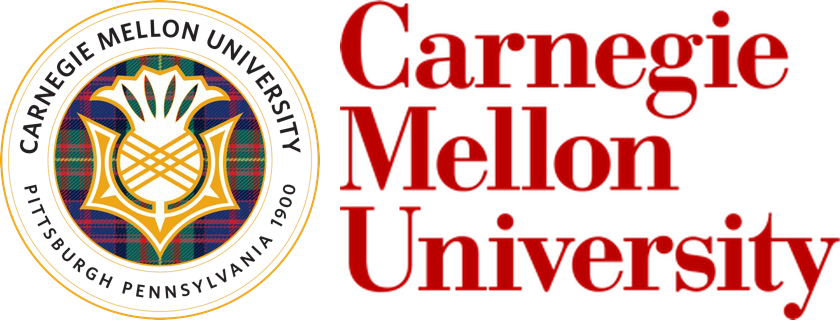}%
  \end{minipage}\hfill%
  \begin{minipage}[c][.85cm][c]{.087\linewidth}
    \centering\includegraphics[width=.94\linewidth,height=.50cm,keepaspectratio]{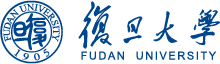}%
  \end{minipage}\hfill%
  \begin{minipage}[c][.85cm][c]{.087\linewidth}
    \centering\includegraphics[width=.94\linewidth,height=.70cm,keepaspectratio]{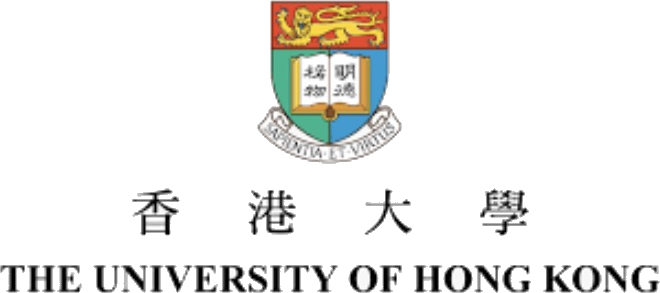}%
  \end{minipage}}
\newcommand{\projecturl}{https://robotworldai.github.io/}
\newcommand{\codeurl}{https://github.com/robotworldai/robotworld}
\newcommand{\bench}{RobotWorld}

\newcommand{\paperlogo}{%
  \includegraphics[height=1.18em]{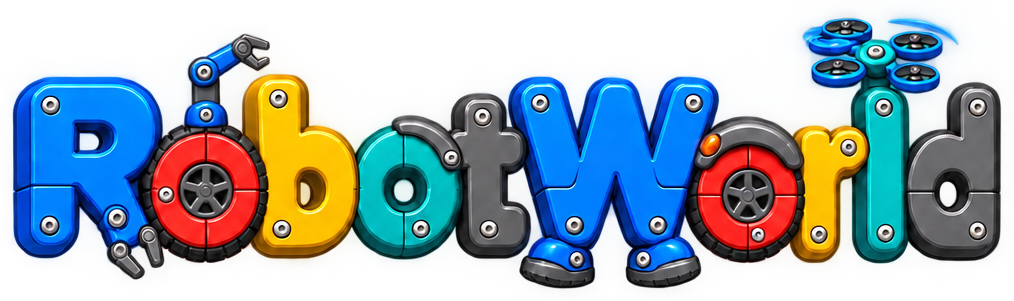}}
\newcommand{\papermascot}{%
  \includegraphics[trim=120bp 115bp 29bp 232bp,clip,
    width=\linewidth,height=2.4cm,keepaspectratio]{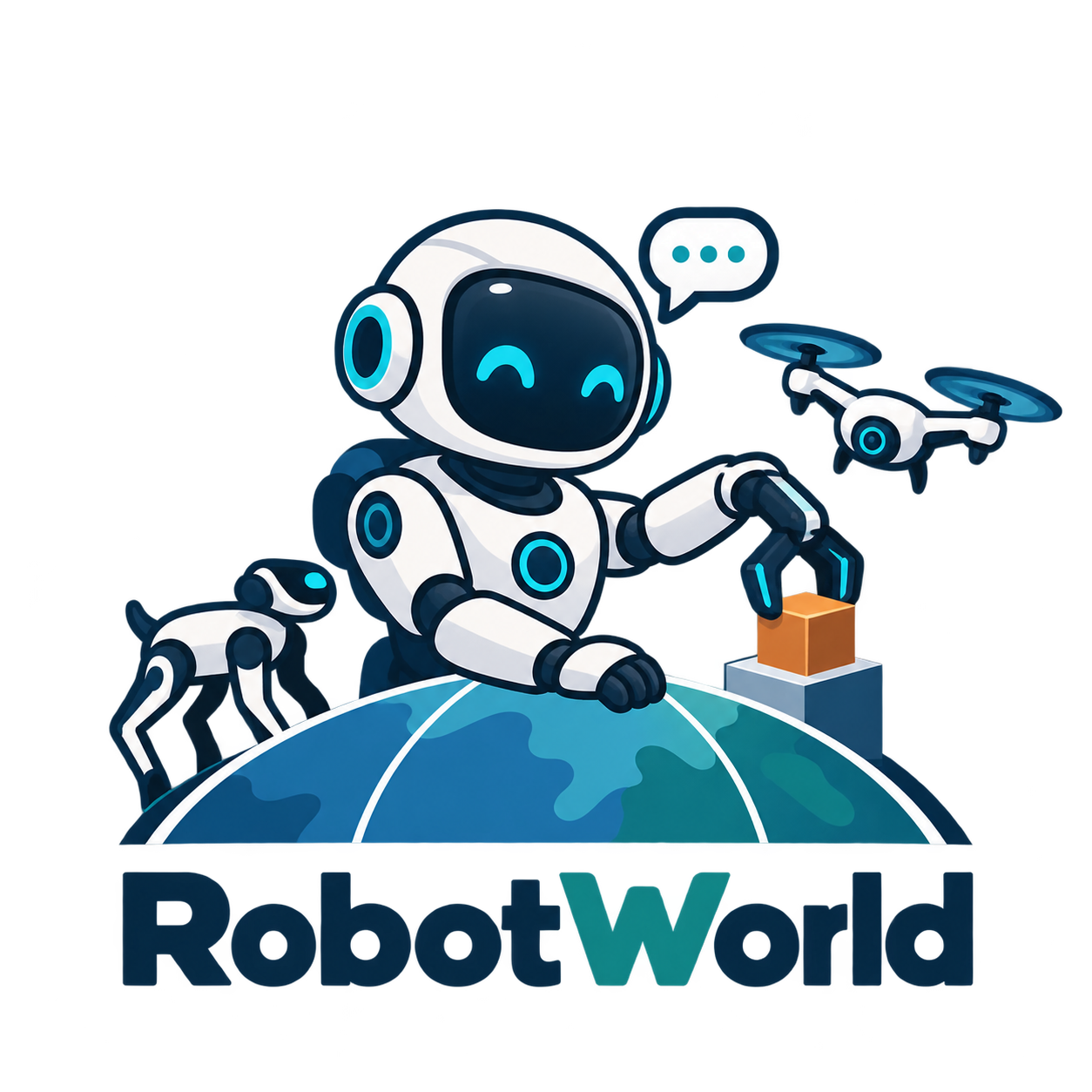}}
\abstract{
General-purpose agents increasingly write code, use tools, and complete complex digital tasks, raising the question of how far these capabilities carry into the physical world. To investigate this, we introduce \textbf{\bench{}}, a challenging simulation testbed for \emph{robot use}: turning instructions and observations into physical task execution through robot interfaces. Its 84 tasks span manipulation, mobile manipulation, locomotion, driving, and aerial control, with explicit interaction budgets and executable success checks. By analysing task outcomes alongside execution traces, we identify both the capabilities that transfer and the gaps that prevent reliable completion. Furthermore, we find that current agents can construct sophisticated perception and control workflows, including image segmentation, camera calibration, spatial estimation, and dynamics-based computation. These capabilities, however, do not consistently compose into successful behaviour: agents lose task-relevant object states despite reaching commanded poses, fail to correct ineffective actions, recover too late, or mistake unfinished tasks for completion. This uneven transfer also differs across models: Astra succeeds more often on spatial and constrained-contact goals, whereas Opus 5.5 succeeds more often on continuous-balance and timed-interaction goals. By linking these outcomes to execution behaviour, \bench{} provides both a rigorous proving ground and an empirical account of the remaining capability gaps, thereby establishing concrete targets for training and designing more reliable physical-world agents.}
\hypersetup{pdftitle={\papertitle},pdfauthor={RobotWorld authors pending}}
\hypersetup{pdftitle={\papertitle},pdfauthor={Author One, Author Two}}

\begin{document}
\addtocontents{toc}{\protect\setcounter{tocdepth}{-1}}
\maketitle
\begingroup
\renewcommand{\thefootnote}{}
\footnotetext{\textbf{Contact:} \mail{robotworld.ai@gmail.com}}
\endgroup

\begin{figure}[H]
\centering
\includegraphics[width=\linewidth]{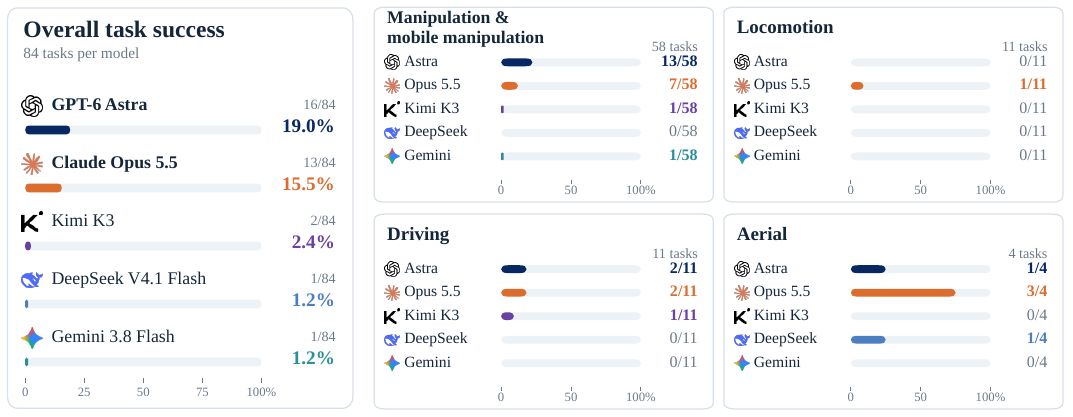}
\captionsetup{font=small,skip=2pt}
\caption{\textbf{RobotWorld leaderboard.} Left: overall success across 84 tasks. Right: success in four displayed groups (Manipulation and Mobile manipulation are pooled), with counts shown beside each bar.  All panels use the same 0--100\% success scale. Each model has one retained outcome per task under the evaluation budget rules.}
\label{fig:front-leaderboard}
\end{figure}

\clearpage

% Optional first-page figure: see snippets.tex.
% \begin{quote}
% \textbf{Positioning.} \bench{} evaluates whether a general-purpose model can turn task instructions and sensory feedback into successful physical control across diverse robot bodies and dynamics in simulation.
% \end{quote}
% Section dispatcher: edit the individual files below.
\section{Introduction}
General-purpose agents are becoming useful actors in the digital world. They develop software, operate browsers, and carry out multi-step knowledge-work tasks through frontier harnesses and systems such as Operator and Claude Cowork~\citep{operator2025,cowork2026}. Their capabilities extend beyond producing answers: they inspect an environment, use tools, observe the consequences, and revise their actions. Robotics presents a natural next question: \textbf{how far can these capabilities take agents into the physical world?} Recent studies of frontier models, including GPT-6 Astra, have made robot control an increasingly concrete possibility~\citep{clauderobotics2026,astrapolicies2026}. Community experiments now include robot-arm control and simulated dexterous apple grasping, while real-time robot rollouts also expose the gap between an impressive demonstration and practical execution speed~\citep{chooi2026robot,huang2026dexlab,duan2026realtime}. The challenge is to establish which capabilities already transfer, which remain missing, and what should improve next.

These experiments have also prompted debate about what progress in robot use should mean. Malik questions whether pick-and-place demonstrations establish competence in dexterity and dynamics; Isola and Fu emphasise the potential of agents that use controllers and robotics tools~\citep{malik2026control,isola2026controllers,fu2026harnesses}. This motivates a broader empirical question: across which physical demands can a general-purpose agent act reliably, with what control support, and where does its feedback loop fail? Physical interaction tests how perception, reasoning, and program construction work together. An agent may estimate a grasp from images, write a segmentation program, fit camera geometry, or calculate a controller from approximate dynamics. Each is useful only insofar as it supports successful action. Describing a grasp does not secure the object; reaching a target pose does not establish insertion; recovering balance for an instant does not satisfy sustained stability. Across different robot bodies, the agent must interpret its observations, anticipate action consequences, correct errors, and judge completion. We use \emph{robot use} to describe this ability to turn instructions and feedback into task execution through robot interfaces.

Recent work is turning general-purpose models into agents that execute robot tasks through programs, tools, and feedback. EmbodiedSWE studies coding agents that construct executable solutions for long-horizon dexterous tasks~\citep{embodiedswe}. PyRUA-Lean investigates how programmatic action composition and selective observation improve execution efficiency~\citep{pyrualean2026}, while RPG uses simulation practice and failure diagnosis to refine reusable robot skills~\citep{rpg2026}. These advances make systematic evaluation increasingly important: across which robot bodies and physical demands do current agents act reliably, and where does execution break down? We study this question across manipulation, locomotion, driving, and flight, connecting task outcomes to how agents combine perception, computation, and feedback. Our aim is to turn promising demonstrations into measurable capability gaps and concrete targets for future training and agent design.

\begin{figure}[!t]
  \centering
  \includegraphics[width=1\linewidth]{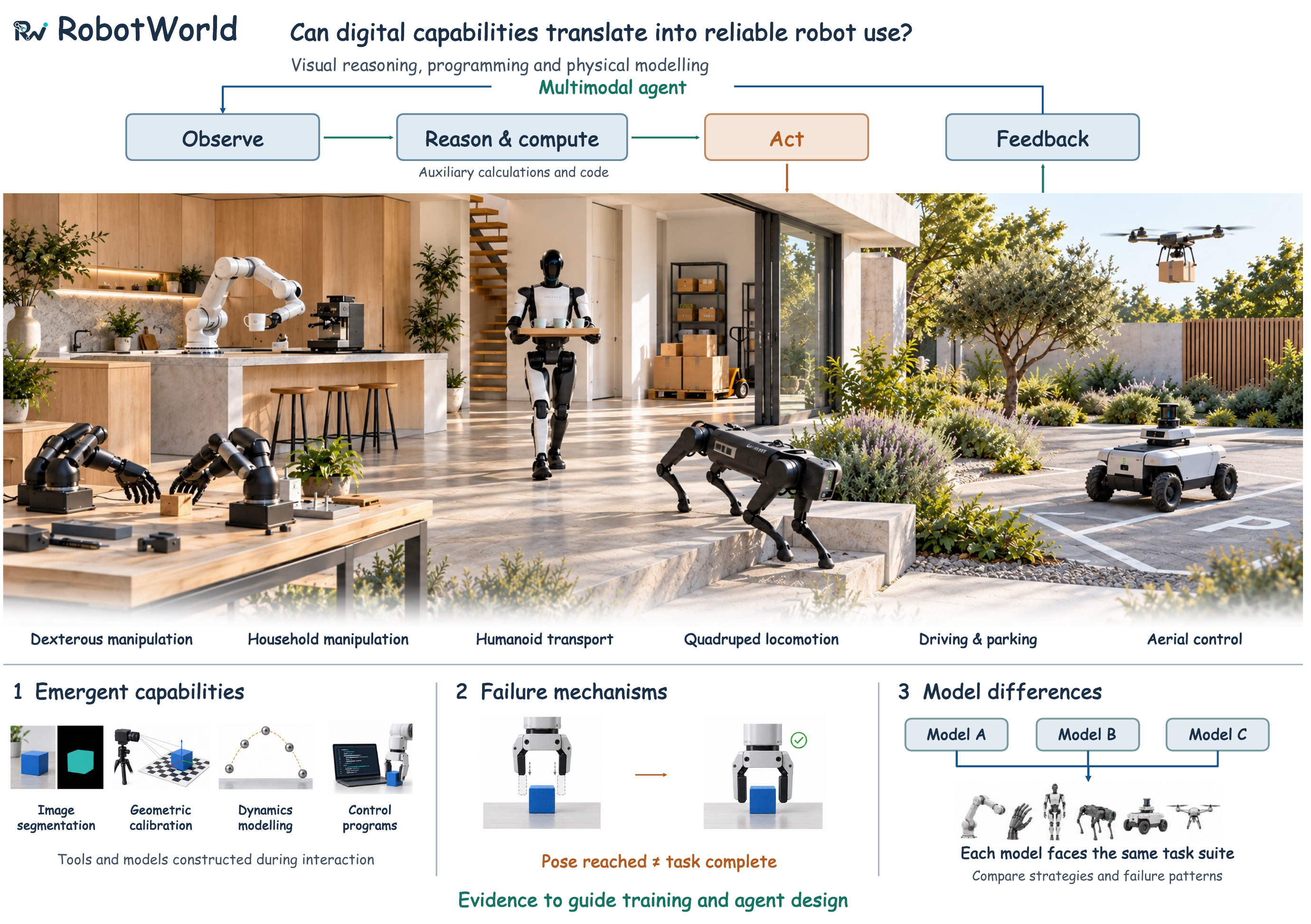}
  \caption{\textbf{Overview of \bench{}.} RobotWorld evaluates whether general-purpose multimodal models can turn digital capabilities into reliable robot use. Agents interpret instructions and observations, perform auxiliary computation, and act through a feedback loop across diverse robots and tasks. Trajectory analysis examines emerging capabilities---including image segmentation, geometric calibration, dynamics modelling and control program generation---alongside failure mechanisms and differences between models on shared tasks. These findings inform training and agent design. The scene illustrates task families evaluated in separate simulation environments.}
  \label{fig:overview}
\end{figure}

\textbf{We introduce \bench{}, a simulation benchmark of robot use across diverse tasks and embodiments.} It brings together 84 task specifications with documented robot interfaces, task budgets, and executable success checks (see Figure~\ref{fig:overview}). Multimodal agents interact through a shared execution harness, with each task's observations and controller assistance specified. Recorded actions, tool interactions, and outcomes support analysis beyond aggregate scores. Sections~\ref{sec:environment}--\ref{sec:evaluation} describe the environments, benchmark construction, and evaluation protocol; our central empirical question is what these executions reveal about agents' physical capabilities.

Our trajectory analysis reveals both substantial capabilities and persistent gaps. Agents construct perception and control workflows involving colour segmentation, camera-geometry fitting, spatial estimation, and numerical dynamics calculations. They also use feedback to revise motions and recover from execution errors. Yet these local capabilities do not consistently compose into task completion: a robot may reach the requested pose while losing the object, remain in navigation without reaching the next subgoal, regain stability too late, or stop correcting an unfinished task because it believes the goal is already met. These failures locate weaknesses in state interpretation, action revision, temporal coordination, and completion assessment.

The models also exhibit different task and interaction profiles. Across the task groups analysed in Section~\ref{sec:model-profiles}, Astra succeeds more often on spatial and constrained-contact goals, whereas Opus 5.5 succeeds more often on continuous-balance and timed-interaction goals. Among eight tasks solved by both, Astra uses fewer tool calls in seven but fewer control steps in only six, showing that interaction cost and physical execution cost need not agree. Together with the observed workflows, these patterns suggest different ways of composing visual-spatial reasoning, explicit computation, and feedback control. We treat this as an exploratory explanation of the model profiles. By connecting outcomes to execution mechanisms, \bench{} aims to make evaluation useful for directing progress towards more capable physical-world agents.

Our contributions are twofold:
\begin{enumerate}[leftmargin=*,itemsep=3pt]
\item \textbf{A challenging, rigorous proving ground for robot use.} \bench{} gives the community a systematic way to test whether general-purpose agents can turn digital capabilities into successful physical action. Its 84 tasks from 20 source projects span diverse embodiments and control demands, with explicit interfaces, budgets, and executable success checks. Substantial remaining headroom makes it a testbed for developing and measuring the next generation of physical-world agents.
\item \textbf{An empirical capability profile and directions for improvement.} Through execution traces and cross-model comparisons, we reveal how current agents combine perception, computation, and control, where reliable task completion breaks down, and how strengths and weaknesses differ across models. These findings identify concrete priorities for future training and agent design, including grounding perception and computation in action, recovering from execution errors, coordinating actions over time, and recognising task completion.
\end{enumerate}

\section{Related Work}
\label{sec:related}
% Comparison is scoped to the cited protocols, including their evaluated variants.
% Direct actions include numerical actions and symbolic action sequences supplied
% without generating a control program; they need not be low-level motor commands.
% CaP-X: paper Sections 2--4; code REPL; Section 4.2 evaluates navigation in BEHAVIOR.
% EmbodiedBench: Sections 4.1 and Appendix B; discrete commands and 7-D arm actions.
% VLABench: Sections 3.2 and 4; direct VLA actions and code-based VoxPoser workflow
% are separate evaluations, not mixed calls within a single episode.
% EmbodiedEval: Section 3.2; discrete movement / interaction / answer actions.
% EAI: Section 2.5 evaluates explicit symbolic action sequences in the simulator;
% navigation is part of its household action vocabulary, not legged balance control.
% EmbodiedSWE: Appendix B.1 includes three locomanipulation tasks; B.3 supplies
% a learned locomotion controller, not a separate body-stability task objective.
% Mixed calls means distinct direct-action and code-execution interfaces selected
% within one episode; primitive calls made inside generated Python do not count.
% RobotWorld code control is optional and disabled in the primary evaluation.
\begin{table}[t]
\centering\small\color{black}
\DeclareRobustCommand{\rwtick}{\rwcheck}
\DeclareRobustCommand{\rwno}{\rwcross}
\caption{Control interfaces and task coverage (\rwtick: included; \rwno: not reported in the cited protocols). Direct actions are model-emitted commands or action values; code control executes generated programs. Mixed calls alternate these interfaces within one episode. Balance denotes tasks with an explicit robot-stability objective. Optional \bench{} code control is included.}
\label{tab:related_evaluation_scope}
\label{tab:related_protocols}
\rwtable
\setlength{\tabcolsep}{6pt}
\renewcommand{\arraystretch}{1.65}
\resizebox{\linewidth}{!}{%
\begin{tabular}{@{}l*{6}{c}@{}}
\toprule
\rwhead{Evaluation} & \multicolumn{3}{c}{\rwhead{Control interfaces}} & \multicolumn{3}{c}{\rwhead{Beyond manipulation}} \\
\cmidrule(lr){2-4}\cmidrule(lr){5-7}
 & \rwhead{Direct actions} & \rwhead{Code control} & \rwhead{Mixed calls} & \rwhead{Navigation} & \rwhead{Balance} & \rwhead{Driving / flight} \\
\midrule
CaP-X~\citep{capx} & \rwno & \rwtick & \rwno & \rwtick & \rwno & \rwno \\
EmbodiedBench~\citep{embodiedbench} & \rwtick & \rwno & \rwno & \rwtick & \rwno & \rwno \\
VLABench~\citep{vlabench} & \rwtick & \rwtick & \rwno & \rwno & \rwno & \rwno \\
EmbodiedEval~\citep{embodiedeval2025} & \rwtick & \rwno & \rwno & \rwtick & \rwno & \rwno \\
Embodied Agent Interface~\citep{eai} & \rwtick & \rwno & \rwno & \rwtick & \rwno & \rwno \\
EmbodiedSWE-Bench~\citep{embodiedswe} & \rwno & \rwtick & \rwno & \rwtick & \rwno & \rwno \\
\midrule
\rowcolor[gray]{0.94}
\textbf{\bench{} (ours)} & \rwtick & \rwtick & \rwtick & \rwtick & \rwtick & \rwtick \\
\bottomrule
\end{tabular}%
}
\end{table}

\paragraph{Foundation Models for Robotics.}
Vision--language--action (VLA) models such as RT-2, OpenVLA and $\pi_0$ generate actions from visual observations and language instructions~\citep{rt2,openvla,pi0}; $\pi_{0.5}$ further studies open-world generalisation through heterogeneous co-training~\citep{pi05}. World--action models (WAMs) connect action generation with future-observation prediction. Recent work examines complementary aspects of this coupling: SelfWAM models action-conditioned visual consequences~\citep{selfwam2026}, OpenWAM systematically studies world--action pretraining and information flow~\citep{openwam2026}, and Rolling-WAM distributes joint video--action denoising across replanning cycles to improve responsiveness~\citep{rollingwam2026}. These directions study closely related capabilities: selecting actions from sensory inputs and modelling how actions change the world. Robot-use agents can express both through reasoning, programs and feedback. Our trajectories include direct visual estimates used to choose motions, as well as geometric and dynamics calculations used to anticipate motion and derive control commands. \bench{} examines how these capabilities are combined during execution and whether they produce reliable task completion across diverse physical demands.

\paragraph{Agents for Robot Use.}
Agents use reasoning, programs, and tools to connect instructions to physical action. Code as Policies and VoxPoser generate executable control logic and spatial representations for robot motion~\citep{codeaspolicies,voxposer}, while Show-Harness exposes semantic robot actions to VLMs~\citep{showharness}. EmbodiedSWE studies coding agents that construct executable solutions for dexterous and mobile manipulation~\citep{embodiedswe}. Recent work develops how such agents use execution feedback: PyRUA-Lean composes robot primitives into Python cells with conditional checks, local retries, and selective observation~\citep{pyrualean2026}; ASPIRE accumulates validated program repairs into reusable skills~\citep{aspire2026}; and RPG uses simulation practice and failure diagnosis to revise a shared skill library and system prompt~\citep{rpg2026}. These approaches develop complementary ways to organise robot execution, from decisions within an episode to skills retained across tasks. RLE-Bench examines the broader use of coding agents for robotics research and engineering, including policy learning, perception, mechanical design, and interactive control~\citep{rlebench}. Our focus is the breadth and reliability of robot use: how agents combine available capabilities during execution, and which mechanisms support or obstruct task completion.

\paragraph{Benchmarks for Embodied Agents.}
CaP-X evaluates coding agents for manipulation under different primitive abstractions and feedback settings, with extensions to mobile manipulation~\citep{capx}. EmbodiedBench evaluates visual agents across high-level planning and lower-level navigation and manipulation, with capability-oriented task subsets~\citep{embodiedbench}. VLABench examines language-conditioned manipulation and long-horizon reasoning, evaluating both action policies and VLM-based workflows~\citep{vlabench}; EmbodiedEval spans navigation, object and social interaction, and embodied question answering~\citep{embodiedeval2025}. At the decision-module level, Embodied Agent Interface separately evaluates goal interpretation, subgoal decomposition, action sequencing, and transition modelling~\citep{eai}. These agent evaluations build on a wider ecosystem of robot-learning task suites, including RLBench, CALVIN, LIBERO, BEHAVIOR-1K, and RoboCasa365~\citep{rlbench,calvin,libero,behavior1k,robocasa365}. \bench{} brings together diverse embodiments and physical demands, including spatial manipulation, sustained stabilisation, and interaction with moving targets. Alongside task outcomes, we analyse feedback use, recovery, completion assessment, and interaction cost to characterise model strengths and failure mechanisms. Table~\ref{tab:related_evaluation_scope} compares control interfaces and coverage beyond manipulation. Direct commands and generated programs are distinct entry points: \bench{} can interleave both within one episode, while retaining the declared native controllers.

\begingroup
\clubpenalty=10000
\widowpenalty=10000
\section{RobotWorld Environment}
\label{sec:environment}
\bench{} provides a common interaction protocol for robot use across heterogeneous simulators and embodiments. Given an instruction, an agent inspects observations, performs auxiliary computation, and issues robot commands. Execution returns observations and feedback for the next decision. The overall architecture is shown in Figure~\ref{fig:framework}. It connects observations, auxiliary computation, robot tools, and evaluation while preserving each source system's physics, action semantics, and controller assistance.

\begin{figure}[t]
  \centering
  \includegraphics[width=\linewidth]{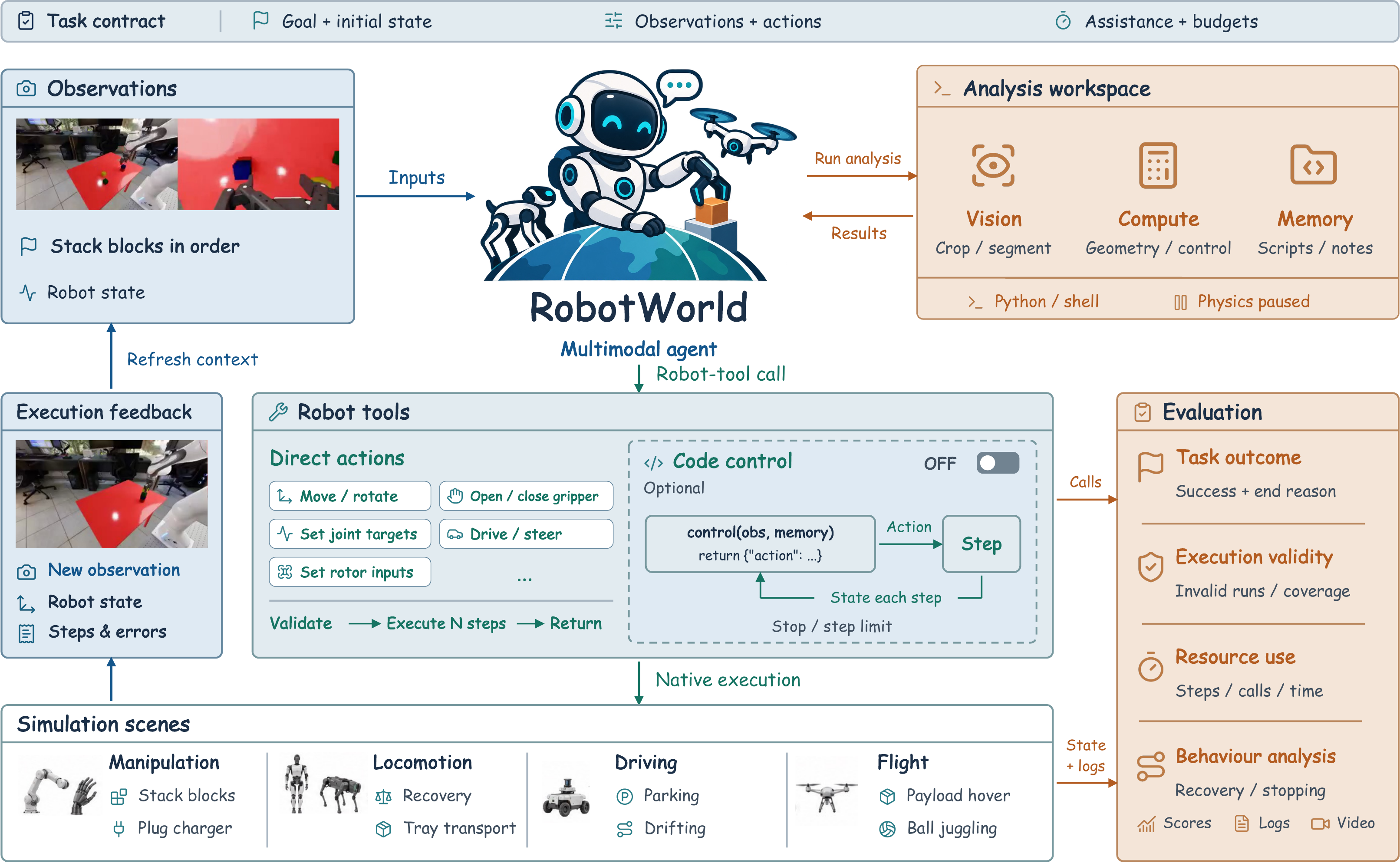}
  \caption{\textbf{The \bench{} framework.} Under a task-specific observation, action, and budget contract, a multimodal agent alternates between analysis and robot execution, using returned observations and execution feedback to revise its actions. Direct actions execute bounded command segments; optional code control, disabled by default, runs generated programs with feedback at each control step. The analysis workspace supports perception, computation, and memory while physics is paused. Evaluation records task outcomes, execution validity, resource use, and behaviour across manipulation, locomotion, driving, and flight.}
  \label{fig:framework}
\end{figure}

\subsection{Task Definition and Interaction Loop}
A task specifies both a physical goal and the conditions under which the agent must achieve it. We represent task $i$ as
\begin{equation}
\mathcal T_i=(\mathcal E_i,\rho_i,g_i,O_i,A_i,C_i,B_i,V_i),
\end{equation}
where $\mathcal E_i$ is the environment, $\rho_i$ defines initial conditions, and $g_i$ is the goal. The observation interface $O_i$, action interface $A_i$, and controller assistance $C_i$ define the agent's access to the robot. Budgets $B_i$ bound interaction, and evaluator $V_i$ assesses the executed trajectory. Together, these elements form a task contract: they expose differences in sensing and control that a shared tool protocol alone would otherwise conceal.

\Needspace{5\baselineskip}
At interaction $k$, the agent selects a response using the goal, interface documentation $D_i$, and permitted history $h_k$:
\begin{equation}
z_k\sim\pi_\theta(\cdot\mid g_i,D_i,h_k).
\end{equation}
The response may request robot execution, invoke workspace computation, or produce text. Model parameters remain fixed during an episode; observations, conversation history, and permitted workspace contents evolve as the agent acts. A frontier harness manages model conversations and tools, while separate environment processes execute robot requests and return feedback. As shown in Figure~\ref{fig:interaction-loop}, a turn proceeds from observation and optional computation to a bounded execution request and returned feedback. Task evaluation follows a separate state-and-event path.

\begin{figure}[!htbp]
\centering
\includegraphics[width=\linewidth]{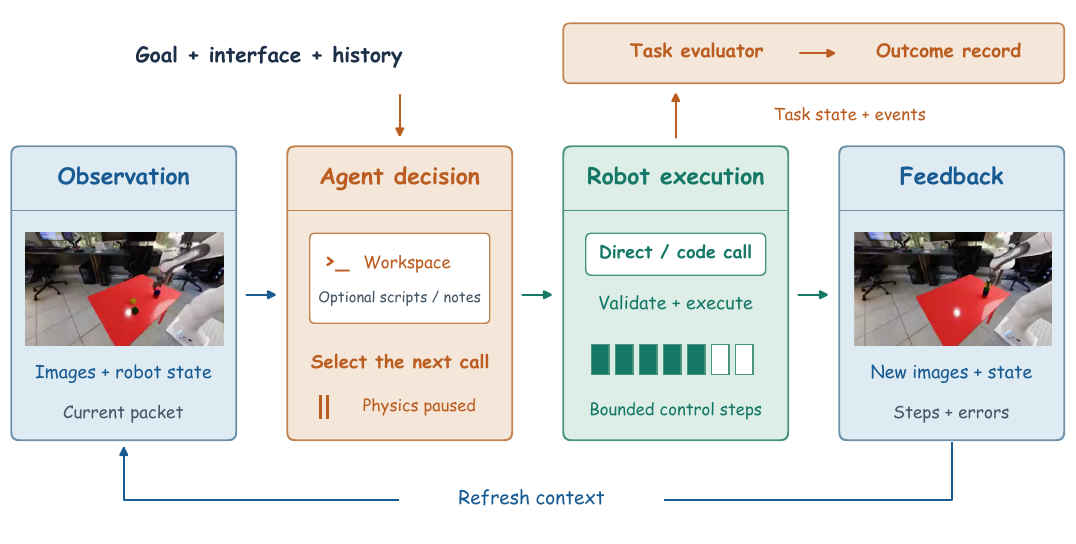}
\caption{\textbf{Environment interaction loop.} Observations and task context inform the agent's next robot call, optionally supported by workspace computation. Native execution validates the request and advances physics within its step limit; returned observations, completed steps, and errors inform the next decision. Physics pauses during reasoning and offline computation. The evaluator checks task state and events independently. Camera views illustrate observations from a stacking episode, rather than the endpoints of a single call; the execution strip is schematic.}
\label{fig:interaction-loop}
\end{figure}

\subsection{Observations and Analysis Workspace}
\noindent\textbf{Permitted observations.} The task instruction and interface documentation accompany camera images, available robot state, and execution feedback. Observation channels are task-specific: an interface may expose joint positions or object measurements in addition to images. Evaluator-only state and cameras used solely to review trajectories remain outside the agent interface. Consequently, the information used to judge an outcome can be richer than the information available to achieve it.

\Needspace{3\baselineskip}
\noindent\textbf{Auxiliary computation.} Where enabled, an isolated workspace supports image inspection, geometric estimation, numerical calculations, and the development of control programs. Scripts and notes can retain intermediate results within an episode. Workspace tools operate on supplied observations and permitted files; they neither access simulator internals nor advance physics. Workspace access and code control are configured independently: analysing an image does not require permission to run a program inside the control loop.

\noindent\textbf{Observation history.} Timestamped packets distinguish newly returned observations from earlier evidence. Analysis and text-only continuation retain the latest packet without implying that the scene has changed. Action records and workspace notes support later decisions, subject to the configured history window. The experimental setup specifies the retained history and tool access used for model comparison.

\Needspace{29\baselineskip}
\subsection{Robot Actions and Execution Feedback}
\label{sec:assistance}
Robot execution converts the agent's request into a bounded segment of physical interaction. Direct actions are always available; optional code control adds a feedback loop within a call. As shown in Figure~\ref{fig:control_modes}, both modes return control to the agent, which can revise its next request using the resulting observations.

% Let wrapfig determine the number of wrapped lines from the figure height.
\begin{wrapfigure}{r}{0.49\textwidth}
\centering
\includegraphics[width=\linewidth]{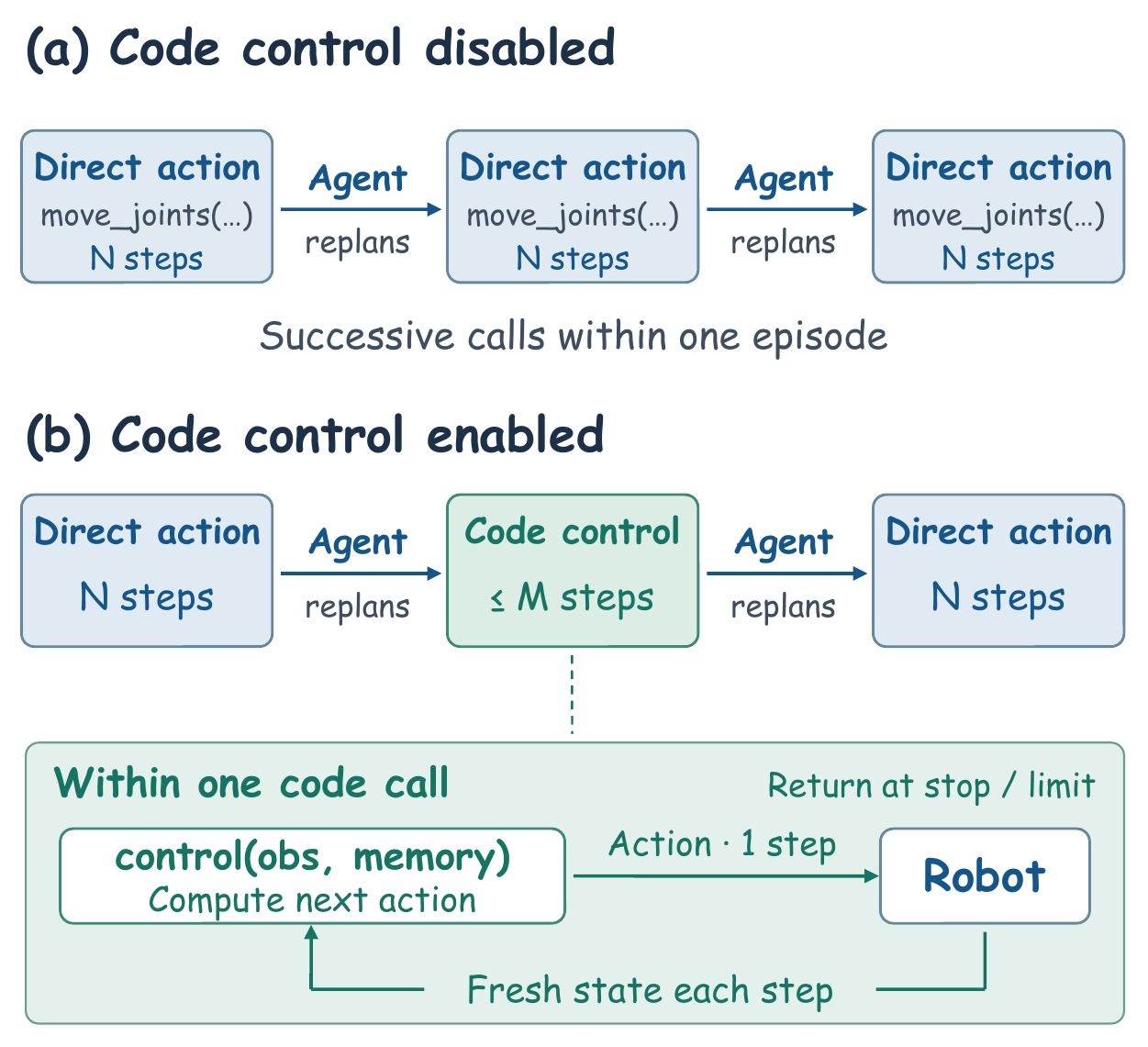}
\captionsetup{font=small,skip=3pt}
\caption{\textbf{Control is chosen per call.} Direct actions and code segments can alternate within one episode. The agent replans between calls; within a code segment, the program updates actions from fresh state at each step, up to its limit of $M$~steps.}

\label{fig:control_modes}
\end{wrapfigure}

\noindent\textbf{Direct actions.} A robot-tool call contains action values or targets together with a step count. For example, \texttt{move\_joints} specifies joint targets. The environment validates the request, executes it through the native controller, and returns permitted observations and execution feedback. This feedback includes completed steps and reported errors, allowing a rejected command to be distinguished from a command that executed but failed to achieve its intended effect.

\Needspace{3\baselineskip}
\noindent\textbf{Code control.} Where supported and enabled, an additional tool runs an agent-generated feedback program~\citep{codeaspolicies}. At each control step, the program reads permitted observations, updates local memory, and selects an action without another model call. Execution is bounded by a step limit and may end earlier on a stop condition, error, or episode termination. If the episode continues, the agent can revise the program or return to direct actions. Thus, code control changes the frequency at which a generated program receives feedback, while the agent still replans between tool calls.

\noindent\textbf{Control semantics and assistance.} Interface documentation specifies units, coordinate frames, bounds, and absolute or incremental commands. Both execution modes retain these semantics and the task's native controllers. Motion generation, such as cuRobo~\citep{curobo} in RoboDojo, and low-level servos are distinguished from learned task or balance policies. These forms of assistance affect the control problem presented to the agent and are part of the task contract; a common protocol does not make all action spaces or control demands equivalent.

\subsection{Episode Execution and Records}
\noindent\textbf{Initialisation and time.} Each episode begins from the prescribed task initialisation. Physics advances during robot execution and pauses while the agent reasons or computes offline. Let $n_k$ denote the cumulative number of native control steps. A request that completes $d_k$ steps gives: $n_{k+1}=n_k+d_k$, with $d_k=0$ for workspace computation or text. A control step may contain several physics substeps; the native control interval determines simulated duration. Control steps, tool interactions, and wall-clock time therefore measure different aspects of resource use.

\noindent\textbf{Call and episode boundaries.} Completing a command segment or returning a stop signal from a program ends that call. Native termination and configured episode limits determine whether further interaction is possible. Neither a completed call nor an agent's textual completion claim establishes task success. The evaluator assesses the relevant states and events during execution, including temporal or history-dependent conditions when required by the task.

\noindent\textbf{Outcome records.} Episode records retain observations, requested and completed actions, checker outcomes, and available errors or termination reasons. These records connect an outcome to the physical execution that produced it and support trajectory and video review. They also allow resource consumption to be examined separately from success. Section~\ref{sec:benchmark} specifies what the task checkers require; the experimental setup defines the model configurations, resource limits, and aggregation used for the reported comparisons.
\FloatBarrier
\endgroup

\begingroup
\clubpenalty=10000
\widowpenalty=10000
\Needspace{24\baselineskip}
\section{RobotWorld Benchmark}
\label{sec:benchmark}
\begin{wrapfigure}{r}{0.49\textwidth}
\centering
\includegraphics[width=\linewidth]{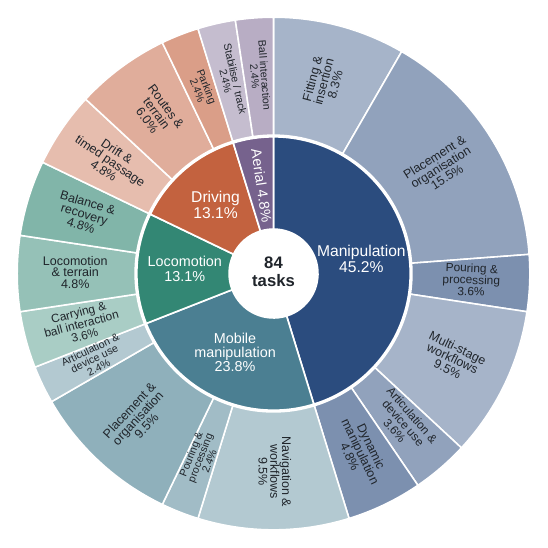}
\captionsetup{font=small,skip=4pt}
\caption{\textbf{Task composition by domain and primary task type.} The inner ring shows five domains and the outer ring their primary task types. Each task is counted once. Percentages use all 84 tasks, not action frequencies.}
\label{fig:task-composition}
\end{wrapfigure}

The environment defines how agents interact with robots. The benchmark defines what they must accomplish and what evidence establishes completion. \bench{} comprises 84 tasks from 20 source projects, spanning manipulation, mobile manipulation, locomotion, driving, and aerial control. Its construction brings these tasks under explicit instructions, interaction conditions, and executable success criteria, enabling outcomes to be interpreted in terms of the physical demands agents encounter.

\subsection{Task Scope and Selection}

Task selection targets complementary forms of robot use: establishing object relations, operating articulated objects, coordinating multiple stages, maintaining stability, and acting on moving targets. Variants contribute distinct tasks when their goals or physical constraints change, such as entry direction, terrain, timing, or coordination. Aliases and alternative control modes do not increase the task count.

\noindent\textbf{Domains and task types.} Figure~\ref{fig:task-composition} summarises the suite at two levels. The five primary domains contain 38 manipulation tasks, 20 mobile manipulation tasks, 11 locomotion tasks, 11 driving tasks, and four aerial tasks. Manipulation uses fixed-base arm or hand interfaces; mobile manipulation uses the mobile base and arm interfaces of RoboCasa and BEHAVIOR-1K. Locomotion covers legged motion and whole-body balance or coordination. Within each domain, a primary type describes the task's main objective, such as fitting and insertion, balance and recovery, or precision parking. Each task contributes once to this composition. These labels characterise the required activity rather than the frequency of low-level actions in a particular model's trajectory. Together, manipulation and mobile manipulation account for 69.0\% of the suite, so the aggregate should be interpreted alongside results for the other domains.

\Needspace{3\baselineskip}
\noindent\textbf{Shared control requirements.} Tasks also share control requirements across domains. \emph{Spatial goals} require particular positions, orientations, destinations, or paths. \emph{Constrained contact} covers interactions such as insertion, hanging, pouring, and articulated-object operation, beyond unconstrained pick-and-place. \emph{Multiple subgoals} requires completing several objects or stages while preserving progress; a routine approach--grasp--lift sequence alone does not qualify. \emph{Balance and tracking} requires sustained regulation of posture, support, or a changing reference. \emph{Timed interaction} requires coordination with an independently moving target or constraint, as in conveyor picking, interception, or moving-gate passage.

These requirements overlap: a task can involve spatial positioning, constrained contact, and several subgoals. Primary types describe what the agent is asked to do; requirement labels describe the challenges involved. Neither constitutes a model performance score or a decomposition into independent abilities.

\subsection{Task Construction and Adaptation}
Manipulation and mobile manipulation tasks draw on BEHAVIOR-1K~\citep{behavior1k}, RoboCasa~\citep{robocasa2024,robocasa365}, RoboLab~\citep{robolab}, RoboDojo~\citep{robodojo}, and Bench2Dex~\citep{bench2dex}, alongside additional control tasks. Driving and aerial environments include WheeledLab~\citep{wheeledlab}, OmniDrones~\citep{omnidrones}, and VolleyBots~\citep{volleybots,jugglerl}. Appendix~\ref{app:spec} attributes all 20 source projects to their papers or software releases.

Each benchmark task combines source simulation assets with an instruction, an initialisation procedure, an interaction contract, and a success checker. We distinguish the origin of the objective from changes to the interface or scenario. An inherited goal can be exposed through a modified robot interface, and a new goal can use an existing scene and controller. Treating these as separate dimensions makes the benchmark's additions explicit.

\noindent\textbf{Objectives and checkers.} When a source defines a suitable goal and executable success condition, the benchmark retains them. For example, selected Bench2Dex tasks use source scene anchors and native event and stability checks. In environments designed around a continuing reward, a reward signal alone may not establish completion of an instruction. Selected locomotion and aerial tasks therefore receive explicit completion objectives, such as reaching a destination after disturbances or stabilising a payload within specified bounds. Surviving to the horizon is insufficient unless it satisfies the stated objective.

\noindent\textbf{Interfaces and scenarios.} Environment adapters expose permitted observations, documented commands, controller assistance, and bounded execution through the interaction protocol in Section~\ref{sec:environment}. Scenario changes specify the conditions in which those interfaces are used. Seven \bench{} driving scenarios, for example, introduce constraints including narrow supports, restricted parking spaces, and a moving gate. Scene geometry and disturbances determine the problem faced by the agent; entry direction, final alignment, or valid gate passage determine whether it has been solved.

\noindent\textbf{Task specification.} Source identity, initialisation, observation and action interfaces, assistance, budgets, and checker definitions jointly identify an evaluation configuration. Changes to the objective are treated as \bench{} task definitions rather than implied changes to the source benchmark's results. Native rewards or auxiliary scores remain distinct from completion under the declared goal. Task-level details and checker thresholds are retained with the benchmark specification.

\subsection{Success Criteria}
\label{sec:qa}
Success is determined from task-relevant simulator states and events in the executed trajectory. Checkers assess physical outcomes rather than agreement with a reference action sequence, allowing different strategies to succeed when they satisfy the same conditions. Figure~\ref{fig:success-criteria} illustrates six overlapping families of evidence. A task may combine several families; each illustration shows a component of a full task checker.

\begin{figure}[!t]
\centering
\includegraphics[width=\linewidth]{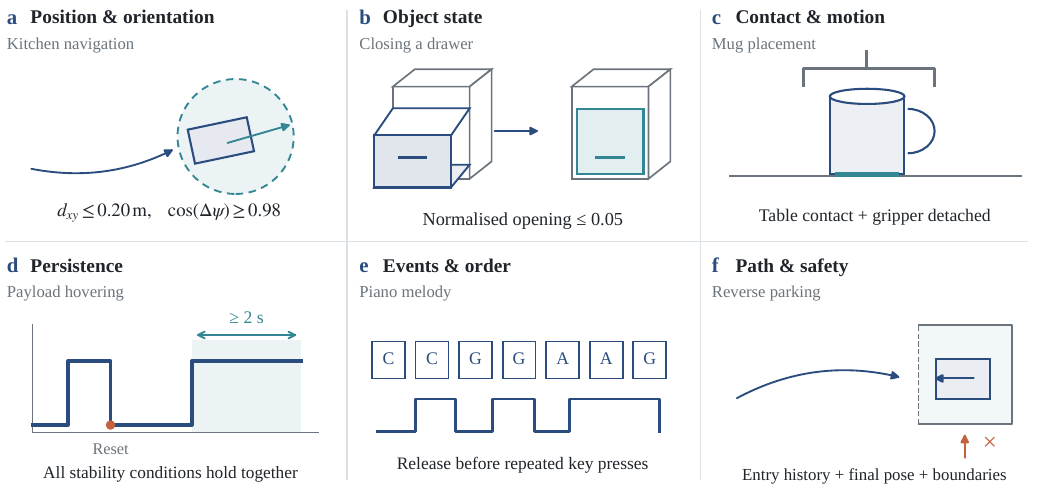}
\caption{\textbf{What establishes task completion.} Six families of success evidence, illustrated schematically. (a) Kitchen navigation checks position and heading. (b) Drawer closure checks normalised joint opening. (c) Mug placement requires support and release in addition to its spatial goal. (d) Payload hovering requires all stability bounds to hold together for at least the final two seconds; a broken continuous hold resets its duration. (e) Repeated piano notes require release before re-pressing. (f) Reverse parking combines entry history, final pose, and boundary constraints. Thresholds are specific to the illustrated tasks; these are checker components, not universal rules for all 84 tasks.}
\label{fig:success-criteria}
\end{figure}

\noindent\textbf{Position and orientation.} Geometric predicates test distances, alignment, insertion depth, containment, or relative placement. Kitchen navigation requires both the target position and heading. An apparent alignment in the image plane need not satisfy a three-dimensional geometric goal.

\noindent\textbf{Object state.} Articulation and semantic checks recognise conditions such as a closed drawer, an activated appliance, or frozen food. Object identities, counts, and quantified relations may also matter: placing exactly three qualifying fruits on a plate differs from placing at least three.

\noindent\textbf{Contact and motion.} Support, release, or velocity constraints distinguish completed placement from a transient or still-held configuration. The white-mug task, for example, checks table contact and gripper detachment as well as placement near the centre. These conditions are explicit when required; geometric containment alone does not certify physical contact.

\Needspace{3\baselineskip}
\noindent\textbf{Persistence.} Hold conditions require simultaneous satisfaction over a duration, whereas fixed-window conditions constrain an entire specified interval. A violation resets a continuous hold; a violation within a fixed evaluation window cannot be erased by subsequent recovery. These checks operate at the evaluator's sampling rate. An arrival goal receives no additional dwell requirement unless its checker specifies one.

\noindent\textbf{Events and order.} Counters and state machines recognise distinct ball hits, repeated recoveries, or ordered key presses. Repeated piano notes require a release event before the next press. Temporal ordering is imposed only when the goal requires it: a prescribed final stack order constrains spatial relations without necessarily prescribing the order of picks.

\Needspace{3\baselineskip}
\noindent\textbf{Path and safety.} History-dependent conditions check route checkpoints, entry direction, or forbidden boundary crossings. A valid final parking pose cannot compensate for a recorded path violation when the task requires reverse entry and boundary clearance.

These criteria connect an instruction to observable completion. A favourable final image, a finished tool call, and the agent's own success claim can each be insufficient. Which evidence is needed follows from the task definition, independently of the strategy used to attempt it.

\subsection{Validation and Coverage}
% Typography, rules and symbols match tables/related_evaluation_scope.tex.
% Compact row spacing and width keep the coverage matrix secondary to the text.
% Marks retain the audited representative-task labels from the coverage matrix.
\begin{table}[t]
\centering\small\color{black}
\begingroup
\DeclareRobustCommand{\covtick}{\rwcheck}
\DeclareRobustCommand{\covno}{\rwcross}
\captionsetup{width=.82\linewidth}
\caption{Representative task coverage by embodiment (\covtick: represented; \covno: no selected example). Labels overlap and describe task requirements, not model success.}
\label{tab:coverage}
\rwtable
\setlength{\tabcolsep}{6pt}
\renewcommand{\arraystretch}{1.12}
\resizebox{.82\linewidth}{!}{%
\begin{tabular}{@{}l*{5}{c}@{}}
\toprule
\rwhead{Embodiment} & \multicolumn{3}{c}{\rwhead{Goal requirements}} & \multicolumn{2}{c}{\rwhead{Dynamic control}} \\
\cmidrule(lr){2-4}\cmidrule(lr){5-6}
 & \rwhead{\shortstack{Spatial\\goals}} & \rwhead{\shortstack{Constrained\\contact}} & \rwhead{\shortstack{Multiple\\subgoals}} & \rwhead{\shortstack{Balance /\\tracking}} & \rwhead{\shortstack{Timed\\interaction}} \\
\midrule
Fixed-base arms & \covtick & \covtick & \covtick & \covtick & \covtick \\
Dexterous / bimanual & \covtick & \covtick & \covtick & \covtick & \covno \\
Mobile manipulators & \covtick & \covtick & \covtick & \covno & \covno \\
Humanoids / bipeds & \covno & \covtick & \covtick & \covtick & \covtick \\
Quadrupeds & \covno & \covno & \covno & \covtick & \covno \\
Wheeled / wheel-legged & \covtick & \covno & \covtick & \covtick & \covtick \\
Aerial robots & \covno & \covno & \covno & \covtick & \covtick \\
\bottomrule
\end{tabular}%
}
\endgroup
\end{table}

\noindent\textbf{Checker and execution evidence.} Validation distinguishes whether a checker implements its specification from whether the goal can be reached through legal actions. Positive, negative, and boundary fixtures test scoring logic, including missing evidence and failure precedence in the \bench{} state-checking framework. Scene probes, state and event replay, and trajectory or video inspection provide complementary evidence about physical execution. A constructed positive state can test a predicate but does not demonstrate a physically achievable solution.

\noindent\textbf{Feasibility and limits.} A successful action trajectory establishes reachability under its tested initial conditions and interface. Reference controllers may use additional state, with that access distinguished from agent observations, while acting through the declared control interface. Reference coverage is incomplete; checker validation does not establish a solution for every task or robustness across initial conditions. The scope of a success claim is further limited by the physical relations represented in its checker and the rate at which they are sampled.

\noindent\textbf{Coverage across embodiments.} Table~\ref{tab:coverage} maps representative requirements across seven embodiment groups. A marked cell indicates a selected example. Objectives are not systematically paired across bodies, so this coverage does not isolate morphology effects. Domain composition, overlapping requirements, and task checkers jointly define the scope of the results in Section~\ref{sec:evaluation}.
\FloatBarrier
\endgroup

% Shared colour and table treatment for the experimental results.
\definecolor{expink}{HTML}{161D2C}
\definecolor{expopus}{HTML}{D17A56}
\definecolor{expshared}{HTML}{246998}
\definecolor{expstripe}{HTML}{F4F6F8}
\definecolor{expheader}{HTML}{E8EDF2}
\definecolor{expmuted}{HTML}{66717D}

\section{Benchmarking Multimodal Agents on \bench{}}
\label{sec:evaluation}
We evaluate whether multimodal agents can complete physical tasks through the interfaces introduced in Section~\ref{sec:environment}. We first compare success across the full benchmark and its five task domains, then examine the time and interactions used by each model.

\subsection{Experimental Setup}
\noindent\textbf{Models and tasks.} We evaluate GPT-6 Astra, Claude Opus 5.5, Kimi K3, DeepSeek V4.1 Flash and Gemini 3.8 Flash on all 84 tasks. The experiments were collected from 30 September to 6 October 2026, with one episode per model--task pair, giving 420 scored episodes. We use the same task initialisation protocols, control-step horizons and scoring rules across models.

\noindent\textbf{Interaction settings.} Each agent receives the task goal, permitted observations and tool documentation. It issues direct robot-tool calls and replans from the returned feedback; environment-side code control is disabled. Auxiliary computation and image inspection follow the permissions of each environment. The observation history contains the current observation and up to four earlier observations sampled every two observation rounds. Agents cannot access hidden checker state or reference solutions. Appendix~\ref{app:execution} describes the available tools, and Appendix~\ref{app:models} gives representative system and developer instructions.

\subsection{Evaluation Protocol and Metrics}
\label{sec:evaluation_protocol}
\noindent\textbf{Episode budgets.} Each episode starts from the prescribed task state and ends when the task terminates or its budget is exhausted. Each task has a predefined control-step horizon. A separate non-action budget limits auxiliary calls, zero-step robot requests and otherwise action-free turns. On 83 tasks, the limits are 15 consecutive events and 30, 60 or 120 total events, depending on the task; volleyball 1v1 uses 20 and 150.\footnote{The task-specific total-budget tiers were calibrated from Astra interaction records and applied across models.} Executed control steps reset the consecutive counter, while API retries do not consume this budget. The numerical limits and accounting rules are detailed in Appendices~\ref{app:execution} and~\ref{app:results}.

\noindent\textbf{Task success.} Completion is determined by the executable checks described in Section~\ref{sec:qa}. We use the outcome at the applicable budget boundary, including retrospective adjudication when a recorded episode continued past that boundary. An unfinished task at the boundary counts as a failure. For a task set $\mathcal G$, the task-equal success rate of model $m$ is
\begin{equation}
\operatorname{SR}_{\mathcal G}(m)=\frac{1}{|\mathcal G|}\sum_{i\in\mathcal G}S_i(\xi_{i,m}),
\end{equation}
where $\xi_{i,m}$ is the evaluated episode and $S_i\in\{0,1\}$ is its task checker. Native rewards and partial scores are not averaged across tasks. Since each pair has one evaluated episode, the results describe this evaluation rather than variability over repeated trials.

\noindent\textbf{Resource use.} We measure elapsed wall-clock time, robot requests and all recorded tool calls. The last quantity includes robot requests, shell calls and explicit image-view calls; images already returned in robot feedback are not counted again. We report medians and interquartile ranges across tasks, including unsuccessful episodes. These measures describe the full recorded interaction, whereas task success is assessed at the evaluation boundary. Appendix~\ref{app:results} provides the numerical distributions and an additional breakdown of control steps and auxiliary calls.

\subsection{Main Results}
\label{sec:main-results}
\noindent\textbf{Task completion remains limited across all five models.} As shown in Figure~\ref{fig:front-leaderboard}, Astra completes 16 of 84 tasks (19.0\%), followed by Opus 5.5 with 13 (15.5\%), Kimi K3 with 2 (2.4\%), and DeepSeek V4.1 Flash and Gemini 3.8 Flash with 1 each (1.2\%). Astra leads Opus 5.5 by three tasks, or 3.6 percentage points, but even the highest-scoring model leaves 68 tasks unfinished. The low completion rates indicate that access to observations, robot tools and auxiliary computation does not yet translate into reliable execution across the benchmark's diverse embodiments and objectives.

\begin{figure}[!htbp]
\centering
\includegraphics[width=\linewidth]{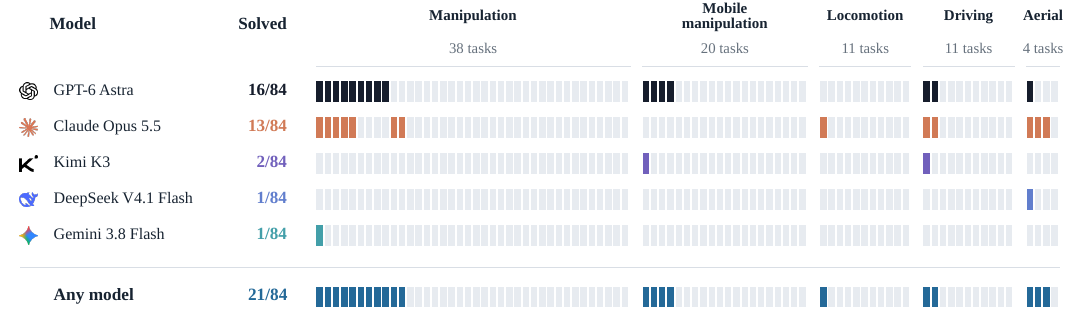}
\caption{\textbf{Success coverage across the 84 tasks.} Each aligned column represents one task, grouped by its primary domain. Coloured cells indicate success and grey cells indicate failure. Tasks are ordered identically for all models, with successful tasks grouped within each domain. The bottom row marks tasks solved by at least one model: 21 tasks are covered, leaving 63 unsolved. Each model has one evaluated episode per task; counts use the outcomes at the applicable evaluation boundary.}
\label{fig:experiment-coverage}
\end{figure}

\noindent\textbf{Different models solve partly different tasks.} Astra and Opus 5.5 share eight successes; Astra solves eight additional tasks and Opus 5.5 solves five. Their combined coverage is 21/84 tasks (25.0\%). As shown in Figure~\ref{fig:experiment-coverage}, the successes of Kimi, DeepSeek and Gemini fall within this set, leaving 63 tasks unsolved by any evaluated model. Table~\ref{tab:main-results} reports the five-domain results for every model. This union describes coverage by the evaluated models, rather than the performance of an agent that can select the best model in advance.

\begin{table}[!htbp]
\centering\small
\caption{\textbf{Main results across five task domains.} Entries give successful tasks / evaluated tasks; overall success rates are in parentheses. One episode is retained per model--task pair. Figure~\ref{fig:front-leaderboard} pools the first two domains for compact presentation.}
\label{tab:main-results}
\begingroup\renewcommand{\arraystretch}{1.22}\setlength{\tabcolsep}{5pt}
\begin{tabularx}{\linewidth}{@{}Xccccc c@{}}
\toprule
\rowcolor{expheader}
\textbf{Model} & \textbf{Manip.} & \shortstack{\textbf{Mobile}\\\textbf{manip.}} & \textbf{Locomotion} & \textbf{Driving} & \textbf{Aerial} & \textbf{Overall}\\
\midrule
Astra & \textbf{9/38} & \textbf{4/20} & 0/11 & \textbf{2/11} & 1/4 & \textbf{16/84 (19.0\%)}\\
\rowcolor{expstripe}
Opus 5.5 & 7/38 & 0/20 & \textbf{1/11} & \textbf{2/11} & \textbf{3/4} & 13/84 (15.5\%)\\
Kimi K3 & 0/38 & 1/20 & 0/11 & 1/11 & 0/4 & 2/84 (2.4\%)\\
\rowcolor{expstripe}
DeepSeek V4.1 Flash & 0/38 & 0/20 & 0/11 & 0/11 & 1/4 & 1/84 (1.2\%)\\
Gemini 3.8 Flash & 1/38 & 0/20 & 0/11 & 0/11 & 0/4 & 1/84 (1.2\%)\\
\bottomrule
\end{tabularx}\endgroup
\end{table}

The successful episodes cover a range of robot tasks. Astra completes tasks such as charger insertion, drawer closure and ordered block stacking, while Opus 5.5 succeeds at quadruped push recovery, payload hovering and drone juggling. Kimi succeeds at drawer closure and courtyard driving; Gemini succeeds at conveyor matching; DeepSeek succeeds at volleyball 1v1. Representative Astra and Opus 5.5 episodes are shown in Figure~\ref{fig:successful-trajectories}. Appendix~\ref{app:results} lists all 420 outcomes, and Appendix~\ref{app:trajectories} follows selected successes through observations, decisions and execution feedback.

\begin{figure}[!htbp]
\centering
\input{figures/success_trajectories}
\caption{\textbf{Representative successful trajectories of Astra (left) and Opus 5.5 (right).} Each strip contains five chronological video frames. The first two rows compare shared successes; the remaining rows show tasks solved only by the displayed model in the illustrated trajectory comparison. Sampling intervals vary, and final frames may precede recorded success.}
\label{fig:successful-trajectories}
\end{figure}

\subsection{Performance Across Task Domains}
\noindent\textbf{Separating mobile manipulation reveals a larger gap between the leading models.} As shown in Table~\ref{tab:main-results}, Astra solves 9/38 manipulation tasks (23.7\%) and Opus 5.5 solves 7/38 (18.4\%). In mobile manipulation, Astra completes 4/20 tasks (20.0\%), while Opus 5.5 completes none. Kimi K3 completes the drawer-closing task, giving 1/20 (5.0\%); DeepSeek and Gemini complete none. The combined panel in Figure~\ref{fig:front-leaderboard} therefore pools two distinct result profiles: 13/58 for Astra and 7/58 for Opus 5.5.

Within fixed-base manipulation, Astra and Opus 5.5 share five successes. Astra solves four additional tasks, and Opus 5.5 solves two, giving a union of 11/38. Astra's four mobile-manipulation successes bring the pair's coverage in the two manipulation domains to 15/58. Gemini's conveyor-matching success and Kimi's drawer-closing success do not expand that union. The aligned task columns in Figure~\ref{fig:experiment-coverage} expose these shared and model-specific successes without treating partially completed tasks as solved.

\noindent\textbf{Dynamic-control successes favour Opus 5.5 on the evaluated tasks.} Opus 5.5 completes 3/4 aerial tasks (75.0\%), compared with 1/4 each for Astra and DeepSeek. It also records the only locomotion-domain success, quadruped push recovery, for 1/11 (9.1\%). Its lower aggregate score therefore coexists with successes in domains where Astra is less successful. The aerial set is small: one task changes the rate by 25 percentage points. These are task-level observations from a single retained episode per pair, rather than estimates of repeated-run reliability.

\noindent\textbf{Driving and locomotion remain broadly unsolved.} Astra and Opus 5.5 each complete courtyard and hairpin driving, yielding 2/11 (18.2\%); Kimi completes courtyard driving, yielding 1/11 (9.1\%). No evaluated model completes the remaining nine driving tasks or ten of the eleven locomotion tasks. Across all five models, the domain unions are 11/38 manipulation, 4/20 mobile manipulation, 1/11 locomotion, 2/11 driving and 3/4 aerial. The 63 tasks outside these unions show that the benchmark's difficulty extends beyond a single embodiment or interface. Section~\ref{sec:analysis} examines the execution mechanisms behind selected successes and failures.

\subsection{Recorded Resource Consumption}
\label{sec:recorded-resources}
\begin{figure}[!htbp]
\centering
\includegraphics[width=\linewidth]{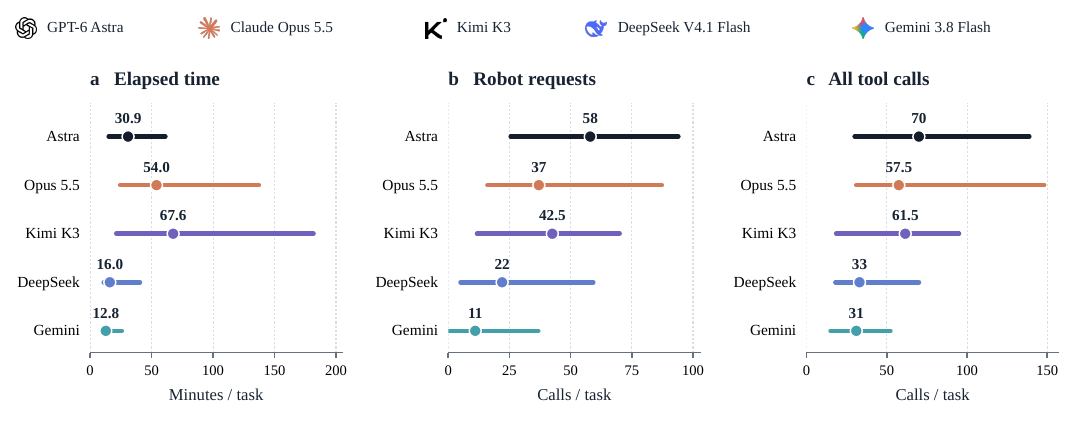}
\caption{\textbf{Recorded resource consumption.} Points and labels show medians; horizontal segments span the 25th--75th percentiles across 84 retained runs per model, including failed runs. Models use the same colours and row order in every panel. Elapsed time includes setup, waiting and execution. Total tool calls include robot requests, shell calls and explicit image-view calls. Full logs may extend beyond retrospective adjudication cutoffs; the intervals describe variation across tasks, rather than uncertainty over repeated trials.}
\label{fig:resource-consumption}
\end{figure}

\noindent\textbf{Longer episodes do not consistently correspond to more successful tasks.} As shown in Figure~\ref{fig:resource-consumption}, median elapsed times are 30.9 minutes for Astra, 54.0 for Opus 5.5, 67.6 for Kimi, 16.0 for DeepSeek and 12.8 for Gemini. Astra achieves the most successes with a shorter median episode than either Opus 5.5 or Kimi. Kimi has the longest median duration but completes only two tasks. These aggregate differences show that time spent interacting with the environment alone does not explain the success ranking.

\noindent\textbf{Interaction frequency and elapsed time give different orderings.} Median robot-request counts are 58, 37, 42.5, 22 and 11 for Astra, Opus 5.5, Kimi, DeepSeek and Gemini, respectively; median totals including auxiliary tools are 70, 57.5, 61.5, 33 and 31. Astra therefore makes more robot requests per median episode than Opus 5.5 or Kimi despite its shorter median elapsed time. Across tasks, durations also vary widely: the interquartile range is 15.2--61.2 minutes for Astra, 24.5--137.3 for Opus 5.5 and 21.3--181.7 for Kimi. Call counts alone do not capture these timing differences.

Elapsed time includes setup, waiting and execution, and full logs may extend beyond a retrospectively adjudicated cutoff. In addition, short episodes can reflect early failure or stopping. These measurements therefore describe resource consumption, rather than time to success or a ranking of efficiency. The matched-trajectory analyses in Section~\ref{sec:analysis} and Appendix~\ref{app:controlled} examine resource use for comparable task outcomes.

\subsection{Success, Token Consumption, and Estimated Cost}
\label{sec:cost-tradeoffs}
\begin{figure}[!htbp]
\centering
\includegraphics[width=\linewidth]{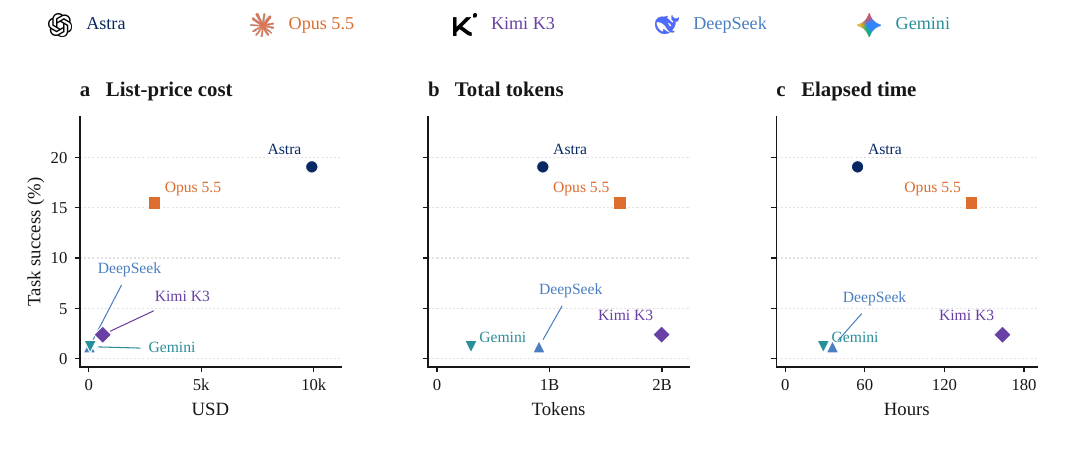}
\caption{\textbf{Success versus aggregate resource consumption.} Each point represents one model evaluated on 84 tasks; all panels use the same success axis. (a) Estimated list-price cost. (b) Total tokens, including cached input and output. (c) Cumulative elapsed time, including unsuccessful runs. All values follow the project website: valid per-task mean multiplied by 84. Token and cost estimates use 57 valid records for Kimi and 84 for each other model; all elapsed-time totals use 84 records. Missing usage is estimated by this scaling, rather than counted as zero. Resource accounting covers full recorded runs and may extend beyond retrospective scoring cutoffs; these plots do not measure cost or time to first success.}
\label{fig:cost-tradeoffs}
\end{figure}

We compare the overall success rate with three aggregate resource measures in Figure~\ref{fig:cost-tradeoffs}: estimated API cost at list prices, total tokens, and cumulative elapsed time. All three measures use the project website's aggregate accounting: the mean over scored tasks with a valid recorded value, multiplied by 84. Missing or zero-flagged token usage is excluded from the mean; Kimi's token and cost estimates therefore extrapolate from 57 valid records, while the other models have 84. Tokens include cached input and output, and costs are list-price estimates rather than invoices. All 84 elapsed durations are available for each model; these website-reported durations include simulation, API latency and waiting, rather than the duration of a parallel evaluation campaign. Appendix~\ref{app:results} gives the aggregate values and accounting scope.

\noindent\textbf{The strongest result has the highest reported cost, but not the largest token volume.} Astra reaches 19.0\% success at a reported \$9,913 and 941M tokens. Opus 5.5 reaches 15.5\% at \$2,916 and 1.63B tokens. Thus Astra's estimated expenditure is about 3.4 times Opus 5.5's, while its token volume is lower. Different model prices and cache accounting prevent raw token totals from being interpreted as a common monetary budget. Kimi K3 consumes the most reported tokens (2.00B), with \$624 estimated cost, but completes only two tasks. DeepSeek records one success at 907M tokens and \$34; Gemini records one at 303M tokens and approximately \$67.

\noindent\textbf{More aggregate computation or elapsed time does not ensure broader task coverage.} Cumulative elapsed times are 54.6 hours for Astra, 140.3 for Opus 5.5, 163.8 for Kimi, 35.7 for DeepSeek and 28.7 for Gemini. Astra completes the most tasks while using less elapsed time than Opus 5.5 or Kimi; Kimi consumes the most time and tokens without comparable success. These comparisons describe the evaluated model--harness combinations. They do not isolate inference speed, hardware, pricing effects or a causal benefit of extra computation. In particular, an early failed episode can consume few resources. The matched successful-task comparisons in Section~\ref{sec:analysis} therefore complement these aggregate plots by conditioning on shared completed tasks.

\FloatBarrier

\begingroup
\setlength{\textfloatsep}{11pt plus 2pt minus 2pt}
\setlength{\intextsep}{10pt plus 2pt minus 2pt}
\setlength{\parskip}{4pt plus 1pt minus 1pt}
\newenvironment{analysistakeaway}{\begin{tcolorbox}[enhanced,colback=kpale,colframe=kblue!65!black,
boxrule=.65pt,arc=5pt,left=7pt,right=7pt,top=8pt,bottom=5pt,
before skip=7pt,after skip=7pt,fontupper=\small,
attach boxed title to top left={xshift=8pt,yshift=-\tcboxedtitleheight/2},
boxed title style={colback=kblue!65!black,colframe=kblue!65!black,arc=2pt,boxrule=0pt},
title=Takeaway,fonttitle=\small\bfseries]}{\end{tcolorbox}}
\section{Analysis}
\label{sec:analysis}
The following case studies and cross-task profiles use the same retained five-model evaluation reported in Section~\ref{sec:evaluation}. We relate decisions to recorded execution and evaluator outcomes. Success uses the applicable evaluation boundary; resource distributions describe full retained logs, while explicitly identified milestone comparisons use matched progress within an episode. Paired Astra--Opus 5.5 cases examine different routes to comparable progress, and Kimi and DeepSeek cases expose failures in state tracking and continuation. These examples establish mechanisms in individual episodes, rather than their frequency across the benchmark.
\subsection{Feedback-Driven Adaptation and Recovery}
\label{sec:feedback-recovery}
Figure~\ref{fig:feedback-recovery} illustrates action refinement, reconfiguration after rejection, and sustained recovery. In K3's drawer-closing episode, reduced end-effector displacement prompted shorter pushing segments, followed by withdrawal to check closure. Native success occurred during withdrawal: the useful pattern was feedback-dependent refinement and verification.

\begin{figure}[!t]
  \centering
  \includegraphics[width=\linewidth]{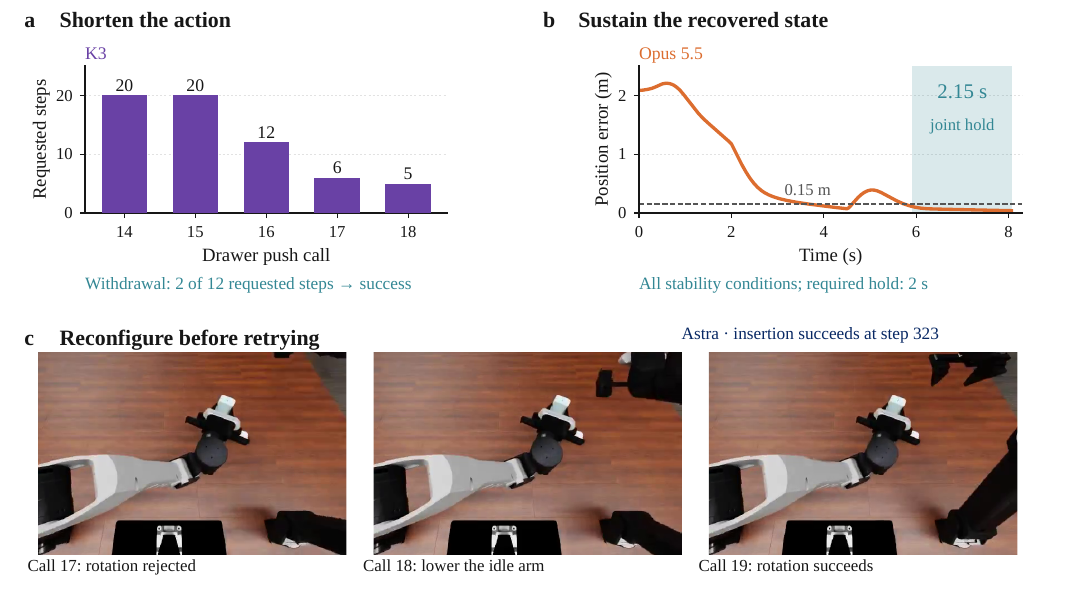}
\caption{\textbf{Feedback-dependent correction and its limits.} (a) Kimi K3 shortens successive drawer pushes; withdrawal executes two of twelve requested steps before success. (b) Opus 5.5 sustains payload hover: shading marks all stability conditions jointly satisfied for 2.15 seconds, exceeding the two-second requirement. (c) Astra changes the idle arm's position after a downward camera rotation is rejected at step 196: it lowers the arm by step 207 and successfully retries the rotation by step 230. The episode subsequently completes charger insertion at step 323. Frames show the corresponding states from the retained recording; the rejected call itself executes no steps.}
  \label{fig:feedback-recovery}
\end{figure}

Recovery can require changing the configuration from which an action is attempted. In charger insertion, Astra tried rotating the idle arm's wrist camera downward to inspect alignment. The planner rejected the command at step 196. Astra then lowered and brought the idle arm nearer before retrying the same requested pitch; this rotation executed successfully, as shown in Figure~\ref{fig:feedback-recovery}c. Further insertion adjustments, release and withdrawal produced native success at step 323. This sequence illustrates feedback-dependent reconfiguration, although it does not isolate that change as the cause of final success.

Dynamic recovery additionally requires sustaining the recovered state. Opus 5.5 combined offline linear-quadratic regulator calculations with repeated state feedback in drone payload hover. Payload error decreased, and the final interval met the joint position, velocity, swing, and attitude conditions, as shown in Figure~\ref{fig:feedback-recovery}b. Here, auxiliary computation supported successful execution through commands checked against subsequent observations.

\begin{analysistakeaway}
\textbf{Feedback matters when it changes execution.} Effective recovery adapts the command, revises a failing strategy, and verifies that the recovered state persists.
\end{analysistakeaway}

\subsection{Failure Modes}
\label{sec:failure-modes}
In K3's pouring episode, the initial approach knocked the bottle onto its side. The agent then reported that its first closure missed the fallen bottle. Repeated left-arm attempts did not establish a grasp; at step 240 it switched to the right arm, whose first repositioning ended at step 272. The remaining budget was consumed by further approach motions, ending without success at step 400, as shown in Figure~\ref{fig:failure-mechanisms}a. Reaching successive arm poses did not restore the object state needed for pouring.

\begin{figure}[!t]
  \centering
  \includegraphics[width=\linewidth]{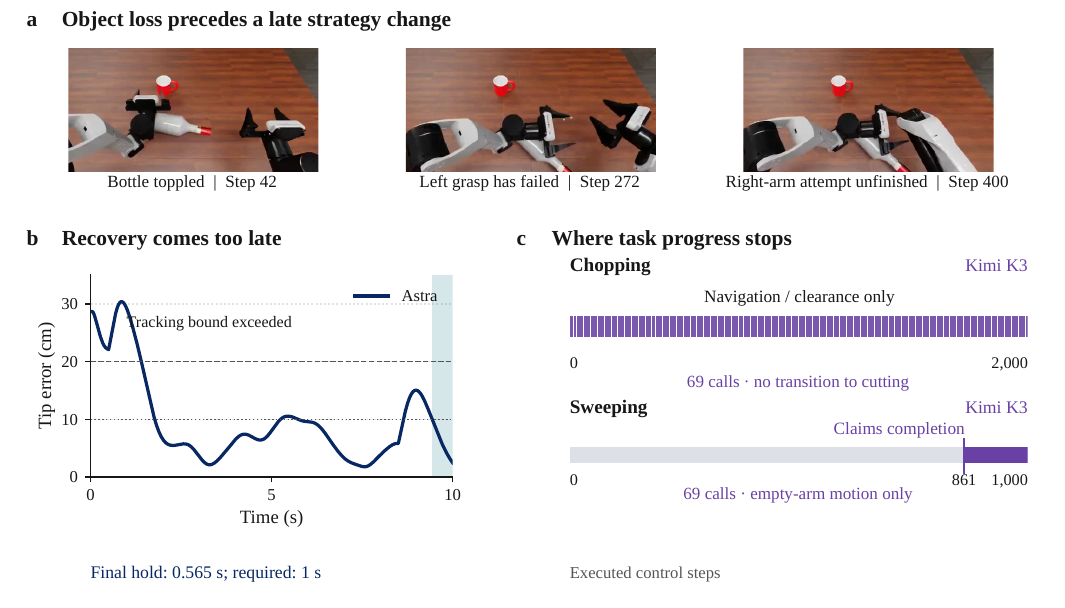}
\caption{\textbf{Execution without task completion.} (a) Kimi K3 topples the bottle during approach, fails to secure it with the left arm, and switches to a right-arm attempt late in the 400-step episode. Images are sampled from the current run at the labelled control steps. (b) Astra crosses the tracking bound (20\,cm) before recovering below the terminal bound (10\,cm); shading marks all terminal conditions, including speed and tilt. The final joint hold lasts only 0.565 seconds. (c) Chopping never advances to the next subgoal: 69 positive-step calls cover the entire 2,000-step episode. Sweeping ends with 69 calls that vary only the empty left arm's height over the final 139 steps, while the model asserts completion and the native task remains unsuccessful. Each bar spans its own episode horizon. White dividers mark chopping calls; the purple sweeping segment marks the final 139 steps. The chopping count covers the whole episode; the sweeping count covers only this final segment.}
  \label{fig:failure-mechanisms}
\end{figure}

Preparation can also consume the entire horizon. All 69 positive-step calls in K3's vegetable-chopping episode concerned surveying, navigation, or clearance around furniture. The final command still attempted kitchen entry; no cutting command appeared before the 2,000-step limit; see Figure~\ref{fig:failure-mechanisms}c. The observed bottleneck was reaching the workspace, so this failure does not diagnose the unattempted cutting skill.

Misjudging completion can similarly prevent further correction. In block sweeping, K3 described the task as complete after returning the empty left arm home at step 861. Its final 69 calls changed only that arm's vertical target, advancing another 139 control steps without resuming sweeping. Native success remained false at the 1,000-step horizon, as shown in Figure~\ref{fig:failure-mechanisms}c. This episode includes earlier rejected commands and revised targets, but the terminal failure is continued execution without corrective task action.

Recovery must also arrive in time. Astra's humanoid table-tennis episode ended at 3.58\,s when its base fell below 0.50\,m despite balance corrections. In inverted-pendulum tracking, Astra completed the horizon with only 2.45\,cm final tip error, but violated the tracking tolerance when checking began at 0.5\,s. Its final stable interval was 0.565\,s rather than the required second; see Figure~\ref{fig:failure-mechanisms}b. A favourable final frame can therefore conceal temporal failure.

\begin{analysistakeaway}
\textbf{Motion is not task progress.} Failures include losing the required object state, remaining in preparation, mistaking an unfinished state for completion, and recovering too late.
\end{analysistakeaway}

\subsection{Control Strategies Across Tasks}
\label{sec:control-strategies}
Figure~\ref{fig:control-strategies} compares sustained contact control in drone juggling; Figure~\ref{fig:stacking-strategies} shows the different orderings of perception, computation and action in stacking.

\begin{figure}[!htbp]
  \centering
  \includegraphics[width=\linewidth]{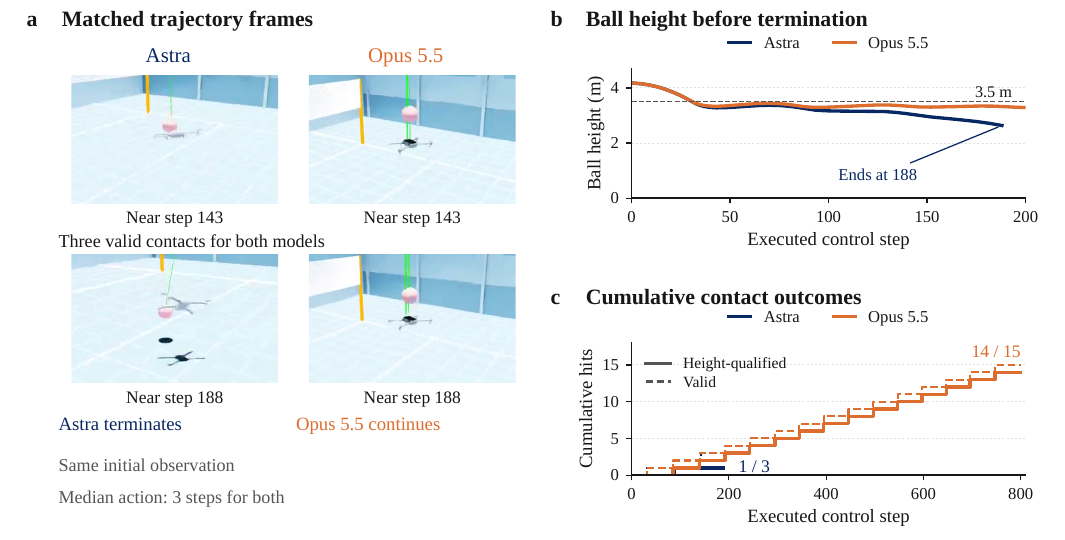}
\caption{\textbf{Sustaining contact under the same initial observation.} (a) Matched frames from Astra and Opus 5.5 near steps 143 and 188. Both have made three valid contacts near step 143; Astra terminates at 188 while Opus 5.5 continues. (b) Ball height over the first 200 steps; the dashed line marks the 3.5\,m qualification threshold. (c) Valid and height-qualified contact counts, ending at each trajectory's termination. The first contact is not height-qualified. Astra ends with 3 valid / 1 qualified contacts; Opus 5.5 reaches 15 / 14 at step 800. Both use a median of three executed steps per action. Frames are nearest samples from 12-fps recordings.}
  \label{fig:control-strategies}
\end{figure}

Both models performed offline dynamics calculations for drone juggling. Astra used position and attitude feedback with rotor thrust allocation. Opus 5.5's short rotor-action segments levelled the vehicle, accelerated before impact, reduced thrust after contact, and repositioned beneath the ball. Opus 5.5 sustained 15 valid hits, including 14 height-qualified hits, over 800 steps; Astra recorded three valid hits, one height-qualified, before ball-too-low termination at step 188. Both had a median of three executed steps per call. The distinction is sustained contact coordination, not simply building a model or using shorter segments.

Stacking contrasts direct visual estimation with programmatically assisted localisation. Astra approached the blue block from successive camera observations, closed the gripper at step 157, and lifted at 179 before external image-processing or camera-fitting computation. It introduced a camera-projection fit only at step 204, after carrying blue towards red. Opus 5.5 wrote colour-segmentation code before moving, then sampled robot motions to fit camera geometry and compute targets.

\begin{figure}[!htbp]
\centering
\includegraphics[width=\linewidth]{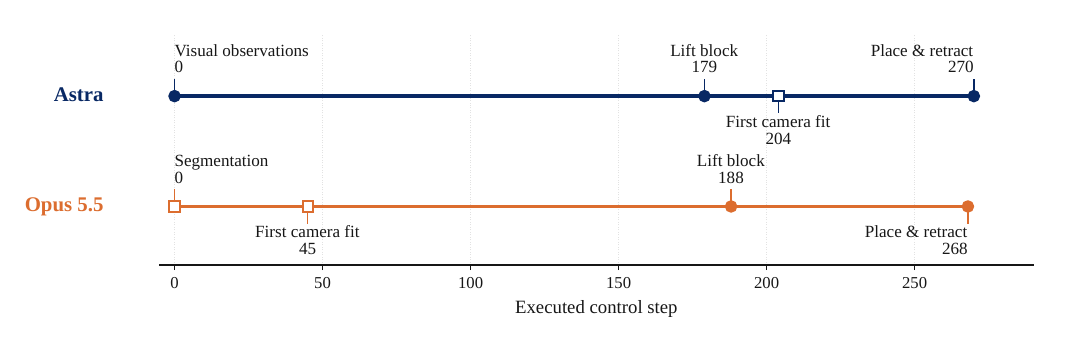}
\caption{\textbf{Different routes from observation to grasping.} Milestones from the two stacking episodes on a shared control-step axis. Filled circles mark observations or physical milestones; open squares mark explicit computation, during which physics is paused. Astra closes the gripper at step 157 and lifts at 179, before its first camera fit at 204. Opus 5.5 writes colour-segmentation code before motion, samples the arm and fits camera geometry at step 45, then lifts at 188. Both reach first placement and retraction at similar physical steps (270 versus 268), but their information-gathering sequences differ. Astra completes the stack at step 635; Opus 5.5 reaches the applicable auxiliary-call budget at step 313. These observations describe strategies in the illustrated episodes and do not identify the models' training data.}
\label{fig:stacking-strategies}
\end{figure}

Astra reused the grasp geometry and stack centre for later blocks, checking the resulting state through lifts and retractions. Opus 5.5 repeatedly refined segmentation and geometric estimates. Astra finished in 635 control steps; Opus 5.5 reached its auxiliary budget at step 313 without completion. Section~\ref{sec:interaction-efficiency} compares the interaction costs of their shared first-block milestone.

\begin{analysistakeaway}
\textbf{Useful estimates must become sustained control.} Both models build numerical controllers; their trajectories differ in coordinating repeated contacts and in obtaining and reusing spatial estimates.
\end{analysistakeaway}

\subsection{Completion Assessment and Decisions to Continue}
\label{sec:completion-assessment}
Table~\ref{tab:completion-assessment} distinguishes these judgements using stated assessments and subsequent actions. The relevant distinction is whether the next command tests or repairs an unresolved task condition: execution can continue even after corrective search has stopped.

\begin{table}[!htbp]
\centering\small
\caption{\textbf{Completion judgements and subsequent execution.} Holding and return-to-home commands advance simulation even when they do not advance the unfinished task. In white-mug placement, the tool closes without a final observation or success score; the release request may itself trigger success and does not establish post-success over-action.}
\label{tab:completion-assessment}
\begingroup
\setlength{\tabcolsep}{5pt}
\renewcommand{\arraystretch}{1.22}
\begin{tabularx}{\linewidth}{@{}*{4}{>{\raggedright\arraybackslash}X}@{}}
\toprule
\rowcolor{expheader}
\textbf{Case / model} & \textbf{Agent judgement} & \textbf{Subsequent behaviour} & \textbf{Recorded result} \\
\midrule
Block sweeping\newline Kimi K3 & Declares completion at step 861 & 69 empty-arm height commands; 139 more steps & Unfinished at the 1,000-step horizon \\
\addlinespace[5pt]
\rowcolor{expstripe}
Stove navigation\newline DeepSeek & No further motion needed at step 345 & 105 more steps with zero base-velocity commands & Unfinished at the 450-step horizon \\
\addlinespace[5pt]
Hanging mugs\newline Opus 5.5 & 35 steps deemed insufficient to grasp and hang a mug & Returns the arm home over 35 steps & Unfinished at the 800-step horizon \\
\addlinespace[5pt]
\rowcolor{expstripe}
White mug\newline Opus 5.5 & Requests gripper release at step 340 & One of 15 requested steps executes & Native success at step 341 \\
\bottomrule
\end{tabularx}
\endgroup
\end{table}

In DeepSeek's stove-navigation episode, visibility, proximity, and apparent counter contact supported its claim of completion. A continuation prompt elicited repositioning, but the model then judged further motion unwarranted and issued holding commands. The final 105 steps advanced simulation using zero base-velocity commands, while native success remained false. Continuation restored tool use without restoring corrective search. Kimi's sweeping episode exhibits the same distinction through nonzero motion: the empty arm changes height, but those commands do not revisit the unfinished sweeping objective, as shown in Figure~\ref{fig:failure-mechanisms}c.

Opus 5.5 instead recognised failure during mug hanging: opening the gripper dropped the yellow mug. It first returned home, then made another approach with 83 steps remaining. At step 765, it judged the final 35 steps insufficient to grasp and hang a mug and spent them returning the arm home. This was a judgement of infeasibility, not a claim of success; the trace does not establish that recovery was impossible.

Successful termination leaves a different ambiguity. Opus 5.5 requested a 15-step release after observing step 340 in white-mug placement; success occurred at step 341, followed by closure without a final observation or score. The release may itself have triggered success, so the plan does not establish unnecessary post-success action. In this placement episode, native success ends execution, whereas sustained-stability tasks require continued control beyond an instantaneously favourable state. A continuation rule must therefore distinguish the agent's own assessment from the task's actual termination and temporal requirements.

\begin{analysistakeaway}
\textbf{Continuation requires judging both completion and recoverability.} An agent can stop correcting because it believes the goal is met, or because it believes further progress is infeasible.
\end{analysistakeaway}

\subsection{Interaction Efficiency}
\label{sec:interaction-efficiency}
Figure~\ref{fig:interaction-efficiency} separates robot, shell, and explicit image-view calls from executed control steps for matched scenes and seeds. Images embedded in robot feedback and accompanying prose add no separate call. These are execution counts, independent of API latency.

\begin{figure}[!htbp]
\centering
\includegraphics[width=\linewidth]{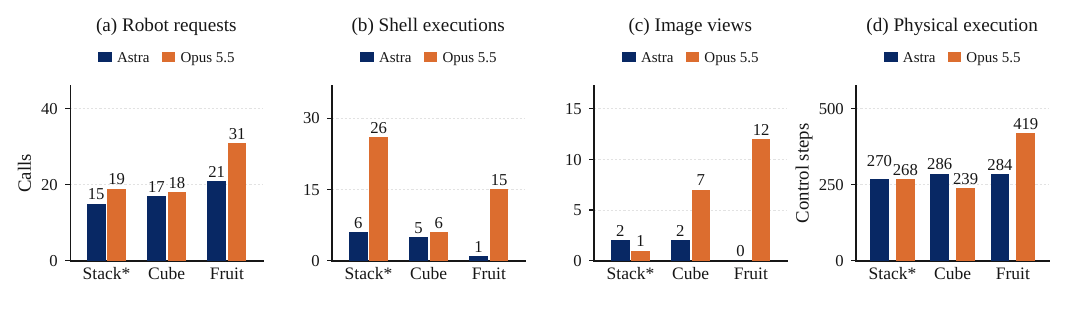}
\caption{\textbf{Tool calls and physical execution measure different costs.} Stacking ends at the shared first blue-on-red placement and retraction, before Opus 5.5's step-313 budget cutoff. Cube and fruit comparisons cover full successful episodes. Separate panels count robot requests, shell executions, explicit image-view calls and control steps. Total tool calls (Astra / Opus 5.5) are 23 / 46 for stacking, 24 / 31 for cube and 22 / 58 for fruit. Images embedded in robot feedback are not additional image-view calls.}
\label{fig:interaction-efficiency}
\end{figure}

For first-block placement and retraction, Astra used 23 tool calls versus Opus 5.5's 46, despite nearly identical physical execution (270 versus 268 steps). Before lifting blue, Astra used ten robot calls and two shell calls; Opus 5.5 used 13 robot calls and 25 auxiliary calls. The localisation strategies in Section~\ref{sec:control-strategies} thus reached comparable progress at different interaction costs, including analysis and software preparation.

Cube placement reverses the physical-cost ordering. Astra's initial grasp slipped, requiring realignment and a deeper grasp. Opus 5.5 performed additional localisation and contact checks without the same drop-and-regrasp sequence. Both succeeded: Astra used fewer tool calls (24 versus 31), but more control steps (286 versus 239).

In fruit arrangement, both costs increased together: Astra used 22 tool calls and 284 steps; Opus 5.5 used 58 calls and 419 steps, including extra contact-relief motions, gripper checks, and placement adjustments. Since the models chose different lemons first, the comparison uses full successful episodes. Reporting both costs reveals how information gathering and physical recovery contribute to reaching a matched outcome.

\begin{analysistakeaway}
\textbf{Interaction cost and physical cost can diverge.} Comparable progress may require twice as many tool calls; fewer tool calls need not mean fewer control steps.
\end{analysistakeaway}

\subsection{Cross-Task Model Profiles}
\label{sec:model-profiles}
Across 84 matched task records, Figure~\ref{fig:model-profile-interactions} compares interaction distributions and Figure~\ref{fig:model-profile-demands} groups outcomes by task requirements.

\begin{figure}[!htbp]
\centering
\includegraphics[width=\linewidth]{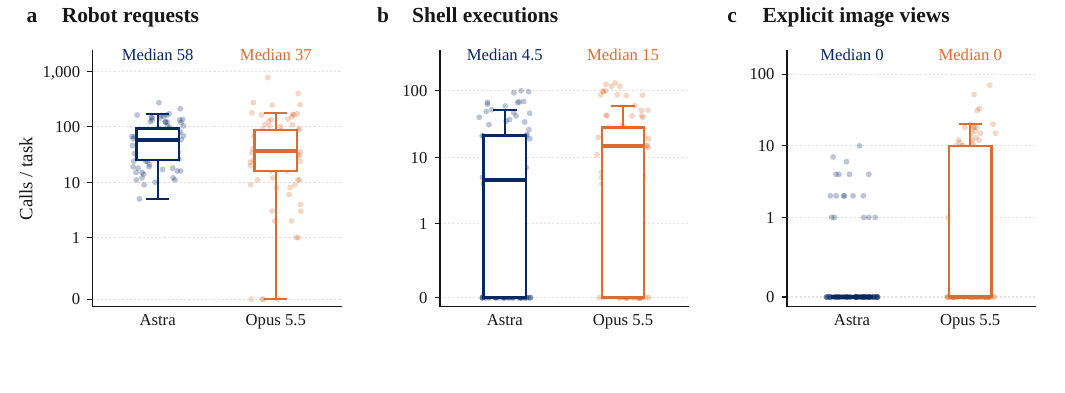}
\caption{\textbf{Interaction distributions across 84 tasks.} Points represent the full retained episode for each task, including failures; boxes show medians and interquartile ranges, with whiskers extending to the most extreme points within 1.5 interquartile ranges. Robot requests include rejected and zero-step requests; shell counts exclude polling. Explicit image views exclude images embedded in feedback. Astra has zero explicit views on 66/84 tasks and Opus 5.5 on 43/84, so both medians are zero; their upper quartiles are 0 and 10 calls, respectively. Symmetric-log axes retain zeros. These distributions describe tool use, not efficiency or success-conditioned cost.}
\label{fig:model-profile-interactions}
\end{figure}

In the full retained logs, Opus 5.5 makes 2,095 shell calls and 568 explicit image-view calls, compared with Astra's 1,435 and 56. Median shell counts are 15 versus 4.5 per task. Thus, external computation and explicit inspection are more common in Opus 5.5's recorded interactions. Counts alone do not reveal how useful those calls were: Figure~\ref{fig:stacking-strategies} supplies the missing ordering and purpose for one matched task.

Among eight tasks solved by both models, Astra uses fewer tool calls in seven and fewer control steps in six, as shown in Figure~\ref{fig:shared-success-costs}. Conveyor grasping reverses both orderings; cube placement reverses the step ordering. This extends Section~\ref{sec:interaction-efficiency} beyond individual examples. Across all tasks, call totals also reflect the duration of unsuccessful attempts and are not an efficiency ranking.

\begin{figure}[!htbp]
\centering
\includegraphics[width=\linewidth]{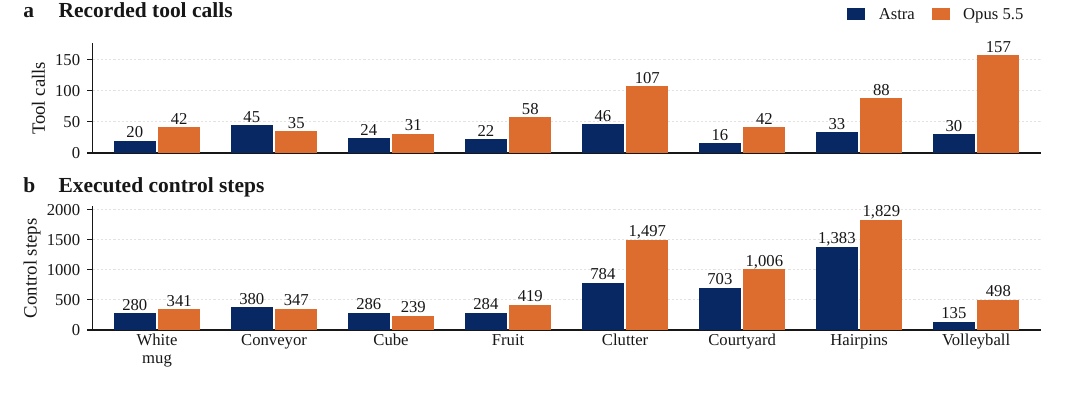}
\caption{\textbf{Resource use on the eight tasks solved by both models.} Absolute tool-call and control-step counts cover each full retained successful episode. Tool calls include robot requests, shell executions and explicit image views; control steps use the final episode count. Astra uses fewer tool calls on seven tasks and fewer control steps on six. These paired costs condition on shared success; the all-task distributions in Figure~\ref{fig:model-profile-interactions} include unsuccessful runs.}
\label{fig:shared-success-costs}
\end{figure}

Five overlapping goal labels cover all tasks: spatial placement/navigation, constrained contact, continuous stabilisation, timed interaction, and multiple subgoals. Ordinary approach--grasp--lift alone does not qualify as multiple subgoals.

\begin{figure}[!htbp]
\centering
\includegraphics[width=.94\linewidth]{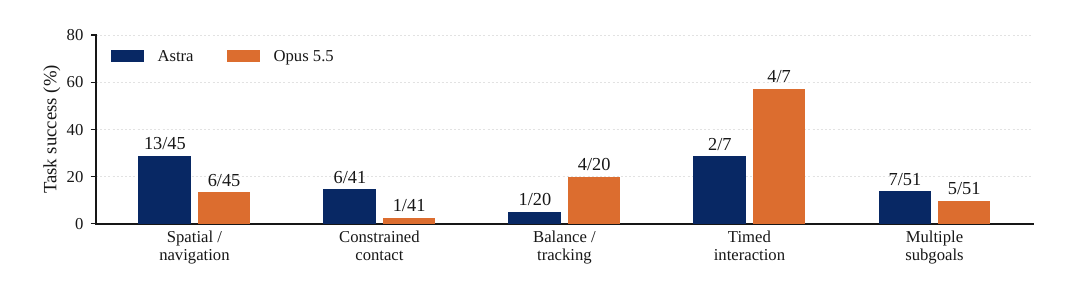}
\caption{\textbf{Exploratory task-demand profiles.} Bars show success rates with exact counts. Labels overlap and both models are evaluated on the same tasks within each group. All 84 tasks are scored for both models. Group membership is assigned from task requirements, not inferred from the outcome; these descriptive differences do not isolate component abilities.}
\label{fig:model-profile-demands}
\end{figure}

Astra succeeds on 13/45 spatial tasks versus Opus 5.5's 6/45, and on 6/41 constrained-contact tasks versus 1/41. Opus 5.5 leads on continuous balance/tracking (4/20 versus 1/20) and timed interaction (4/7 versus 2/7), including quadruped push recovery and cup-based ball catching as well as drones. Overlapping groups describe related views of the same task set.

\begin{analysistakeaway}
\textbf{The model profiles differ across task demands.} Astra uses fewer tool calls on seven of eight jointly solved tasks and succeeds more often on spatial/contact goals; Opus 5.5 succeeds more often on continuous and timed dynamic goals.
\end{analysistakeaway}

We hypothesise that these profiles reflect different ways of composing perception, computation, and control. Astra may translate visual observations into spatial estimates and actions more effectively, reducing external inspection during placement and alignment. Opus 5.5 more often expresses geometric or physical relationships through programs, which may aid prediction and timing in some dynamic tasks. Both routes occur in both models: Astra also builds numerical controllers, and successful Opus 5.5 control need not use many shell calls. Robot use may therefore draw on the transfer and composition of visual-spatial reasoning, program construction, and closed-loop agent execution. This interpretation remains an exploratory hypothesis rather than an isolated component measurement. The comparisons do not identify training-data differences, and neither tool volume nor a single successful strategy establishes a general capability advantage.

Taken together, the cases identify three concrete requirements for reliable robot use: preserve the object and subgoal state across motions, turn failure feedback into a changed task-directed action, and maintain the required conditions for the full evaluation interval. Figures~\ref{fig:feedback-recovery} and~\ref{fig:failure-mechanisms} show why local pose accuracy alone is insufficient; Figures~\ref{fig:stacking-strategies} and~\ref{fig:shared-success-costs} show why useful computation must also be assessed against physical progress and matched outcomes.

\FloatBarrier
\endgroup

\section{Discussion and Limitations}
Our results suggest that progress in robot use depends on how agents combine capabilities throughout an episode. Visual estimation, program construction, and dynamics reasoning can each support useful actions, yet reliable completion requires the agent to track their consequences, revise a failing strategy, and assess progress towards the goal. This motivates training and agent designs that strengthen the connection between perception, computation, and feedback-driven execution. The contrasting model profiles offer hypotheses about where these connections differ; they do not isolate the contribution of vision, coding, or control. Targeted comparisons that vary access to visual observations, auxiliary computation, and controller support could test these hypotheses and identify which interventions improve task completion.

The transition from simulation to physical deployment remains an open test. Our results measure performance under the observations and control interfaces available in each task. These interfaces provide different degrees of assistance, and pausing simulation during model inference removes an important real-world timing constraint. Physical robots must act despite inference delays, sensor noise, and discrepancies in contact and dynamics. Evaluating the same agent strategies in real-time loops and on hardware would establish which capabilities survive these demands and where faster feedback or additional control support is needed.

The suite spans diverse tasks and embodiments, but its coverage is finite and uneven, with related tasks inherited from shared source projects. Results therefore characterise performance on this suite rather than a representative distribution of all robot use. Public task descriptions, code, or demonstrations may also have appeared in model training data. Extending evaluation to newly constructed tasks, varied initial conditions, and additional embodiments would test whether the observed strategies transfer beyond familiar settings and help distinguish reusable physical competence from task-specific success.

\section{Conclusion}
\bench{} provides a challenging, systematic proving ground for agents moving from digital tasks towards robot use. Across diverse tasks and embodiments in simulation, our analysis reveals both the scope of emerging capabilities and the difficulty of composing them into reliable physical action. Agents can construct perception and control workflows, estimate spatial relationships, reason about dynamics, and revise actions from feedback. Yet successful execution also requires preserving task-relevant states, correcting ineffective actions, meeting temporal constraints, and recognising when a goal has actually been achieved. Differences across models further show that strengths in one class of physical demands do not imply uniformly strong robot use. These findings motivate progress in how perception, computation, and action work together throughout an episode. By connecting measurable outcomes with observable execution patterns, \bench{} gives the community a basis for identifying training and agent-design priorities and testing whether proposed improvements lead to more reliable robot behaviour. We hope this shared proving ground helps turn the promise of physical-world agents into cumulative, measurable progress.

\clearpage
\bibliographystyle{assets/plainnat}
\bibliography{references}
\clearpage
\appendix
% Appendix-only, editable task cards and trace excerpts.
\urlstyle{same}
\definecolor{appnavy}{HTML}{243C55}
\definecolor{appblue}{HTML}{EAF2F8}
\definecolor{appgreen}{HTML}{EAF4EE}
\definecolor{apporange}{HTML}{FCF0E8}
\definecolor{appgray}{HTML}{F1F3F5}
\definecolor{appline}{HTML}{B8C3CD}
\newtcolorbox{appcard}{enhanced,colback=white,colframe=appline,
 boxrule=.5pt,arc=0pt,left=9pt,right=9pt,top=8pt,bottom=8pt,
 before skip=5pt,after skip=5pt}
\newtcolorbox{appblock}[2]{enhanced,colback=#1,colframe=#1,
 boxrule=0pt,arc=0pt,left=7pt,right=7pt,top=5pt,bottom=5pt,
 title={#2},colbacktitle=#1,coltitle=appnavy,
 fonttitle=\sffamily\bfseries\fontsize{9.3}{11}\selectfont,
 toptitle=4pt,bottomtitle=1pt,before skip=5pt,after skip=5pt}
\newtcolorbox{appexcerpt}[1]{enhanced,breakable,colback=appgray,
 colframe=appline,boxrule=.35pt,arc=0pt,left=7pt,right=7pt,
 top=5pt,bottom=5pt,colbacktitle=appnavy,coltitle=white,
 title={#1},fonttitle=\sffamily\bfseries\fontsize{9}{10.5}\selectfont,
 before skip=5pt,after skip=7pt}
\newcommand{\appsmall}{\fontsize{9.4}{11.3}\selectfont\setlength{\parskip}{4pt}}
\newcommand{\appcode}[1]{{\small\texttt{\detokenize{#1}}}}
\newcommand{\appstage}[2]{\par\noindent\textbf{Stage #1.} #2\par}
\newcommand{\appverdict}[2]{\noindent\textbf{#1}\quad #2\par}
\newcommand{\appfact}[2]{\par\noindent
 \begin{minipage}[t]{.18\linewidth}\raggedright\color{appnavy}\sffamily\bfseries\fontsize{8}{9.6}\selectfont\mbox{#1}\end{minipage}\hfill
 \begin{minipage}[t]{.79\linewidth}\raggedright #2\end{minipage}\par\vspace{9pt}}
\newtcolorbox{appprompt}[2]{enhanced,breakable,colback=appgray,
 colframe=white,boxrule=0pt,arc=0pt,borderline west={2pt}{0pt}{#1},
 left=9pt,right=8pt,top=5pt,bottom=6pt,
 title={#2},colbacktitle=appgray,coltitle=#1,
 fonttitle=\sffamily\bfseries\fontsize{9.3}{11}\selectfont,
 toptitle=5pt,bottomtitle=1pt,before skip=6pt,after skip=8pt}

\section{Benchmark Specification and Task Profiles}
\label{app:spec}
\subsection{Task inventory and reading guide}
This appendix describes the retained evaluation snapshot dated 6 October 2026: 84 tasks from 20 source groups, with five scored model records per task. VolleyBots 1v1 is included. Source groups record software provenance; they are not counts of distinct robot embodiments. The five primary domains contain 38 manipulation tasks, 20 mobile manipulation tasks, 11 locomotion tasks, 11 driving tasks, and 4 aerial tasks. Mobile manipulation comprises the ten RoboCasa and ten BEHAVIOR-1K tasks; their mobile base and arm interfaces distinguish them from fixed-base manipulation. Locomotion includes legged motion, whole-body balance and body-coordinated interaction.

The eight profiles below connect each task objective to what the model can sense, what its tools command, and what the evaluator checks. Each profile pairs actual recorded views with a concise interface specification. The complete task inventory appears in Appendix~\ref{app:results}; the registered tool catalog and original prompt examples appear in Appendices~\ref{app:execution} and~\ref{app:models}. Task objectives here are condensed summaries; quoted prompt text is identified separately.
The inventory credits each originating project. Paper citations identify the corresponding published work or preprint; software citations identify projects for which no separate project paper was found. Digit uses the Isaac Lab environment, TTRL uses the PACE project, and the VolleyBots source includes the JuggleRL task.
\par\begingroup\small\rwtablemono\renewcommand{\arraystretch}{1.12}\begin{longtable}{@{}p{.25\linewidth}r p{.62\linewidth}@{}}
\toprule\textbf{Task source} & \textbf{Tasks} & \textbf{Reference}\\\midrule\endhead
HumanoidSoccer & 1 & \citep{humanoidsoccer} \\
ReflexBench & 1 & \citep{reflexbench} \\
Digit & 1 & \citep{isaaclab} \\
Flamingo & 1 & \citep{flamingo}; software \\
Go2 Push & 1 & \citep{go2locomotion}; software \\
AI-CPS & 3 & \citep{aicps} \\
OmniIsaacGymEnvs & 1 & \citep{omniisaacgymenvs}; software \\
OmniDrones & 2 & \citep{omnidrones} \\
Bench2Dex & 9 & \citep{bench2dex} \\
RoboCasa & 10 & \citep{robocasa2024,robocasa365} \\
RoboLab & 10 & \citep{robolab} \\
BEHAVIOR-1K & 10 & \citep{behavior1k} \\
WheeledLab & 11 & \citep{wheeledlab} \\
Robot Lab & 1 & \citep{robotlab}; software \\
SteadyTray & 1 & \citep{steadytray} \\
TTRL & 1 & \citep{pace} \\
VolleyBots & 2 & \citep{volleybots,jugglerl} \\
Wheel-Legged & 2 & \citep{wheelleggedlab}; software \\
Wheeled Quadruped & 1 & \citep{wheeledquadruped}; software \\
RoboDojo & 15 & \citep{robodojo} \\
\bottomrule\end{longtable}\endgroup
\clearpage\subsection{Primary task types}
The primary types below form an editorial partition based on the task objective. A grasp, contact event, or intermediate motion can occur in several types; these counts are task counts, not frequencies of atomic actions in trajectories. Secondary demands and success-check families can overlap.
\par\begingroup\fontsize{8.8}{10.5}\selectfont\rwtablemono\renewcommand{\arraystretch}{1.05}
\begin{longtable}{p{.27\linewidth}p{.52\linewidth}r}
\toprule\textbf{Domain} & \textbf{Primary task type} & \textbf{Tasks}\\\midrule\endhead
Manipulation & Fitting \& insertion & 7 \\
Manipulation & Placement \& organisation & 13 \\
Manipulation & Pouring \& processing & 3 \\
Manipulation & Multi-stage workflows & 8 \\
Manipulation & Articulation \& device use & 3 \\
Manipulation & Dynamic manipulation & 4 \\
Mobile manipulation & Navigation \& workflows & 8 \\
Mobile manipulation & Pouring \& processing & 2 \\
Mobile manipulation & Placement \& organisation & 8 \\
Mobile manipulation & Articulation \& device use & 2 \\
Locomotion & Carrying \& ball interaction & 3 \\
Locomotion & Locomotion \& terrain & 4 \\
Locomotion & Balance \& recovery & 4 \\
Driving & Drift \& timed passage & 4 \\
Driving & Routes \& terrain & 5 \\
Driving & Precision parking & 2 \\
Aerial & Stabilisation \& tracking & 2 \\
Aerial & Ball interaction & 2 \\
\bottomrule\end{longtable}\endgroup

\begingroup\appsmall
\noindent\textbf{Manipulation (38 tasks).} Fixed-base arms and dexterous hands establish object relations or operate objects within a workspace. Placement and organisation includes sorting, stacking and reorientation. Fitting and insertion includes chargers, tubes and supported placements such as hanging a mug. Pouring and processing is distinguished by the material-transfer objective. A multi-stage workflow combines distinct goals, such as storing multiple items or using a device; routine approach, grasp and lift motions alone do not create this category. Dynamic manipulation involves a moving target or sustained object regulation.

\noindent\textbf{Mobile manipulation (20 tasks).} The ten RoboCasa and ten BEHAVIOR-1K tasks expose a mobile base together with manipulation tools. The domain describes the available embodiment and control problem; it does not require every successful episode to move the base. For example, drawer closure remains a mobile-manipulation task even when the arm performs the decisive final motion. Primary task types still describe the objective: placement, articulation, processing or navigation and household workflows. This separates mobility from the object operation being evaluated.

\noindent\textbf{Locomotion (11 tasks).} This domain includes legged movement, terrain traversal, posture recovery and whole-body coordination. Balance and recovery tasks emphasise the body state rather than distance travelled. Carrying and ball interaction covers tasks where motion of the body must be coordinated with a load or a ball, such as tray transport and humanoid ball interaction. Wheel-legged platforms remain in this domain when their task tests posture or legged mobility rather than vehicle steering along a route.

\noindent\textbf{Driving (11 tasks).} Vehicle tasks use speed and steering to traverse routes, negotiate terrain, drift or park. Drift and timed passage is separated from ordinary route following because a transient manoeuvre or moving constraint is central to the objective. Precision parking instead emphasises the terminal pose and alignment. A driving trajectory can contain all three behaviours while still contributing only once to the inventory under its primary objective.

\noindent\textbf{Aerial (4 tasks).} Two tasks test payload stabilisation or inverted-pendulum tracking, and two test drone--ball interaction. The former require sustained regulation of multiple state variables; the latter also depend on contact histories and timing. Across all five domains, the finer task types are descriptive categories, while the executable checker remains the authority for success. Secondary requirements such as constrained contact or multiple subgoals can overlap and are analysed separately in Section~\ref{sec:analysis}.

\par\endgroup

\clearpage
\subsection{Task profile 01: Ordered block stacking}
\label{app:card-astra-62}
\par\begingroup\appsmall
{\color{appnavy}\sffamily\bfseries RoboLab \quad / \quad Placement and organisation}\par\smallskip{}Fixed-base DROID: Franka arm and Robotiq gripper\par\medskip
\noindent\textbf{Objective.} Stack the blocks from bottom to top in the order red, blue, green, yellow.\par\medskip
\begin{center}
\begin{minipage}[b]{.485\linewidth}\centering\includegraphics[width=\linewidth,height=135pt,keepaspectratio]{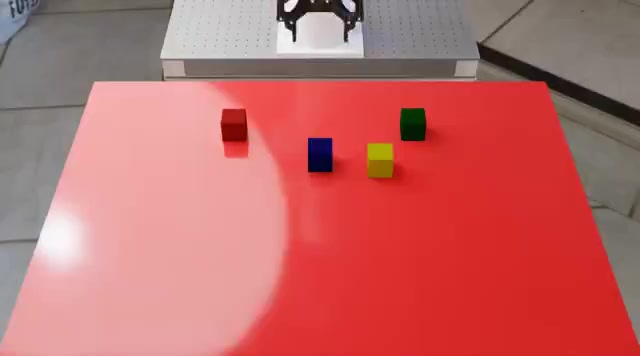}\par{\scriptsize Video 0.10 s}\end{minipage}\hfill
\begin{minipage}[b]{.485\linewidth}\centering\includegraphics[width=\linewidth,height=135pt,keepaspectratio]{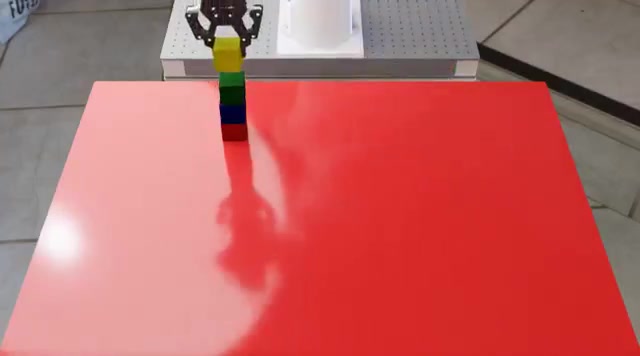}\par{\scriptsize Video 40.98 s}\end{minipage}
\end{center}{\small Astra; external review camera. The late frame precedes the terminal gripper request.}\par\medskip
\appfact{01 / SENSE}{RGB camera views and measured arm, gripper and end-effector state. Absolute IK targets use robot-root metres and WXYZ orientation. The commanded point is the gripper flange, not the fingertips.}
\appfact{02 / ACT}{\tool{move\_eef}, \tool{set\_gripper}, and coordinated robot commands. Ordinary segments request 1--30 steps at 15 Hz. Opening is \tool{gripper\_close=0}; closing is 1.}
\appfact{03 / VERIFY}{The native stacked relation checks the complete red--blue--green--yellow order. The audited source uses a default 1 cm relation tolerance. An executed approach or a visually plausible partial stack does not establish completion.}
\appfact{EPISODE}{1,350 control steps (90 simulated seconds). This retained run ends successfully at result step 635; the last completed event is step 634.}
\appfact{04 / EXECUTION}{The ordered tower requires repeated object localisation, grasping, lifting, placement and release. Progress on one block must survive later approaches: neither reaching a target pose nor closing the gripper establishes that the intended block is being carried. The observation after release is needed to distinguish a supported placement from a block still held by the robot.}
\appfact{05 / TRAJECTORY}{In the retained Astra episode, the gripper closes at step 157 and the block is lifted by step 179. Camera fitting first appears at step 204, after the initial grasp, and the first placement-and-retraction milestone is step 270. The complete ordered-stack result is successful at step 635. These milestones illustrate a visual estimate followed by geometric refinement, rather than calibration before every motion.}
\noindent\textbf{Retained outcomes.} Astra succeeds, Opus 5.5 fails, Kimi K3 fails, DeepSeek V4.1 Flash fails, Gemini 3.8 Flash fails. One episode per model.
\begin{appprompt}{appnavy}{Reading the episode}Pose arrival, object retention, support and release are separate facts. Preserve clearance for the whole carried block.\end{appprompt}
{\scriptsize\textbf{Record:} \nolinkurl{astra-62}, \nolinkurl{BlockStackingSpecifiedOrderTask}. Recorded views and outcomes refer to this retained episode.}\par\endgroup

\clearpage
\subsection{Task profile 02: Plugging in a charger}
\label{app:card-task-338}
\par\begingroup\appsmall
{\color{appnavy}\sffamily\bfseries RoboDojo \quad / \quad Fitting and insertion}\par\smallskip{}Two fixed ARX X5 arms with parallel-jaw grippers\par\medskip
\noindent\textbf{Objective.} Plug the charger into the power strip.\par\medskip
\begin{center}
\begin{minipage}[b]{.485\linewidth}\centering\includegraphics[width=\linewidth,height=135pt,keepaspectratio]{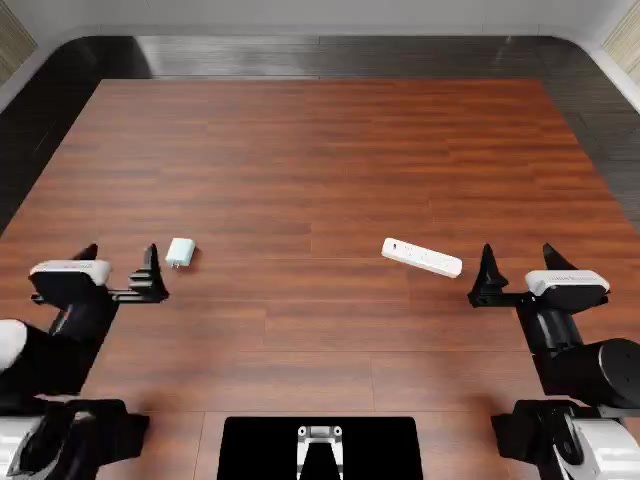}\par{\scriptsize Video 0.10 s}\end{minipage}\hfill
\begin{minipage}[b]{.485\linewidth}\centering\includegraphics[width=\linewidth,height=135pt,keepaspectratio]{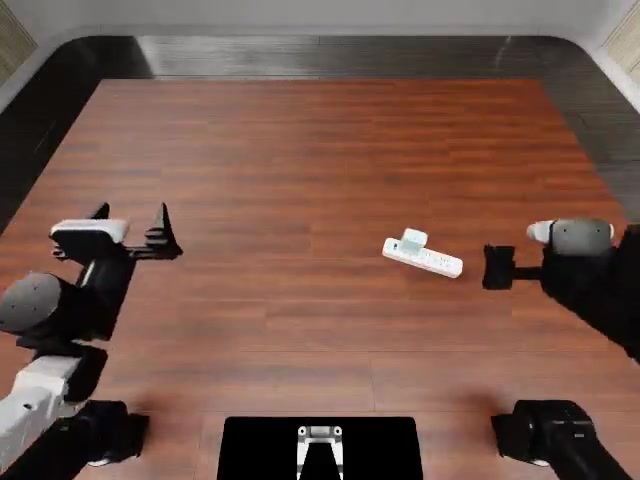}\par{\scriptsize Video 12.61 s}\end{minipage}
\end{center}{\small Astra; recorded head-camera view.}\par\medskip
\appfact{01 / SENSE}{RGB views plus both grasp-point poses and contextual joint angles. Named Cartesian targets are absolute world coordinates; wrist-angle parameters use degrees relative to a downward reference.}
\appfact{02 / ACT}{\tool{move\_eef} changes named pose and gripper fields. Unnamed fields hold their measured call-start values. Opening is 1; closing is 0. The planner chooses trajectory length; the caller does not specify a step count.}
\appfact{03 / VERIFY}{The audited native task requires the charger in the socket region, a 1.5 cm depth threshold and its specified axis aligned upward within 10 degrees. The robot must also satisfy the return-to-home condition.}
\appfact{EPISODE}{400 control steps. The retained Astra episode succeeds at step 323 after 42 robot requests.}
\appfact{04 / EXECUTION}{The control sequence must establish a grasp, transport the charger, orient it relative to the power strip, insert it and return the robot to its required final configuration. Because both arms occupy the workspace, a feasible target for one end effector may still produce a rejected motion when the other arm obstructs the trajectory.}
\appfact{05 / TRAJECTORY}{The retained Astra trajectory contains a rejected rotation at call 17, with the step counter at 196. The next call lowers the idle arm, reaching step 207. Retrying the rotation then succeeds at step 230; the episode subsequently passes the insertion and return conditions at step 323. The feedback therefore changes the configuration before the same intended rotation is attempted again.}
\noindent\textbf{Retained outcomes.} Astra succeeds, Opus 5.5 fails, Kimi K3 fails, DeepSeek V4.1 Flash fails, Gemini 3.8 Flash fails. One episode per model.
\begin{appprompt}{appnavy}{Reading the episode}Lateral alignment alone is insufficient: insertion depth, orientation and robot return state also matter.\end{appprompt}
{\scriptsize\textbf{Record:} \nolinkurl{task-338}, \nolinkurl{plug_in_charger}. Recorded views and outcomes refer to this retained episode.}\par\endgroup

\clearpage
\subsection{Task profile 03: Closing a kitchen drawer}
\label{app:card-task-7}
\par\begingroup\appsmall
{\color{appnavy}\sffamily\bfseries RoboCasa \quad / \quad Articulation and device use}\par\smallskip{}RoboCasa PandaOmron mobile manipulator\par\medskip
\noindent\textbf{Objective.} Close the designated drawer.\par\medskip
\begin{center}
\begin{minipage}[b]{.485\linewidth}\centering\includegraphics[width=\linewidth,height=135pt,keepaspectratio]{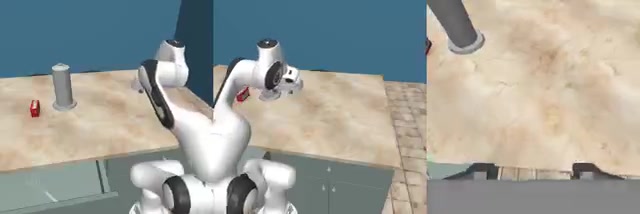}\par{\scriptsize Video 0.10 s}\end{minipage}\hfill
\begin{minipage}[b]{.485\linewidth}\centering\includegraphics[width=\linewidth,height=135pt,keepaspectratio]{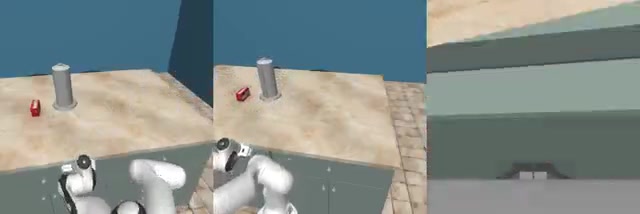}\par{\scriptsize Video 12.93 s}\end{minipage}
\end{center}{\small Kimi K3; recorded multi-camera strip.}\par\medskip
\appfact{01 / SENSE}{RGB and robot proprioception. Reported end-effector pose is base-relative, with XYZW quaternions. The active action is a normalised 12-dimensional controller vector.}
\appfact{02 / ACT}{Arm commands use OSC increments, with translation up to 0.05 m and rotation up to 0.5 rad per repeated step. Repeating a nonzero increment repeats motion. Base and torso commands are normalised inputs.}
\appfact{03 / VERIFY}{All door joints belonging to the designated drawer must have normalised opening at most 0.05. The checker evaluates the articulated state; reaching the handle is an intermediate action.}
\appfact{EPISODE}{450 control steps. Kimi K3 completes this retained episode at step 265 using 19 robot requests.}
\appfact{04 / EXECUTION}{A mobile manipulator must establish a usable base-to-drawer relation before the arm can apply a closing motion. End-effector commands are increments in the robot base frame, so their effect depends on both the current arm pose and the duration of repetition. Small segments near the drawer allow fresh images to reveal progress or unintended contact.}
\appfact{05 / TRAJECTORY}{Kimi K3 shortens its five late pushing requests to 20, 20, 12, 6 and 5 steps. The subsequent withdrawal asks for 12 steps, but the environment executes only two before reporting success at step 265. The shortened final request reflects episode termination by the native checker. It does not mean that the full requested retreat was necessary for the drawer to count as closed.}
\noindent\textbf{Retained outcomes.} Astra succeeds, Opus 5.5 fails, Kimi K3 succeeds, DeepSeek V4.1 Flash fails, Gemini 3.8 Flash fails. One episode per model.
\begin{appprompt}{appnavy}{Reading the episode}The RoboCasa incremental interface differs from RoboLab absolute IK despite similarly named tools.\end{appprompt}
{\scriptsize\textbf{Record:} \nolinkurl{task-7}, \nolinkurl{CloseDrawer}. Recorded views and outcomes refer to this retained episode.}\par\endgroup

\clearpage
\subsection{Task profile 04: Pouring into a cup}
\label{app:card-task-25}
\par\begingroup\appsmall
{\color{appnavy}\sffamily\bfseries RoboDojo \quad / \quad Pouring and processing}\par\smallskip{}Two fixed ARX X5 arms with parallel-jaw grippers\par\medskip
\noindent\textbf{Objective.} Pour liquid from the bottle into the cup.\par\medskip
\begin{center}
\begin{minipage}[b]{.485\linewidth}\centering\includegraphics[width=\linewidth,height=135pt,keepaspectratio]{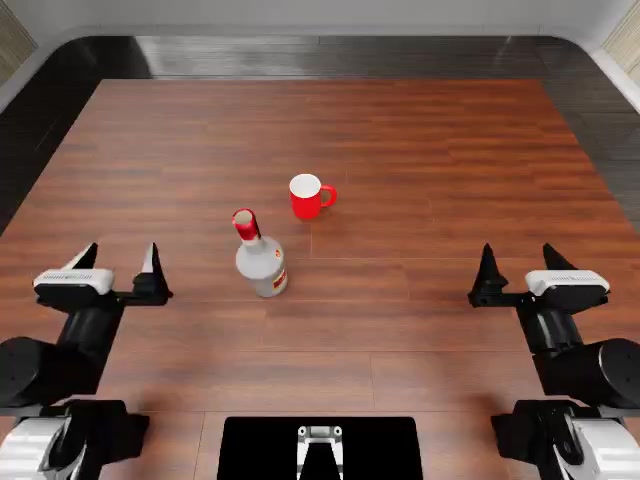}\par{\scriptsize Video 0.10 s}\end{minipage}\hfill
\begin{minipage}[b]{.485\linewidth}\centering\includegraphics[width=\linewidth,height=135pt,keepaspectratio]{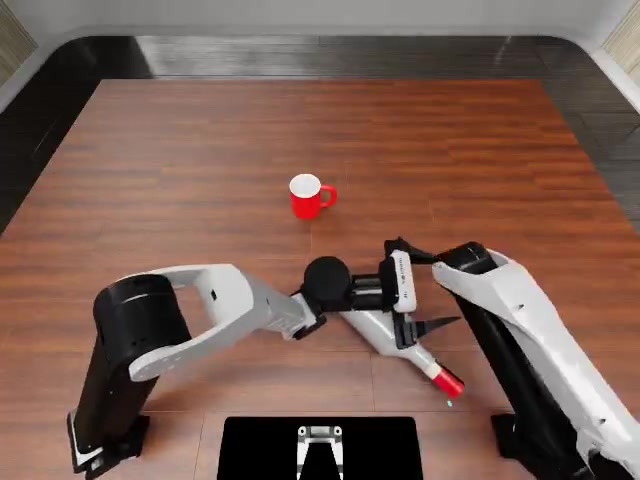}\par{\scriptsize Video 15.52 s}\end{minipage}
\end{center}{\small Kimi K3; recorded head-camera view; unsuccessful episode.}\par\medskip
\appfact{01 / SENSE}{The same world-frame grasp-point interface as charger insertion. RGB observations identify the bottle and cup; no privileged object-coordinate query is supplied.}
\appfact{02 / ACT}{\tool{move\_eef} requests bounded planned motion. Gripper change follows arm arrival in a combined request. The tool reports arrival residuals and planner failures.}
\appfact{03 / VERIFY}{The audited native liquid test is triggered when the bottle becomes upright again within 30 degrees. It uses a cup-transfer threshold of 0.97 and bottle-residual threshold of 0.15 under the original fluid-filtering rule.}
\appfact{EPISODE}{400 control steps. The illustrated Kimi K3 run consumes the horizon without completing the task; 27 robot requests are recorded.}
\appfact{04 / EXECUTION}{Pouring couples object retention, bottle orientation and the relative position of the receiving cup. A successful tool return establishes the executed arm motion, but the next observation must establish whether the bottle remains in the gripper and whether its opening is above the cup. The liquid test is separate from the arm-arrival residual.}
\appfact{05 / TRAJECTORY}{In this Kimi K3 episode, the approach topples the bottle at step 42. Repeated left-arm attempts fail to establish a usable grasp; the agent changes strategy at step 240 and repositions by step 272. Its right-arm attempt remains unfinished when the 400-step horizon is exhausted. The example exposes a delay between losing the intended object state and making a substantial change to the approach.}
\noindent\textbf{Retained outcomes.} Astra fails, Opus 5.5 fails, Kimi K3 fails, DeepSeek V4.1 Flash fails, Gemini 3.8 Flash fails. One episode per model.
\begin{appprompt}{appnavy}{Reading the episode}Moving a grasp point to the requested pose does not show that a bottle was grasped or that liquid entered the cup.\end{appprompt}
{\scriptsize\textbf{Record:} \nolinkurl{task-25}, \nolinkurl{pour_liquid_into_cup}. Recorded views and outcomes refer to this retained episode.}\par\endgroup

\clearpage
\subsection{Task profile 05: Following a courtyard route}
\label{app:card-task-107}
\par\begingroup\appsmall
{\color{appnavy}\sffamily\bfseries WheeledLab \quad / \quad Routes and terrain}\par\smallskip{}MuSHR wheeled vehicle\par\medskip
\noindent\textbf{Objective.} Follow the road to the green parking patch, align with the direction of travel, and stop.\par\medskip
\begin{center}
\begin{minipage}[b]{.485\linewidth}\centering\includegraphics[width=\linewidth,height=135pt,keepaspectratio]{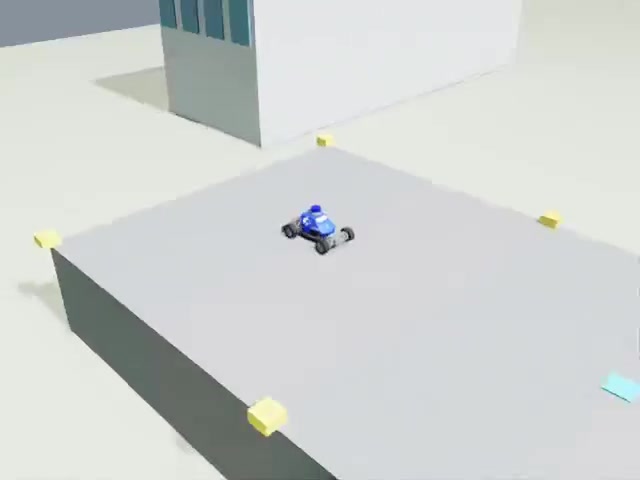}\par{\scriptsize Video 0.10 s}\end{minipage}\hfill
\begin{minipage}[b]{.485\linewidth}\centering\includegraphics[width=\linewidth,height=135pt,keepaspectratio]{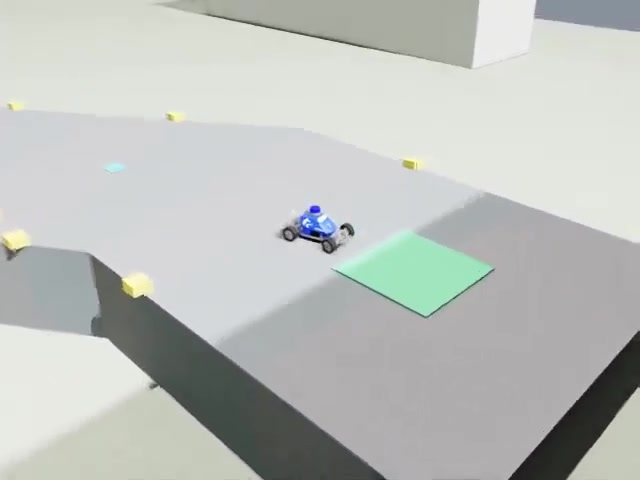}\par{\scriptsize Video 36.54 s}\end{minipage}
\end{center}{\small Kimi K3; external reviewer camera, not the policy front camera.}\par\medskip
\appfact{01 / SENSE}{The policy receives a front RGB camera, wheel and steering encoders, and body IMU attitude/gyro. Global pose, a full map and checkpoint progress are not policy inputs.}
\appfact{02 / ACT}{\tool{drive} supplies normalised speed and steering. In this MuSHR profile, negative speed clamps to zero; there is no reverse. A held segment spans 1--50 control steps at 50 Hz.}
\appfact{03 / VERIFY}{For this archived contract, complete the ordered route and reach the destination within 0.65 m, heading within 25 degrees and ground speed at most 0.12 m/s for 0.8 s. Road and collision conditions remain active.}
\appfact{EPISODE}{2,000 control steps (40 simulated seconds). Kimi K3 succeeds at step 1,881 with 6/6 checkpoints and a recorded 0.8 s parking hold.}
\appfact{04 / EXECUTION}{The route requires steering through ordered checkpoints before satisfying the terminal parking condition. Steering and speed must be inferred from the front-camera scene and local sensors: the external video is useful for reviewing the episode, but it does not supply the agent with a map or global route progress. Shortening a held command creates another opportunity to observe before a turn.}
\appfact{05 / TRAJECTORY}{The retained Kimi K3 result reaches all six checkpoints and records a 0.8-second parking hold, succeeding at step 1,881 of 2,000. Reaching the green region alone would leave heading, speed and dwell requirements unresolved. Conversely, a low final speed away from the required destination would not count as route completion. The trajectory must satisfy the route and terminal checks together.}
\noindent\textbf{Retained outcomes.} Astra succeeds, Opus 5.5 succeeds, Kimi K3 succeeds, DeepSeek V4.1 Flash fails, Gemini 3.8 Flash fails. One episode per model.
\begin{appprompt}{appnavy}{Reading the episode}The dwell interval is specific to this recorded version. Do not substitute a newer arrival-only checker for a historical trajectory.\end{appprompt}
{\scriptsize\textbf{Record:} \nolinkurl{task-107}, \nolinkurl{rw-courtyard}. Recorded views and outcomes refer to this retained episode.}\par\endgroup

\clearpage
\subsection{Task profile 06: Quadruped disturbance recovery}
\label{app:card-task-142}
\par\begingroup\appsmall
{\color{appnavy}\sffamily\bfseries Go2 Push \quad / \quad Balance and recovery}\par\smallskip{}Go2 quadruped, 12 joint-position residuals\par\medskip
\noindent\textbf{Objective.} Recover near the prescribed root goal while remaining upright and slowing down under the retained scene dynamics.\par\medskip
\begin{center}
\begin{minipage}[b]{.485\linewidth}\centering\includegraphics[width=\linewidth,height=135pt,keepaspectratio]{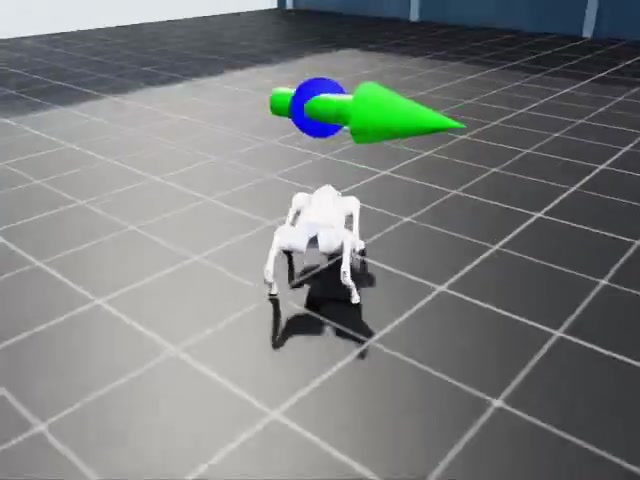}\par{\scriptsize Video 0.10 s}\end{minipage}\hfill
\begin{minipage}[b]{.485\linewidth}\centering\includegraphics[width=\linewidth,height=135pt,keepaspectratio]{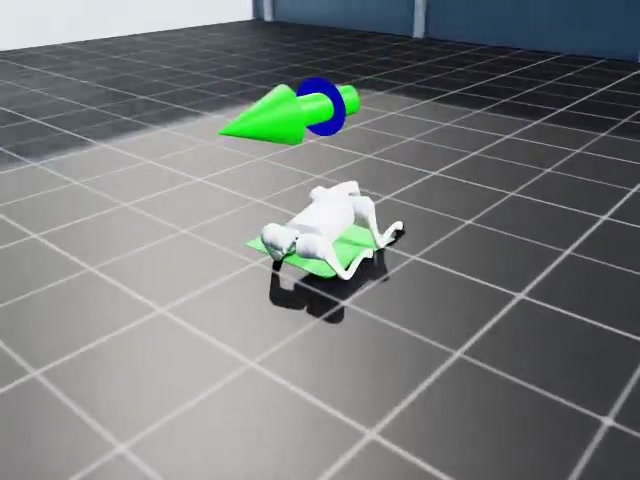}\par{\scriptsize Video 19.40 s}\end{minipage}
\end{center}{\small Opus 5.5; third-person reviewer video.}\par\medskip
\appfact{01 / SENSE}{Named native actor terms include body velocities, projected gravity, commands, joint state and previous action. These are simulator-derived actor observations, not an RGB-only setting.}
\appfact{02 / ACT}{\tool{apply\_action} supplies 12 simultaneous residuals. Joint targets follow $q_{\rm target}=q_{\rm default}+0.25u$ radians. Native motor PD remains active; no learned gait controller is supplied.}
\appfact{03 / VERIFY}{The recorded World-state checker requires, throughout the final 2 s, horizontal goal error at most 0.25 m, horizontal speed at most 0.10 m/s and tilt at most 20 degrees. Native or scene failure overrides success.}
\appfact{EPISODE}{1,000 control steps at 50 Hz (20 simulated seconds). Opus 5.5 completes the full window; final error is 0.192 m, speed 0.0022 m/s and tilt 4.48 degrees.}
\appfact{04 / EXECUTION}{Twelve residual joint targets are applied together, so posture regulation requires coordinating the entire stance. The previous action and measured body motion provide feedback for the next bounded request. The native PD controller converts targets into motor responses; it does not choose a learned recovery policy for the agent.}
\appfact{05 / TRAJECTORY}{The retained Opus 5.5 episode completes the 1,000-step horizon. Its terminal horizontal error, speed and tilt are below their individual thresholds, but these endpoint values alone are not the success test. The checker also requires the conjunction to hold throughout the final two seconds. This separates sustained recovery from briefly crossing a favourable pose during continuing motion.}
\noindent\textbf{Retained outcomes.} Astra fails, Opus 5.5 succeeds, Kimi K3 fails, DeepSeek V4.1 Flash fails, Gemini 3.8 Flash fails. One episode per model.
\begin{appprompt}{appnavy}{Reading the episode}The binary World-state criterion is distinct from the upstream velocity-tracking reward. The task name alone does not measure recovery from every possible disturbance.\end{appprompt}
{\scriptsize\textbf{Record:} \nolinkurl{task-142}, \nolinkurl{quadruped_push_recovery}. Recorded views and outcomes refer to this retained episode.}\par\endgroup

\clearpage
\subsection{Task profile 07: Stabilising a suspended payload}
\label{app:card-task-144}
\par\begingroup\appsmall
{\color{appnavy}\sffamily\bfseries OmniDrones \quad / \quad Stabilisation and tracking}\par\smallskip{}Quadrotor with a 1 m suspended payload\par\medskip
\noindent\textbf{Objective.} Stabilise the payload at its target and suppress swing under native random payload pushes.\par\medskip
\begin{center}
\begin{minipage}[b]{.485\linewidth}\centering\includegraphics[width=\linewidth,height=135pt,keepaspectratio]{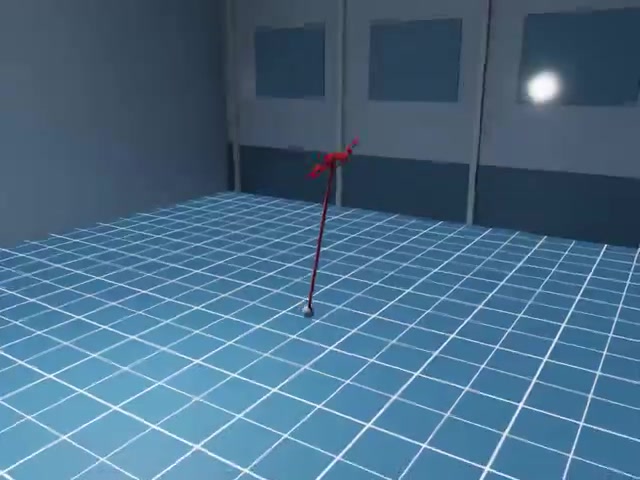}\par{\scriptsize Video 0.10 s}\end{minipage}\hfill
\begin{minipage}[b]{.485\linewidth}\centering\includegraphics[width=\linewidth,height=135pt,keepaspectratio]{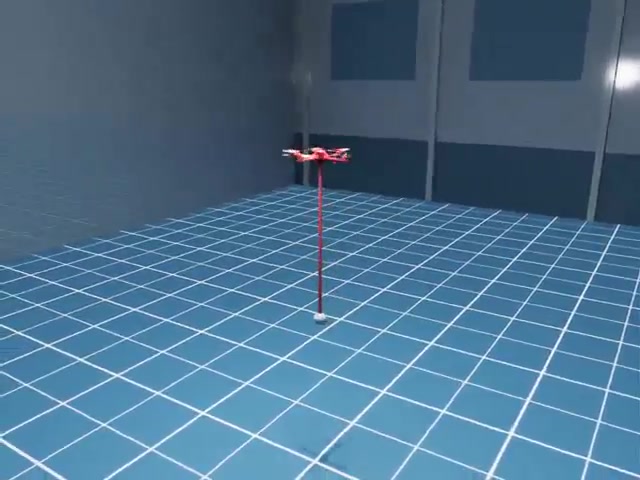}\par{\scriptsize Video 7.84 s}\end{minipage}
\end{center}{\small Opus 5.5; external reviewer video.}\par\medskip
\appfact{01 / SENSE}{The native actor vector exposes relative payload/target geometry, drone attitude and motion, and declared motor-state terms. No hidden evaluator state is returned.}
\appfact{02 / ACT}{\tool{apply\_action} supplies four rotor inputs. Desired internal throttle is $s^*=\sqrt{\operatorname{clip}((u+1)/2,0,1)}$; thrust depends on $s^2$. Motor lag remains active.}
\appfact{03 / VERIFY}{For the final 2 s: payload error at most 0.15 m, payload speed at most 0.20 m/s, suspension swing at most 10 degrees, and drone tilt at most 15 degrees. All bounds must hold together through the end.}
\appfact{EPISODE}{500 control steps at 62 Hz (8.0645 simulated seconds). Opus 5.5 passes with final payload error 0.0394 m and speed 0.0147 m/s.}
\appfact{04 / EXECUTION}{The controller must regulate both the quadrotor and its suspended load. Changing thrust can correct altitude while exciting swing, and motor lag separates the requested rotor input from its immediate physical effect. Feedback therefore needs to track payload displacement and velocity alongside drone tilt, rather than treating an upright drone as sufficient evidence of success.}
\appfact{05 / TRAJECTORY}{The Opus 5.5 trajectory finishes with payload error 0.0394 m and speed 0.0147 m/s. The recorded joint-stability interval lasts 2.15 seconds, exceeding the required two seconds, with all four bounds satisfied together. The episode illustrates maintaining a recovered state until the end instead of stopping at the first apparently stable observation.}
\noindent\textbf{Retained outcomes.} Astra fails, Opus 5.5 succeeds, Kimi K3 fails, DeepSeek V4.1 Flash fails, Gemini 3.8 Flash fails. One episode per model.
\begin{appprompt}{appnavy}{Reading the episode}Stable drone attitude does not imply a stable payload. A single passing frame does not satisfy the temporal window.\end{appprompt}
{\scriptsize\textbf{Record:} \nolinkurl{task-144}, \nolinkurl{drone_payload_hover}. Recorded views and outcomes refer to this retained episode.}\par\endgroup

\clearpage
\subsection{Task profile 08: Repeated drone ball juggling}
\label{app:card-task-158}
\par\begingroup\appsmall
{\color{appnavy}\sffamily\bfseries VolleyBots \quad / \quad Ball interaction}\par\smallskip{}Iris quadrotor with native body-contact dynamics\par\medskip
\noindent\textbf{Objective.} Keep the ball airborne through repeated legal contacts and sufficiently high flight arcs.\par\medskip
\begin{center}
\begin{minipage}[b]{.485\linewidth}\centering\includegraphics[width=\linewidth,height=135pt,keepaspectratio]{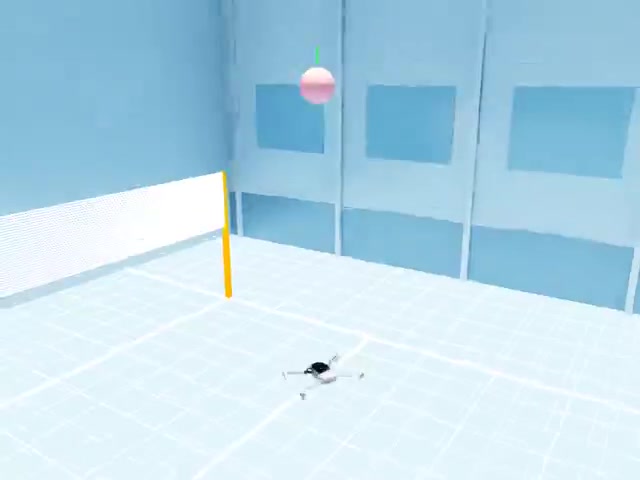}\par{\scriptsize Video 0.10 s}\end{minipage}\hfill
\begin{minipage}[b]{.485\linewidth}\centering\includegraphics[width=\linewidth,height=135pt,keepaspectratio]{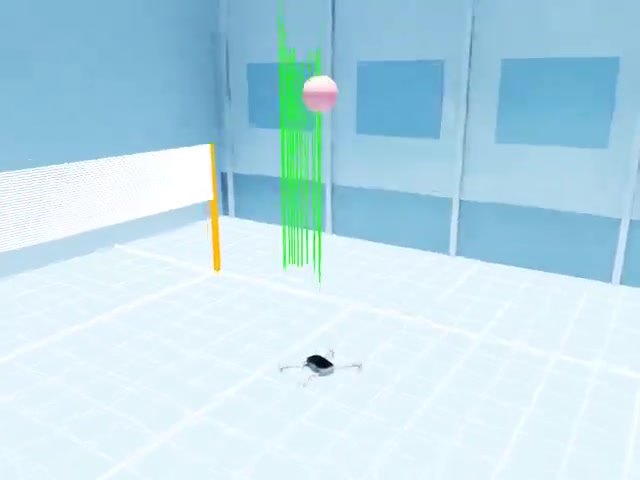}\par{\scriptsize Video 15.52 s}\end{minipage}
\end{center}{\small Opus 5.5; external reviewer video, unavailable to the policy.}\par\medskip
\appfact{01 / SENSE}{A 32-value native actor vector supplies drone pose/motion, anchor and ball-relative terms, ball velocity and episode time. The review camera is not part of the actor input.}
\appfact{02 / ACT}{Four rotor commands through \tool{apply\_action}, held for bounded segments at 50 Hz. No automatic ball-launch or racket assistance is added. Thinking pauses simulator time.}
\appfact{03 / VERIFY}{At least four height-qualified hits are required. A qualifying arc exceeds ball-centre height 3.5 m and is confirmed at the next valid contact. Contacts separated by at most 25 steps are invalid. The final 0.6 s must keep the ball airborne without support contact.}
\appfact{EPISODE}{800 control steps (16 simulated seconds). Opus 5.5 records 15 true hits, 14 height-qualified hits and 274 robot requests, completing the full horizon.}
\appfact{04 / EXECUTION}{The agent alternates phases that bring the drone into contact with the ball and phases that prepare for its next return. Commands expose rotor inputs rather than a high-level strike primitive. Ball position and velocity must be interpreted with the drone state; a rendered near-contact frame cannot establish a legal hit or reconstruct the intervening flight arc.}
\appfact{05 / TRAJECTORY}{Opus 5.5 issues 274 robot requests across the 800-step horizon, with a median three executed steps per request. The checker records 15 valid hits and 14 height-qualified hits. The trajectory also satisfies the final airborne interval. These three facts are distinct: frequent calls alone do not ensure valid contacts, and enough historical hits do not excuse support contact in the final window.}
\noindent\textbf{Retained outcomes.} Astra fails, Opus 5.5 succeeds, Kimi K3 fails, DeepSeek V4.1 Flash fails, Gemini 3.8 Flash fails. One episode per model.
\begin{appprompt}{appnavy}{Reading the episode}Visible proximity is not a contact event. The final position alone cannot reconstruct the legal-hit sequence.\end{appprompt}
{\scriptsize\textbf{Record:} \nolinkurl{task-158}, \nolinkurl{drone_volleyball_solo_juggle}. Recorded views and outcomes refer to this retained episode.}\par\endgroup

\clearpage
\section{Execution, Observations, and Evaluation Records}
\label{app:execution}
\subsection{What the archive contains}
The offline archive contains one self-contained task page for each of the 420 retained model--task records. Each page carries a manifest entry and embedded run, tool and environment records, a merged event trace, and any available videos. We verified the SHA-256 of all 420 pages against the five-model result audit before extracting this appendix. The frozen snapshot is \texttt{snapshot-20261006T025143Z}. The publication-side inventory maps each task to its source group, primary domain, model, local trace ID, outcome and page hash.

The data support three distinct readings. The manifest supplies the \emph{retained scored outcome}; the embedded run preserves the \emph{historical terminal result}; and the event/video stream records the \emph{observed execution}. These may differ after budget adjudication. Appendix~\ref{app:diagnostics} gives a concrete example in which the historical terminal success occurs after the scored boundary. We never replace the retained score with the last visually plausible frame.

\subsection{Registered robot tools across the benchmark}
The following catalog is extracted from the startup tool schemas of all 420 retained records. It describes tools offered to the model, including tools that a particular episode never calls. Tool-name sets agree across the five models within each source group. Parameters and action maps can still vary by task. All 20 source groups are covered: the 13 native-action groups are expanded in the action-map table below.
\par\begingroup\fontsize{9}{11}\selectfont\setlength{\tabcolsep}{5pt}\renewcommand{\arraystretch}{1.16}
\begin{longtable}{>{\raggedright\arraybackslash}p{0.27\linewidth}>{\raggedright\arraybackslash}p{0.07\linewidth}>{\raggedright\arraybackslash}p{0.58\linewidth}}\toprule \textbf{Source group} & \textbf{Tasks} & \textbf{Registered robot tools}\\\midrule\endhead
RoboDojo & 15 & \texttt{move\_eef}\\[3pt]
RoboLab & 10 & \texttt{move\_eef}, \texttt{set\_gripper}, \texttt{move\_robot}\\[3pt]
RoboCasa & 10 & \texttt{move\_eef}, \texttt{move\_base}, \texttt{move\_torso}, \texttt{set\_gripper}, \texttt{move\_robot}\\[3pt]
BEHAVIOR-1K & 10 & \texttt{move\_arms}, \texttt{move\_base}, \texttt{set\_grippers}, \texttt{move\_torso}, \texttt{move\_robot}\\[3pt]
WheeledLab & 11 & \texttt{observe}, \texttt{drive}\\[3pt]
AI-CPS & 3 & \texttt{observe}, \texttt{move\_joints}\\[3pt]
HumanoidSoccer & 1 & \texttt{move\_joints}\\[3pt]
Native-action adapters (13 source groups) & 24 & \texttt{observe}, \texttt{apply\_action}\\[3pt]
\bottomrule\end{longtable}\endgroup
\noindent\textbf{Robot tools and workspace tools.} Shell execution and image-file viewing belong to the frontier harness, not to the robot-tool catalog. They support calculations, inspection of supplied observations and notes; they do not advance physics. An \texttt{observe} call also advances no control steps, but remains subject to the interaction budget. Adapters without a standalone \texttt{observe} tool return observations through motion feedback.

\noindent\textbf{Effective availability.} No retained startup registers \texttt{coding\_control}, reset, score-query or give-up tools. Some inherited prompt passages describe optional coding control; the registered tool list and disabled run setting determine what was actually available. The source supplement \texttt{appendix\_interfaces.json} stores the original instructions and schemas, linked to each run and source-page hash.
\clearpage\subsection{Manipulation tools: arguments and execution meaning}
Most motion tools take a short \texttt{note}, a duration \texttt{steps}, and a \texttt{targets} object. RoboDojo is an exception: its planner chooses the trajectory duration, so the exposed call has no \texttt{steps} argument. End-effector (EEF) coordinates and gripper polarity differ across adapters.
\par\begingroup\fontsize{9}{11}\selectfont\setlength{\tabcolsep}{5pt}\renewcommand{\arraystretch}{1.16}
\begin{longtable}{>{\raggedright\arraybackslash}p{0.13\linewidth}>{\raggedright\arraybackslash}p{0.23\linewidth}>{\raggedright\arraybackslash}p{0.56\linewidth}}\toprule \textbf{Adapter} & \textbf{Tool} & \textbf{Inputs and meaning}\\\midrule\endhead
RoboDojo & \texttt{move\_eef} & \texttt{targets}: selected left/right XYZ (world metres), wrist pitch/roll/yaw (degrees from the downward reference), and gripper opening (0 closed, 1 open). Omitted fields hold measured call-start values. The planner moves the arm before applying a combined gripper change.\\[3pt]
RoboLab & \texttt{move\_eef} & \texttt{targets.position[3]} and/or \texttt{quaternion\_wxyz[4]}: absolute flange pose in the robot-root frame. Omitted pose components hold measured values. \texttt{steps=1..30} at 15 Hz.\\[3pt]
RoboLab & \texttt{set\_gripper} & \texttt{targets.gripper\_close}: 0 opens, 1 closes; the target persists. The arm holds while the segment executes.\\[3pt]
RoboLab & \texttt{move\_robot} & Combines the same arm-pose and gripper fields in one native action. Closure starts with arm motion, not after arrival.\\[3pt]
RoboCasa & \texttt{move\_eef} & \texttt{targets.eef\_delta[6]}: normalised translation and rotation-vector increments in the base frame. Scales are 0.05 m and 0.5 rad per step. \texttt{steps=1..30}; repetition repeats the increment.\\[3pt]
RoboCasa & \texttt{move\_base} & \texttt{targets.base\_motion[3]}: normalised XY/yaw controller inputs, not displacement. Uses base-following mode; gripper target persists.\\[3pt]
RoboCasa & \texttt{move\_torso} & \texttt{targets.torso}: normalised input in [-1,1]; requests a slide-joint increment of 0.05 times the input in metres per tick. Not an absolute torso height.\\[3pt]
RoboCasa & \texttt{set\_gripper} & \texttt{targets.gripper\_close}: 0 opens, 1 closes. No arm/base/torso motion is requested.\\[3pt]
RoboCasa & \texttt{move\_robot} & Combines EEF, base, torso and gripper fields. Optional \texttt{control\_mode=0} updates from achieved arm pose; 1 uses the desired pose for base following.\\[3pt]
\bottomrule\end{longtable}\endgroup
\noindent\textbf{Same name, different motion.} Holding a RoboLab absolute pose repeats a fixed target. Holding a RoboCasa nonzero EEF delta requests an increment every step. A RoboDojo target addresses the grasp point between the jaws; RoboLab addresses the flange. These distinctions determine whether a repeated call holds, accumulates motion, or changes the physical point being controlled.
\clearpage\subsection{Mobile manipulation, driving and named joints}
\par\begingroup\fontsize{9}{11}\selectfont\setlength{\tabcolsep}{5pt}\renewcommand{\arraystretch}{1.16}
\begin{longtable}{>{\raggedright\arraybackslash}p{0.16\linewidth}>{\raggedright\arraybackslash}p{0.21\linewidth}>{\raggedright\arraybackslash}p{0.55\linewidth}}\toprule \textbf{Adapter} & \textbf{Tool} & \textbf{Inputs and meaning}\\\midrule\endhead
BEHAVIOR-1K & \texttt{move\_arms} & Left/right XYZ in robot-root metres and \texttt{left/right\_quat\_xyzw}. Absolute EEF IK; omitted arm holds its measured pose. All tools in this group use \texttt{steps=1..30} at 30 Hz.\\[3pt]
BEHAVIOR-1K & \texttt{move\_base} & \texttt{base\_vx}, \texttt{base\_vy} in local-body m/s (limits +/-0.3), and \texttt{base\_wz} in rad/s (+/-0.5). Arms hold root-relative targets; omitted base velocities are zero.\\[3pt]
BEHAVIOR-1K & \texttt{set\_grippers} & \texttt{left/right\_gripper}: continuous opening in [0,1], with 0 closed and 1 open. Targets persist; EEFs hold measured poses and base motion stops.\\[3pt]
BEHAVIOR-1K & \texttt{move\_torso} & \texttt{trunk\_qpos[4]}: absolute trunk joint positions in radians, using the recorded joint order and limits. This differs from the normalised slide increment in RoboCasa.\\[3pt]
BEHAVIOR-1K & \texttt{move\_robot} & Combines arm poses, gripper openings, base velocities and trunk targets simultaneously. One control step counts once, regardless of how many components are commanded.\\[3pt]
WheeledLab & \texttt{observe} & Returns the current allowed observation without stepping physics. Authored onboard courses expose front RGB, encoders and IMU; the call does not reveal map, world pose or checkpoint progress.\\[3pt]
WheeledLab & \texttt{drive} & \texttt{action[2]} = normalised speed and steering, each in [-1,1]; \texttt{steps=1..50}. Speed scales to 3 m/s wheel targets; steering uses 0.488 followed by the native mapping. Reverse availability is task-specific.\\[3pt]
AI-CPS & \texttt{observe}; \texttt{move\_joints} & Observation costs no physics steps. Motion uses \texttt{arm\_action[7]} in [-1,1] for 1--50 steps. Each tick adds 0.125 times the input in radians to the previous target, before limits and native noise; fingers remain task-controlled.\\[3pt]
HumanoidSoccer & \texttt{move\_joints} & \texttt{joint\_positions}: a dictionary of named G1 joint targets in absolute radians; omitted joints hold measured call-start positions. \texttt{steps=1..50} at 50 Hz. Targets pass through the original PD torque controller. If the recorded ankle-balance assist is enabled, it can adjust ankle targets; it is not a walking policy.\\[3pt]
\bottomrule\end{longtable}\endgroup
\noindent\textbf{Profile-specific details.} The archived courtyard profile clamps negative speed to zero. Reverse-bay and parallel-parking profiles accept signed reverse motion. Likewise, \texttt{move\_joints} names absolute joint positions in HumanoidSoccer but repeated normalised increments in AI-CPS. The supplied action contract, not the tool name alone, determines the mapping.
\clearpage\subsection{Native-vector tools and action maps}
All 13 source groups below register \texttt{observe(note)} and \texttt{apply\_action(note, action, steps)}. Observation returns the current native actor-policy input without stepping physics. Execution supplies the entire vector simultaneously for 1--50 control steps. Native gains, lags, constraints and per-task termination remain active. The vector length is validated against the archived schema; the entries are not universally joint angles or bounded to [-1,1].
\par\begingroup\fontsize{9}{11}\selectfont\setlength{\tabcolsep}{5pt}\renewcommand{\arraystretch}{1.16}
\begin{longtable}{>{\raggedright\arraybackslash}p{0.23\linewidth}>{\raggedright\arraybackslash}p{0.06\linewidth}>{\raggedright\arraybackslash}p{0.63\linewidth}}\toprule \textbf{Source group} & \textbf{Dim.} & \textbf{Meaning of the action vector}\\\midrule\endhead
ReflexBench & 8 & Absolute root-frame EEF XYZ + WXYZ quaternion + gripper sign. Positive gripper opens; non-positive closes.\\[3pt]
Digit & 26 & Joint-position targets with the recorded offsets and 0.5 rad input scale, in runtime joint order.\\[3pt]
Flamingo & 8 & Six actuator-position channels and two wheel-velocity channels (40 rad/s scale). Leg channels use motor-space angles with gear ratio -1.5.\\[3pt]
Go2 Push & 12 & Joint-position residuals: $q_{\rm target}=q_{\rm default}+0.25u$ radians.\\[3pt]
OmniIsaacGymEnvs & 12 & ANYmal joint-position offsets: $q_{\rm target}=q_{\rm default}+0.5u$ radians, tracked by native PD.\\[3pt]
OmniDrones & 4 & Four rotor inputs. -1 requests zero thrust; +1 requests maximum. Zero is half maximum steady-state thrust, not hover.\\[3pt]
Bench2Dex & 52 & Absolute arm and dexterous-finger joint targets in radians. Both UR5 arms and both hands act simultaneously; no binary gripper command.\\[3pt]
Robot Lab & 12 & A1 joint-position residuals with the recorded default posture and 0.25 rad scale; native target clipping remains active.\\[3pt]
SteadyTray & 29 & G1 joint-position targets from per-joint scales plus the default posture. The delayed native actuators remain active.\\[3pt]
TTRL & 21 & Booster T1 joint-position residuals with 0.25 rad scale plus the default posture; raw inputs are clipped to [-100,100].\\[3pt]
VolleyBots & 4 & Four Iris rotor inputs with the native motor mapping and lag; no automatic attitude or ball-tracking policy.\\[3pt]
Wheel-Legged & 6 & Virtual leg angle, leg length and wheel speed for each side. Scales: 0.35 rad, 0.06 m around 0.237 m, and 24 rad/s; native VMC maps them to motor torques.\\[3pt]
Wheeled Quadruped & 4 & Two front-thigh position residuals (0.5 rad scale) and two rear-wheel velocity targets (5 rad/s scale).\\[3pt]
\bottomrule\end{longtable}\endgroup
\noindent\textbf{What feedback establishes.} A valid request can still produce tracking error, collision, a lost grasp or task failure. Conversely, an invalid request can consume an interaction without advancing physics. Inspect the validation receipt, completed-step count and returned observation separately from the evaluator's task outcome.
\clearpage\subsection{Control-step and interaction accounting}
Physics advances through accepted robot execution and is paused during model deliberation and offline calculations. A requested segment can terminate early, be rejected, or finish without meeting the task goal. We therefore distinguish requested steps, confirmed executed steps and the final evaluator's step count. The Astra stacking record, for example, has a retained result at step 635 and completed environment events through step 634; its terminal feedback is unavailable. This one-step discrepancy is preserved rather than silently repaired.

All 420 run records report environment-side code control disabled. Shell analysis is a separate permission; the retained RoboDojo comparison uses Shell-on. Some inherited prompt text describes an optional coding interface. Its mention does not override the effective run setting or demonstrate that a coding-control call was executed.

The scored protocol uses task-specific physical horizons. For 83 tasks the consecutive non-action limit is 15, with total limits of 30, 60 or 120 according to the task. Volleyball 1v1 uses 20 consecutive and 150 total. Auxiliary calls, zero-step robot requests, and otherwise action-free completed turns are counted under the observable-event policy. Confirmed control execution resets the consecutive counter only. API retries are excluded. The logs mark accounting as incomplete where built-in calls, terminal polling, or turn-end identifiers are not fully observable; an absent trace item is not automatically zero resource use.

\subsection{Success checks and their observation window}
Sixty-eight tasks per model use a retained \texttt{native} scoring profile and sixteen use \texttt{world-state-v1}. ``Native'' is a record label: for authored driving scenarios it includes the versioned independent geometry/trajectory evaluator. Native numerical rewards remain diagnostics and are not averaged across heterogeneous tasks. The World-state records expose whether the full evaluation horizon completed, whether the scene was valid, the sampled predicates, and any overriding failures.

For temporal checks, the claim is limited to the logged control ticks, including reset when recorded. It is not a guarantee over unobserved continuous physics. In payload hovering, all four bounds must hold throughout the final two seconds. In juggling, both the event count and the final airborne window matter. The cards in Appendix~\ref{app:spec} show the thresholds used in these retained examples.

\subsection{Frame provenance}
The selected task prompts document current observations and up to four historical observations sampled every two feedback rounds, with step labels. Observation rounds have variable physical duration. The environment state supplied to the agent, the history retained by the host, and the continuous reviewer video are separate records.

Every appendix frame is extracted from an embedded source video with a recorded video timestamp and SHA-256. Frames are not generated illustrations. A video timestamp denotes playback time in that file, not elapsed model time or a guaranteed one-to-one correspondence with a tool event. External driving and aerial cameras are reviewer views, not additional model observations. Stage captions identify trace-event indices separately, and late frames can precede terminal success. The accompanying \texttt{appendix\_frames.json} and \texttt{appendix\_excerpts.json} resolve these references to source files.

\clearpage
\section{Prompts and Agent Instructions}
\label{app:models}
System instructions define the robot, permitted observations and action conventions; developer instructions specify workspace use and interaction rules. Below we reproduce six examples from the evaluated episodes, grouped by embodiment. The task goal and current observations accompany these instructions, while Appendix~\ref{app:execution} describes the callable tools. Quotes are verbatim excerpts with whitespace normalised; separate paragraphs omit intervening text. Environment-side code control is disabled throughout the experiments.
\subsection{Tabletop manipulation}
Both tabletop interfaces expose RGB images and robot state, but their Cartesian targets refer to different physical points. RoboDojo commands the grasp point between the jaws, whereas RoboLab commands the gripper flange. Their gripper conventions also differ: zero closes in RoboDojo and opens in RoboLab. These distinctions are stated explicitly in the instructions.

\begin{appprompt}{appnavy}{RoboDojo / grasp-point control}\appsmall\raggedright
\textbf{System instructions (excerpts)}\par
``You are controlling a real robot embodiment named 'robodojo-arx-x5'. You receive RGB camera images, the current world-frame grasp-point state of both arms in the same 14 dimensions move\_eef takes, arm joint angles as context you cannot command, and a task instruction. Move with move\_eef by naming only the world-frame dimensions you want to change. Cartesian targets must be estimated from RGB; no depth or world-coordinate query is available. Respond with exactly one tool call per turn. After each motion the next observation reports how far the grasp point ended up from what you asked for, so check it before assuming a motion landed.''\par\smallskip
``Gripper 0 is fully closed, 1 is fully open, and commanding 0 always closes as far as the object allows.''\par\smallskip
\textbf{Developer instructions (excerpt)}\par ``Use move\_eef for robot motion. Each result includes fresh observations. Never use shell, files, web or other tools to control the scene or read hidden state. Execute one motion at a time and inspect its result.''\par
\end{appprompt}\par\medskip
\begin{appprompt}{appnavy}{RoboLab / flange-pose control}\appsmall\raggedright
\textbf{System instructions (excerpts)}\par
``You control the fixed-base DROID robot in RoboLab. Complete the exact upstream task instruction using supplied RGB and measured proprioception only. The robot has one Franka arm and a Robotiq gripper; it has no mobile base, torso or head control. All positions are in metres relative to the robot articulation root; all quaternions are WXYZ. ee\_pos/ee\_quat describe the gripper base\_link flange, not the fingertips. eef\_pos/eef\_quat describe a differently oriented reporting frame, NOT the IK target frame.''\par\smallskip
``gripper\_close=0 opens and 1 closes. Use move\_robot to combine arm and gripper in the same native action; closure begins alongside movement, not on arrival. Separate tools are sequential. A tool segment costs steps once; the native rate is 15 Hz.''\par\smallskip
\textbf{Developer instructions}\par ``Shell/code and /workspace are for local calculations and notes. Read-only /observations contains only allowed RGB and proprioception. All movement and fresh observations must use the robot tools. Never read or change simulator/evaluator files or seek external task solutions.''\par
\end{appprompt}\par\medskip
\clearpage\subsection{Mobile manipulation}
Mobile manipulation adds coordination between the arm, gripper, torso and base. RoboCasa uses incremental end-effector commands; BEHAVIOR-1K accepts absolute arm poses and physical base velocities. The instructions explain which components move together and which hold their current targets.

\begin{appprompt}{appnavy}{RoboCasa / incremental mobile manipulation}\appsmall\raggedright
\textbf{System instructions (excerpts)}\par
``You control the standard RoboCasa365 PandaOmron robot using benchmark-specific robot tools. Complete the exact task instruction from the environment. Use only supplied RGB and proprioception. There is no depth sensor enabled in this standard evaluation. EEF pose observations are relative to the robot base; quaternions are XYZW. Actions are the native normalized 12D controller inputs, NOT the absolute IK targets of other benchmarks. move\_eef supplies base-frame OSC increments: normalized translation scales to up to 0.05m and rotation to 0.5rad per step. Repeating a delta repeats motion; use small magnitudes and short segments near objects. Base and torso commands are normalized, not metres or m/s. Gripper close=1/open=0 persists. Unspecified motion inputs are zero. control\_mode=0 uses achieved arm pose; 1 follows desired pose when base moves.''\par\smallskip
\textbf{Developer instructions}\par ``Use shell/code only for local computation and memory in /workspace. Read-only /observations contains RGB PNG and measured robot state. All motion and new observations must use robot tools. Do not access simulator/evaluator files or bypass the interface. A final answer does not terminate the episode.''\par
\end{appprompt}\par\medskip
\begin{appprompt}{appnavy}{BEHAVIOR-1K / coordinated dual-arm motion}\appsmall\raggedright
\textbf{System instructions (excerpts)}\par
``You control an R1Pro mobile dual-arm robot through benchmark-specific move\_robot, move\_base, move\_arms, set\_grippers and (when available) move\_torso tools. Complete the named household task. Use move\_robot to combine both arm poses, base velocities, gripper openings and available trunk targets in one simultaneous bounded action. Each control step applies all components together and counts once toward the budget; separate tool calls execute sequentially, not concurrently. Combined gripper changes begin alongside motion, not after arm arrival.''\par\smallskip
``Use only onboard RGB, depth and proprioception. EEF positions and XYZW quaternions are relative to the robot articulation root, not world coordinates. Read the measured starting poses. Base local +x forward, +y left, +z up; base velocities in m/s and yaw rad/s. Unspecified arms hold their observed pose during the action; gripper targets persist (0 closed, 1 open). Base velocities default to zero on each call.''\par\smallskip
\textbf{Developer instructions (excerpt)}\par ``Use shell, code execution and files in /workspace for computation, planning, maps and persistent memory. Allowed sensor exports are read-only in /observations; each numbered directory includes RGB PNGs, raw depth NPYs and proprioception.''\par
\end{appprompt}\par\medskip
\clearpage\subsection{Driving and flight}
Driving and flight require different observation and control descriptions. The WheeledLab example supplies onboard visual and proprioceptive observations with speed and steering commands. The OmniDrones example instead supplies native actor measurements and direct rotor inputs. Both distinguish policy observations from external review-camera views.

\begin{appprompt}{appnavy}{WheeledLab / onboard driving}\appsmall\raggedright
\textbf{System instructions (excerpts)}\par
``You drive a real-collision simulated MuSHR in a RobotWorld custom driving course. Observation profile robotworld-onboard-v2: front camera, ideal wheel/steering encoders and body IMU gyro/tilt. These simulate sensors a real car can carry; no added sensor noise yet. No global position/yaw, true translational velocity/slip, full map, obstacle coordinates, checkpoint progress, next waypoint, gate pose, gate clearance Boolean, friction map or future perturbations are given. Infer road shape, hazards and stopping locations from the onboard image. Review-camera video and independent evaluator records are NOT policy inputs. Do not try to recover private state through shell/files/network.''\par\smallskip
``Only observe and drive are enabled; no per-tick program execution tool is available.''\par\smallskip
\textbf{Developer instructions}\par ``Use shell only for calculations in /workspace. Only dynamic tools control physics. /observations contains allowed policy observations. A final answer does not end the episode. Non-action interaction budget: 15 consecutive or 30 total calls/empty turns. Auxiliary tool calls and robot calls executing zero control steps each cost one. A normal turn with no executed control steps and no counted calls costs one. Confirmed control steps reset only the consecutive counter. API retries do not consume this budget. The evaluator ends the episode at either limit.''\par
\end{appprompt}\par\medskip
\begin{appprompt}{appnavy}{OmniDrones / direct rotor control}\appsmall\raggedright
\textbf{System instructions (excerpts)}\par
``You directly control a Hummingbird quadrotor through FOUR normalized rotor commands in native order. action\_transform=null; no Lee controller, hover policy, automatic attitude stabilizer or trajectory tracker runs for you. All four commands are simultaneous, each in [-1,1]: -1 targets zero thrust, +1 maximum thrust. Zero is half maximum steady-state thrust, NOT a neutral/hover command. Native actuator first-order throttle response remains active.''\par\smallskip
``Only the original actor policy group is supplied. It may include simulator-derived state, commands or a task's native predictor; consult this task's explicit observation contract. Never claim pure visual control if state is included. Critic/privileged evaluator observations are excluded. Third-person review video is never a policy input. Native sensor images, if any, are labelled separately.''\par\smallskip
\textbf{Developer instructions}\par ``Only dynamic robot tools control physics. Shell is for calculations. Continue until the episode terminates or budget ends. A final answer does not end the episode. Non-action interaction budget: 15 consecutive or 120 total calls/empty turns. Auxiliary tool calls and robot calls executing zero control steps each cost one. A normal turn with no executed control steps and no counted calls costs one. Confirmed control steps reset only the consecutive counter. API retries do not consume this budget. The evaluator ends the episode at either limit.''\par
\end{appprompt}\par\medskip

\clearpage
\section{Full Quantitative Results}
\label{app:results}
Section~\ref{sec:evaluation} presents the aggregate results and domain comparisons. Here we provide the complete task-level outcomes and numerical resource distributions for all five models.
\subsection{All task--model outcomes}
Table~\ref{tab:app-all-results} includes every task exactly once. Columns A, O, K, D and G denote Astra, Opus 5.5, Kimi K3, DeepSeek V4.1 Flash and Gemini 3.8 Flash. A green check denotes retained success; a red cross denotes retained failure. A dagger marks an adjudicated record. The horizon is the task's recorded control-step limit; N and W indicate native and World-state scoring profiles. Task domains and descriptions are provided in Appendix~\ref{app:spec}.
\par\begingroup\fontsize{8.2}{10}\selectfont\setlength{\tabcolsep}{3pt}\renewcommand{\arraystretch}{1.13}
\begin{longtable}{rp{.43\linewidth}rrccccc}
\caption{Complete 84-task inventory and five-model outcomes.}\label{tab:app-all-results}\\
\toprule\textbf{ID}&\textbf{Source / task identifier}&\textbf{Limit}&\textbf{Score}&\textbf{A}&\textbf{O}&\textbf{K}&\textbf{D}&\textbf{G}\\\midrule\endfirsthead
\multicolumn{9}{l}{\small Table~\ref{tab:app-all-results} continued}\\
\toprule\textbf{ID}&\textbf{Source / task identifier}&\textbf{Limit}&\textbf{Score}&\textbf{A}&\textbf{O}&\textbf{K}&\textbf{D}&\textbf{G}\\\midrule\endhead
\midrule\multicolumn{9}{r}{\scriptsize Continued on next page}\\\endfoot\bottomrule\endlastfoot
\rowcolor{appgray}\multicolumn{9}{l}{\textbf{RoboDojo}}\\*
01 & \nolinkurl{hang_mugs} & 800 & N & \rwcross & \rwcross & \rwcross & \rwcross & \rwcross \\
02 & \nolinkurl{sweep_blocks} & 1000 & N & \rwcross & \rwcheck & \rwcross & \rwcross & \rwcross \\
03 & \nolinkurl{pour_liquid_into_cup} & 400 & N & \rwcross & \rwcross & \rwcross & \rwcross & \rwcross \\
04 & \nolinkurl{make_toast} & 1400 & N & \rwcross & \rwcross & \rwcross & \rwcross & \rwcross \\
05 & \nolinkurl{store_laptop_and_headphones} & 800 & N & \rwcross & \rwcross & \rwcross & \rwcross & \rwcross \\
06 & \nolinkurl{insert_tubes} & 500 & N & \rwcross & \rwcross & \rwcross & \rwcross & \rwcross \\
07 & \nolinkurl{plug_in_charger} & 400 & N & \rwcheck & \rwcross & \rwcross & \rwcross & \rwcross \\
08 & \nolinkurl{pour_balls_into_vase} & 600 & N & \rwcross & \rwcross & \rwcross & \rwcross & \rwcross \\
09 & \nolinkurl{play_Xylophone} & 500 & N & \rwcross & \rwcross & \rwcross & \rwcross & \rwcross \\
10 & \nolinkurl{fill_pen_holder} & 1100 & N & \rwcross & \rwcross & \rwcross & \rwcross & \rwcross \\
11 & \nolinkurl{fill_egg_holder} & 700 & N & \rwcross & \rwcross & \rwcross & \rwcross & \rwcross \\
12 & \nolinkurl{make_kong} & 600 & N & \rwcross & \rwcross & \rwcross & \rwcross & \rwcross \\
13 & \nolinkurl{pour_by_language} & 800 & N & \rwcross & \rwcross & \rwcross & \rwcross & \rwcross \\
14 & \nolinkurl{match_and_pick_from_conveyor} & 700 & N & \rwcheck & \rwcheck & \rwcross & \rwcross & \rwcheck \\
15 & \nolinkurl{deposit_coin} & 300 & N & \rwcross & \rwcross & \rwcross & \rwcross & \rwcross \\
\rowcolor{appgray}\multicolumn{9}{l}{\textbf{BEHAVIOR-1K}}\\*
16 & \nolinkurl{carrying_in_groceries} & 2000 & N & \rwcross & \rwcross$^{\dagger}$ & \rwcross & \rwcross & \rwcross \\
17 & \nolinkurl{clean_up_your_desk} & 2000 & N & \rwcross & \rwcross$^{\dagger}$ & \rwcross & \rwcross & \rwcross \\
18 & \nolinkurl{slicing_vegetables} & 2000 & N & \rwcross & \rwcross$^{\dagger}$ & \rwcross & \rwcross & \rwcross \\
19 & \nolinkurl{sorting_vegetables} & 2000 & N & \rwcross & \rwcross & \rwcross & \rwcross & \rwcross \\
20 & \nolinkurl{clean_boxing_gloves} & 2000 & N & \rwcross & \rwcross$^{\dagger}$ & \rwcross & \rwcross & \rwcross \\
21 & \nolinkurl{putting_up_Christmas_decorations_inside} & 2000 & N & \rwcross & \rwcross$^{\dagger}$ & \rwcross & \rwcross & \rwcross \\
22 & \nolinkurl{setting_the_table} & 2000 & N & \rwcross & \rwcross$^{\dagger}$ & \rwcross & \rwcross & \rwcross \\
23 & \nolinkurl{putting_dishes_away_after_cleaning} & 2000 & N & \rwcross & \rwcross$^{\dagger}$ & \rwcross & \rwcross & \rwcross \\
24 & \nolinkurl{can_meat} & 2000 & N & \rwcross & \rwcross & \rwcross & \rwcross & \rwcross \\
25 & \nolinkurl{freeze_pies} & 2000 & N & \rwcross & \rwcross & \rwcross & \rwcross & \rwcross \\
\rowcolor{appgray}\multicolumn{9}{l}{\textbf{RoboCasa}}\\*
26 & \nolinkurl{CountertopCleanup} & 600 & N & \rwcross & \rwcross & \rwcross$^{\dagger}$ & \rwcross & \rwcross \\
27 & \nolinkurl{SortingCleanup} & 3000 & N & \rwcross & \rwcross$^{\dagger}$ & \rwcross$^{\dagger}$ & \rwcross & \rwcross \\
28 & \nolinkurl{CoffeeSetupMug} & 600 & N & \rwcheck & \rwcross & \rwcross$^{\dagger}$ & \rwcross & \rwcross \\
29 & \nolinkurl{CloseDrawer} & 450 & N & \rwcheck & \rwcross & \rwcheck & \rwcross & \rwcross \\
30 & \nolinkurl{NavigateKitchen} & 450 & N & \rwcheck & \rwcross & \rwcross & \rwcross & \rwcross \\
31 & \nolinkurl{PackIdenticalLunches} & 3900 & N & \rwcross & \rwcross & \rwcross & \rwcross & \rwcross \\
32 & \nolinkurl{OrganizeMugsByHandle} & 1350 & N & \rwcross & \rwcross & \rwcross$^{\dagger}$ & \rwcross & \rwcross \\
33 & \nolinkurl{LoadDishwasher} & 1800 & N & \rwcheck & \rwcross & \rwcross & \rwcross & \rwcross \\
34 & \nolinkurl{MicrowaveCorrectMeal} & 1500 & N & \rwcross & \rwcross & \rwcross$^{\dagger}$ & \rwcross & \rwcross \\
35 & \nolinkurl{ResetCabinetDoors} & 3300 & N & \rwcross & \rwcross & \rwcross$^{\dagger}$ & \rwcross & \rwcross \\
\rowcolor{appgray}\multicolumn{9}{l}{\textbf{RoboLab}}\\*
36 & \nolinkurl{ToolOrganizationTask} & 2700 & N & \rwcross & \rwcross$^{\dagger}$ & \rwcross & \rwcross & \rwcross \\
37 & \nolinkurl{NonHammerToolsInRightBinTask} & 2700 & N & \rwcross & \rwcross$^{\dagger}$ & \rwcross & \rwcross & \rwcross \\
38 & \nolinkurl{FoodPacking2CansTask} & 2700 & N & \rwcheck & \rwcross$^{\dagger}$ & \rwcross & \rwcross & \rwcross \\
39 & \nolinkurl{RubiksCubeLeftOfBowlTask} & 450 & N & \rwcheck & \rwcheck & \rwcross & \rwcross & \rwcross \\
40 & \nolinkurl{FruitsOnPlate3Task} & 3000 & N & \rwcheck & \rwcheck & \rwcross & \rwcross & \rwcross \\
41 & \nolinkurl{PutTwoMugsOnShelfTask} & 2700 & N & \rwcross & \rwcross$^{\dagger}$ & \rwcross & \rwcross & \rwcross \\
42 & \nolinkurl{BlockStackingSpecifiedOrderTask} & 1350 & N & \rwcheck & \rwcross$^{\dagger}$ & \rwcross & \rwcross & \rwcross \\
43 & \nolinkurl{ClutterPlasticTask} & 2700 & N & \rwcheck & \rwcheck & \rwcross & \rwcross & \rwcross \\
44 & \nolinkurl{ReorientWhiteMugsTask} & 900 & N & \rwcross & \rwcross & \rwcross & \rwcross & \rwcross \\
45 & \nolinkurl{WhiteMugInCenterOfTableTask} & 450 & N & \rwcheck & \rwcheck & \rwcross & \rwcross & \rwcross \\
\rowcolor{appgray}\multicolumn{9}{l}{\textbf{HumanoidSoccer}}\\*
46 & \nolinkurl{play-soccer} & 300 & N & \rwcross & \rwcross & \rwcross & \rwcross & \rwcross \\
\rowcolor{appgray}\multicolumn{9}{l}{\textbf{AI-CPS}}\\*
47 & \nolinkurl{22} & 300 & N & \rwcross & \rwcross & \rwcross & \rwcross & \rwcross \\
48 & \nolinkurl{23} & 300 & N & \rwcross & \rwcross & \rwcross & \rwcross & \rwcross \\
49 & \nolinkurl{24} & 300 & N & \rwcheck & \rwcross & \rwcross & \rwcross & \rwcross \\
\rowcolor{appgray}\multicolumn{9}{l}{\textbf{WheeledLab}}\\*
50 & \nolinkurl{mushr-drift} & 400 & W & \rwcross & \rwcross$^{\dagger}$ & \rwcross & \rwcross & \rwcross \\
51 & \nolinkurl{f1tenth-drift} & 400 & W & \rwcross & \rwcross & \rwcross & \rwcross & \rwcross \\
52 & \nolinkurl{elevation} & 200 & N & \rwcross & \rwcross & \rwcross & \rwcross & \rwcross \\
53 & \nolinkurl{visual} & 150 & W & \rwcross & \rwcross$^{\dagger}$ & \rwcross & \rwcross & \rwcross \\
54 & \nolinkurl{rw-courtyard} & 2000 & N & \rwcheck & \rwcheck & \rwcheck & \rwcross & \rwcross \\
55 & \nolinkurl{rw-hairpins} & 2000 & N & \rwcheck & \rwcheck & \rwcross & \rwcross & \rwcross \\
56 & \nolinkurl{rw-gate-dock} & 2000 & N & \rwcross & \rwcross & \rwcross & \rwcross & \rwcross \\
57 & \nolinkurl{rw-drift-switch} & 2000 & N & \rwcross & \rwcross & \rwcross & \rwcross & \rwcross \\
58 & \nolinkurl{rw-twin-beam} & 2000 & N & \rwcross & \rwcross & \rwcross & \rwcross & \rwcross \\
59 & \nolinkurl{rw-reverse-bay} & 2000 & N & \rwcross & \rwcross & \rwcross & \rwcross & \rwcross \\
60 & \nolinkurl{rw-parallel-park} & 2000 & N & \rwcross & \rwcross & \rwcross & \rwcross & \rwcross \\
\rowcolor{appgray}\multicolumn{9}{l}{\textbf{Bench2Dex}}\\*
61 & \nolinkurl{41} & 681 & N & \rwcross & \rwcross & \rwcross & \rwcross & \rwcross \\
62 & \nolinkurl{42} & 986 & N & \rwcross & \rwcross & \rwcross & \rwcross & \rwcross \\
63 & \nolinkurl{43} & 1149 & N & \rwcross & \rwcross & \rwcross & \rwcross & \rwcross \\
64 & \nolinkurl{44} & 1080 & N & \rwcross & \rwcross & \rwcross & \rwcross & \rwcross \\
65 & \nolinkurl{45} & 964 & N & \rwcross & \rwcross & \rwcross & \rwcross & \rwcross \\
66 & \nolinkurl{46} & 1061 & N & \rwcross & \rwcross & \rwcross & \rwcross & \rwcross \\
67 & \nolinkurl{47} & 593 & N & \rwcross & \rwcross & \rwcross & \rwcross & \rwcross \\
68 & \nolinkurl{48} & 1213 & N & \rwcross & \rwcross & \rwcross & \rwcross & \rwcross \\
69 & \nolinkurl{49} & 1082 & N & \rwcross & \rwcross & \rwcross & \rwcross & \rwcross \\
\rowcolor{appgray}\multicolumn{9}{l}{\textbf{Digit}}\\*
70 & \nolinkurl{digit_walk_hand_tracking} & 700 & W & \rwcross & \rwcross & \rwcross & \rwcross & \rwcross \\
\rowcolor{appgray}\multicolumn{9}{l}{\textbf{Flamingo}}\\*
71 & \nolinkurl{wheel_legged_jump_balance} & 1000 & W & \rwcross & \rwcross & \rwcross & \rwcross & \rwcross \\
\rowcolor{appgray}\multicolumn{9}{l}{\textbf{Go2 Push}}\\*
72 & \nolinkurl{quadruped_push_recovery} & 1000 & W & \rwcross & \rwcheck & \rwcross & \rwcross & \rwcross \\
\rowcolor{appgray}\multicolumn{9}{l}{\textbf{OmniDrones}}\\*
73 & \nolinkurl{drone_payload_hover} & 500 & W & \rwcross & \rwcheck & \rwcross & \rwcross & \rwcross \\
74 & \nolinkurl{drone_inverted_pendulum_tracking} & 620 & W & \rwcross & \rwcross & \rwcross & \rwcross & \rwcross \\
\rowcolor{appgray}\multicolumn{9}{l}{\textbf{OmniIsaacGymEnvs}}\\*
75 & \nolinkurl{anymal_rough_terrain} & 800 & W & \rwcross & \rwcross & \rwcross & \rwcross & \rwcross \\
\rowcolor{appgray}\multicolumn{9}{l}{\textbf{ReflexBench}}\\*
76 & \nolinkurl{cup_ball_catching} & 100 & N & \rwcross & \rwcheck & \rwcross & \rwcross & \rwcross \\
\rowcolor{appgray}\multicolumn{9}{l}{\textbf{Robot Lab}}\\*
77 & \nolinkurl{a1_front_leg_handstand} & 500 & W & \rwcross & \rwcross & \rwcross & \rwcross & \rwcross \\
\rowcolor{appgray}\multicolumn{9}{l}{\textbf{SteadyTray}}\\*
78 & \nolinkurl{tray_balancing_walk} & 1000 & W & \rwcross & \rwcross & \rwcross & \rwcross & \rwcross \\
\rowcolor{appgray}\multicolumn{9}{l}{\textbf{TTRL}}\\*
79 & \nolinkurl{humanoid_table_tennis_return} & 1500 & W & \rwcross & \rwcross & \rwcross & \rwcross & \rwcross \\
\rowcolor{appgray}\multicolumn{9}{l}{\textbf{VolleyBots}}\\*
80 & \nolinkurl{drone_volleyball_1v1} & 1000 & N & \rwcheck & \rwcheck & \rwcross & \rwcheck & \rwcross \\
81 & \nolinkurl{drone_volleyball_solo_juggle} & 800 & W & \rwcross & \rwcheck & \rwcross & \rwcross & \rwcross \\
\rowcolor{appgray}\multicolumn{9}{l}{\textbf{Wheel-Legged}}\\*
82 & \nolinkurl{wheel_legged_upright_recovery} & 2000 & W & \rwcross & \rwcross & \rwcross & \rwcross & \rwcross \\
83 & \nolinkurl{wheel_legged_rough_terrain} & 2000 & W & \rwcross & \rwcross & \rwcross & \rwcross & \rwcross \\
\rowcolor{appgray}\multicolumn{9}{l}{\textbf{Wheeled Quadruped}}\\*
84 & \nolinkurl{rear_wheel_upright_balance} & 1000 & W & \rwcross & \rwcross & \rwcross & \rwcross & \rwcross \\
\end{longtable}\endgroup
\subsection{Resource distributions}
Tables~\ref{tab:app-resource-summary} and~\ref{tab:app-resource-breakdown} report medians and interquartile ranges (IQRs) across tasks. Time and call counts cover all 84 episodes per model, including failures. Intervals describe task variation, not repeat-trial uncertainty. Full logs can extend beyond the evaluation boundary; timing includes setup, waiting and execution.
\begin{table}[H]\centering\small
\caption{\textbf{Time and interaction use.} Medians and interquartile ranges across tasks.}\label{tab:app-resource-summary}
\begingroup\setlength{\tabcolsep}{3pt}\renewcommand{\arraystretch}{1.2}
\begin{tabularx}{\linewidth}{@{}l*{6}{R}@{}}\toprule
\rowcolor{expheader}\textbf{Model} & \multicolumn{2}{c}{\textbf{Elapsed time (min)}} & \multicolumn{2}{c}{\textbf{Robot requests}} & \multicolumn{2}{c}{\textbf{All tool calls}}\\
 & Median & \textcolor{expmuted}{IQR} & Median & \textcolor{expmuted}{IQR} & Median & \textcolor{expmuted}{IQR}\\\midrule
Astra & \textbf{30.9} & \textcolor{expmuted}{[15.2, 61.2]} & \textbf{58} & \textcolor{expmuted}{[25.5, 94]} & \textbf{70} & \textcolor{expmuted}{[30, 138.8]} \\
\rowcolor{expstripe}
Opus 5.5 & \textbf{54} & \textcolor{expmuted}{[24.5, 137.3]} & \textbf{37} & \textcolor{expmuted}{[16, 87.2]} & \textbf{57.5} & \textcolor{expmuted}{[31, 148]} \\
Kimi K3 & \textbf{67.6} & \textcolor{expmuted}{[21.3, 181.7]} & \textbf{42.5} & \textcolor{expmuted}{[11.8, 70]} & \textbf{61.5} & \textcolor{expmuted}{[18.5, 95]} \\
\rowcolor{expstripe}
DeepSeek & \textbf{16} & \textcolor{expmuted}{[11.1, 40.5]} & \textbf{22} & \textcolor{expmuted}{[5, 59.2]} & \textbf{33} & \textcolor{expmuted}{[18, 70]} \\
Gemini & \textbf{12.8} & \textcolor{expmuted}{[10, 25.8]} & \textbf{11} & \textcolor{expmuted}{[0, 36.8]} & \textbf{31} & \textcolor{expmuted}{[15, 52.2]} \\
\bottomrule\end{tabularx}\endgroup\end{table}

Control-step statistics use the available observed step counts from full logs: 84 episodes for Astra, 80 for Opus 5.5, 82 for Kimi, 75 for DeepSeek and 60 for Gemini. Missing entries are excluded. Images supplied within robot feedback are not additional image-view calls; zero explicit views therefore does not imply absence of visual observations.
\begin{table}[H]\centering\small
\caption{\textbf{Physical execution and auxiliary-tool use.} Medians and interquartile ranges across tasks.}\label{tab:app-resource-breakdown}
\begingroup\setlength{\tabcolsep}{3pt}\renewcommand{\arraystretch}{1.2}
\begin{tabularx}{\linewidth}{@{}l*{6}{R}@{}}\toprule
\rowcolor{expheader}\textbf{Model} & \multicolumn{2}{c}{\textbf{Control steps}} & \multicolumn{2}{c}{\textbf{Shell calls}} & \multicolumn{2}{c}{\textbf{Image views}}\\
 & Median & \textcolor{expmuted}{IQR} & Median & \textcolor{expmuted}{IQR} & Median & \textcolor{expmuted}{IQR}\\\midrule
Astra & \textbf{627} & \textcolor{expmuted}{[298, 1165]} & \textbf{4.5} & \textcolor{expmuted}{[0, 21.2]} & \textbf{0} & \textcolor{expmuted}{[0, 0]} \\
\rowcolor{expstripe}
Opus 5.5 & \textbf{458} & \textcolor{expmuted}{[178.5, 913.8]} & \textbf{15} & \textcolor{expmuted}{[0, 28]} & \textbf{0} & \textcolor{expmuted}{[0, 10]} \\
Kimi K3 & \textbf{449.5} & \textcolor{expmuted}{[158, 918.5]} & \textbf{1} & \textcolor{expmuted}{[0, 14.2]} & \textbf{0} & \textcolor{expmuted}{[0, 1.2]} \\
\rowcolor{expstripe}
DeepSeek & \textbf{347} & \textcolor{expmuted}{[62, 742]} & \textbf{8.5} & \textcolor{expmuted}{[0, 16]} & \textbf{2.5} & \textcolor{expmuted}{[0, 8]} \\
Gemini & \textbf{315} & \textcolor{expmuted}{[78.5, 625]} & \textbf{15} & \textcolor{expmuted}{[9.8, 23]} & \textbf{0} & \textcolor{expmuted}{[0, 2.2]} \\
\bottomrule\end{tabularx}\endgroup\end{table}

\subsection{Aggregate token and cost accounting}
\label{app:aggregate-cost}
The values in Table~\ref{tab:app-aggregate-cost} accompany the three point plots in Figure~\ref{fig:cost-tradeoffs}. All three measures reproduce the project website's valid per-task mean scaled to 84 tasks, as retrieved on 7 October 2026. Missing or zero-flagged usage is excluded from the mean, not counted as zero. Kimi has 57 valid token and cost records; each other model has 84. Thus Kimi's aggregate consumption is an estimate extrapolated to the full task set, not the sum of the recorded values. Tokens include cached input and output and do not sum repeated cumulative updates. Monetary values use the website's list-price estimates rather than invoices. Elapsed time uses all 84 website-reported durations per model, whose rounding can cause small differences from raw timestamps. All measures include failed tasks; full-run consumption can include execution after a retrospective scoring boundary. Provider token definitions, cache accounting and unit prices differ, so token and cost rankings need not coincide.
\begin{table}[H]\centering\small
\caption{\textbf{Aggregate consumption for the five-model evaluation.} Monetary values are reported list-price estimates. Kimi token and cost totals extrapolate from 57 valid records to 84 tasks.}\label{tab:app-aggregate-cost}
\begin{tabular}{@{}lrrr@{}}\toprule
\textbf{Model} & \textbf{Tokens} & \textbf{Estimated USD} & \textbf{Elapsed hours}\\\midrule
Astra & 941M & 9,912.80 & 54.6\\
Opus 5.5 & 1.63B & 2,916.43 & 140.3\\
Kimi K3 & 2.00B & 624.09 & 163.8\\
DeepSeek V4.1 Flash & 907M & 34.49 & 35.7\\
Gemini 3.8 Flash & 303M & 66.68 & 28.7\\\bottomrule
\end{tabular}
\end{table}

\clearpage
\section{Successful Trajectories}
\label{app:trajectories}
We present four deliberately selected successes covering sequential manipulation, dynamic stabilisation, recovery after a rejected action, and competitive interaction. Each case connects observed state, the agent's next decision, and the resulting execution feedback. Every image and quoted excerpt comes from our own retained archive. Selection is qualitative; these examples do not estimate behavioural prevalence.

Event numbers are zero-based indices in the merged trace. ``Returned step'' is the tool event's recorded step, not a guarantee that every requested step executed. Tool notes express the agent's stated intent and can be wrong. Evaluator fields establish the retained outcome; images and visible messages explain the sequence.
\subsection{Astra: building and checking an ordered stack}
\noindent\textbf{Record:} \nolinkurl{astra-62}. \textbf{Outcome:} success. \textbf{Robot requests:} 35. \textbf{Scored steps:} 635/1350.\par
\begin{figure}[H]\centering
\begin{minipage}[b]{0.319\linewidth}\centering\includegraphics[width=\linewidth,height=112pt,keepaspectratio]{figures/appendix/astra-62-0.jpg}\par{\scriptsize Video 0.10\,s}\end{minipage}\hfill
\begin{minipage}[b]{0.319\linewidth}\centering\includegraphics[width=\linewidth,height=112pt,keepaspectratio]{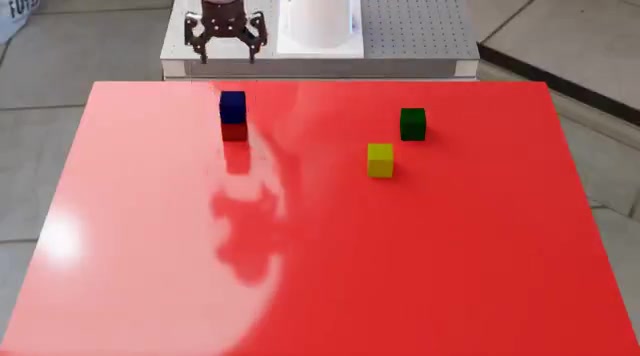}\par{\scriptsize Video 16.90\,s}\end{minipage}\hfill
\begin{minipage}[b]{0.319\linewidth}\centering\includegraphics[width=\linewidth,height=112pt,keepaspectratio]{figures/appendix/astra-62-3.jpg}\par{\scriptsize Video 40.98\,s}\end{minipage}
\caption{\textbf{Astra stacking.} Recorded external-camera frames from \texttt{astra-62}. The final shown frame still precedes the terminal release request. The complete run, rather than this frame alone, supplies the success label.}\label{fig:app-astra-62}\end{figure}
\appstage{1}{The agent leaves the red block as the base and explicitly orders the remaining placements. Three initial movements probe the image-to-robot relation before descent.}
\begin{appexcerpt}{astra-62 | event 4}\appsmall \textbf{Agent message:} ``I'll leave red as the base, then place blue, green, and yellow on top, checking each grasp and placement.
''\end{appexcerpt}
\appstage{2}{After placing blue, the agent requests a vertical retreat before starting the next grasp. This separates release verification from moving toward green; the transcript then reuses the established stack centre.}
\begin{appexcerpt}{astra-62 | event 276 | returned step 270}\appsmall \textbf{Agent tool note:} ``Retract vertically to verify the released blue-on-red stack and clear it before approaching green.''\end{appexcerpt}
\appstage{3}{The same approach--grasp--lift--place sequence is repeated for green and yellow. The final request opens the gripper; the retained evaluator reports success at step 635. Because the final client receipt is missing, we preserve the last completed-event step of 634 separately.}
\begin{appexcerpt}{astra-62 | event 684}\appsmall \textbf{Agent tool note:} ``Release yellow onto green without moving the arm.''\end{appexcerpt}
\clearpage\subsection{Opus 5.5: repeated contacts under rotor-level control}
\noindent\textbf{Record:} \nolinkurl{task-158}. \textbf{Outcome:} success. \textbf{Robot requests:} 274. \textbf{Scored steps:} 800/800.\par
\begin{figure}[H]\centering
\begin{minipage}[b]{0.319\linewidth}\centering\includegraphics[width=\linewidth,height=112pt,keepaspectratio]{figures/appendix/task-158-0.jpg}\par{\scriptsize Video 0.10\,s}\end{minipage}\hfill
\begin{minipage}[b]{0.319\linewidth}\centering\includegraphics[width=\linewidth,height=112pt,keepaspectratio]{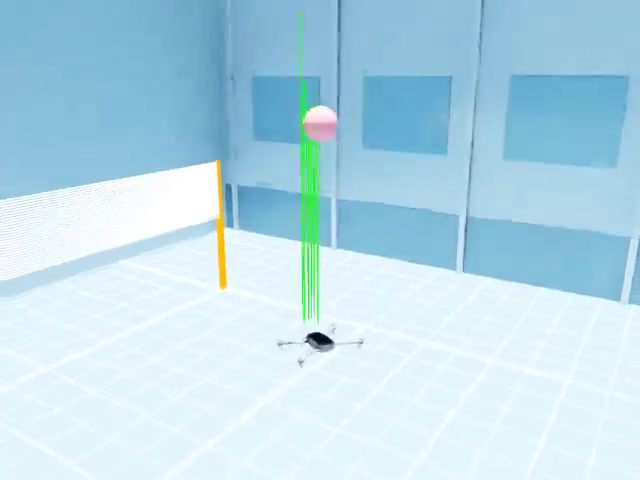}\par{\scriptsize Video 6.40\,s}\end{minipage}\hfill
\begin{minipage}[b]{0.319\linewidth}\centering\includegraphics[width=\linewidth,height=112pt,keepaspectratio]{figures/appendix/task-158-3.jpg}\par{\scriptsize Video 15.52\,s}\end{minipage}
\caption{\textbf{Opus 5.5 juggling.} Reviewer-camera frames from \texttt{task-158}. The policy receives a native actor vector, not these rendered views. Hit counts and final-window validity come from the retained checker.}\label{fig:app-task-158}\end{figure}
\appstage{1}{The first requests use two-step segments to level the vehicle and revise the rotor mixing. A native motor mapping and lag mediate the requested inputs, so command values alone do not establish achieved thrust.}
\begin{appexcerpt}{task-158 | event 12 | returned step 6}\appsmall \textbf{Agent tool note:} ``corrected mixer (roll arm .21, pitch arm .13, equal trim)''\end{appexcerpt}
\appstage{2}{The later trace alternates descent, upward acceleration and post-contact cuts. The note below identifies a planned strike, while the independent event counters establish whether contact and height qualification actually occurred.}
\begin{appexcerpt}{task-158 | event 599 | returned step 437}\appsmall \textbf{Agent tool note:} ``strike 9 accel, tilt trim''\end{appexcerpt}
\appstage{3}{The agent continues until step 800. The retained record reports 15 true hits and 14 height-qualified hits, exceeding the required four, with the final 0.6-second airborne condition satisfied and no recorded overriding failure. The 274 requests illustrate frequent feedback in this particular success.}
\begin{appexcerpt}{task-158 | event 1070 | returned step 800}\appsmall \textbf{Agent tool note:} ``hover to episode end''\end{appexcerpt}
\clearpage\subsection{Gemini: recovering from an unreachable lift}
\noindent\textbf{Record:} \nolinkurl{task-272}. \textbf{Outcome:} success. \textbf{Robot requests:} 48. \textbf{Scored steps:} 511/700.\par
\begin{figure}[H]\centering
\begin{minipage}[b]{0.319\linewidth}\centering\includegraphics[width=\linewidth,height=112pt,keepaspectratio]{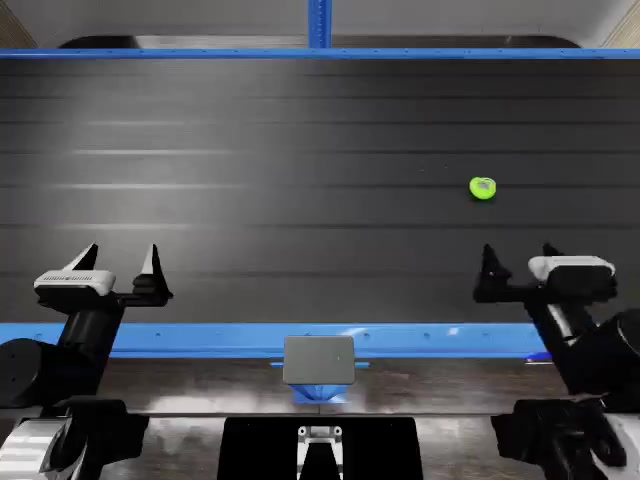}\par{\scriptsize Video 0.10\,s}\end{minipage}\hfill
\begin{minipage}[b]{0.319\linewidth}\centering\includegraphics[width=\linewidth,height=112pt,keepaspectratio]{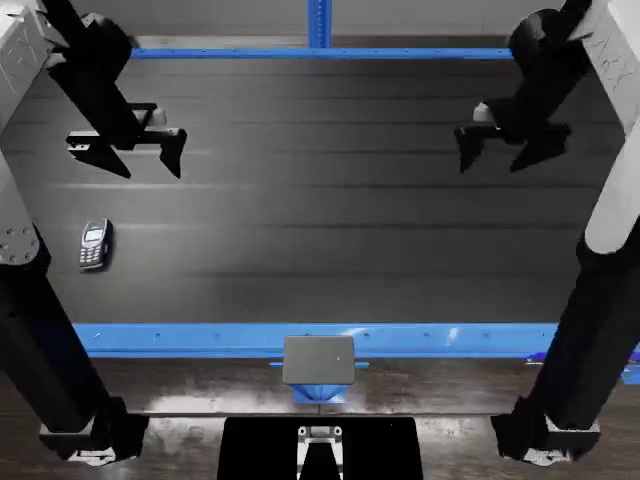}\par{\scriptsize Video 8.20\,s}\end{minipage}\hfill
\begin{minipage}[b]{0.319\linewidth}\centering\includegraphics[width=\linewidth,height=112pt,keepaspectratio]{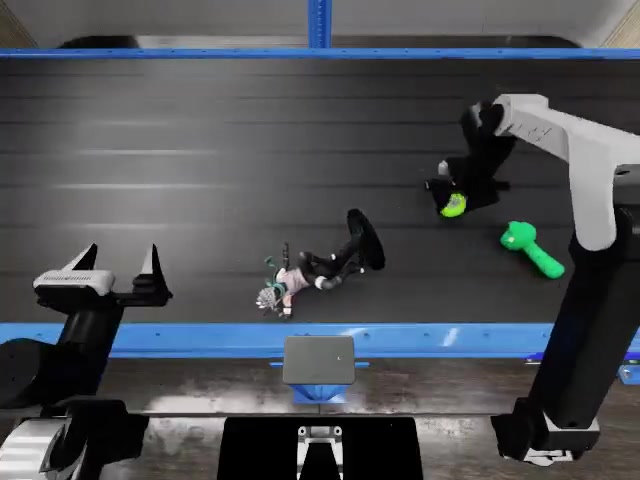}\par{\scriptsize Video 19.88\,s}\end{minipage}
\caption{\textbf{Gemini conveyor matching.} Recorded head-camera frames from \texttt{task-272}. The decisive recovery is resolved by consecutive tool receipts at events 550--559, rather than inferred solely from sampled frames.}\label{fig:app-task-272}\end{figure}
\appstage{1}{The agent tracks a matching green object among distractors and attempts a grasp. The target is moving, and robot motions also advance the conveyor. The following request is the later right-arm grasp at the observed travel lane.}
\begin{appexcerpt}{task-272 | event 537 | returned step 501}\appsmall \textbf{Agent tool note:} ``Descending right arm to grasp height z = 0.7725 at (0.30, -0.02) over the green Tamagotchi and closing the gripper upon arrival.''\end{appexcerpt}
\appstage{2}{A subsequent lift to 0.95 m is rejected without increasing the recorded step beyond 501. The agent lowers the requested height to 0.85 m; the next receipt confirms progress to step 508. This is a concrete revision after negative execution feedback.}
\begin{appexcerpt}{task-272 | event 550 | returned step 501}\appsmall \textbf{Tool feedback:} ``right arm pose unreachable (planner status: Fail).'' [excerpt]\end{appexcerpt}
\begin{appexcerpt}{task-272 | event 551 | returned step 508}\appsmall \textbf{Agent tool note:} ``Lifting right arm with grasped Tamagotchi to z = 0.85, which is a verified reachable intermediate height.''\end{appexcerpt}
\appstage{3}{The final request retracts toward home while maintaining the grasp. The episode ends at step 511 with retained success. The requested motion duration and the actual terminal step differ because evaluation can stop an in-flight segment.}
\clearpage\subsection{DeepSeek: correction, interception and a terminal win}
\noindent\textbf{Record:} \nolinkurl{task-350}. \textbf{Outcome:} success. \textbf{Robot requests:} 39. \textbf{Scored steps:} 241/1000.\par
\begin{figure}[H]\centering
\begin{minipage}[b]{0.319\linewidth}\centering\includegraphics[width=\linewidth,height=112pt,keepaspectratio]{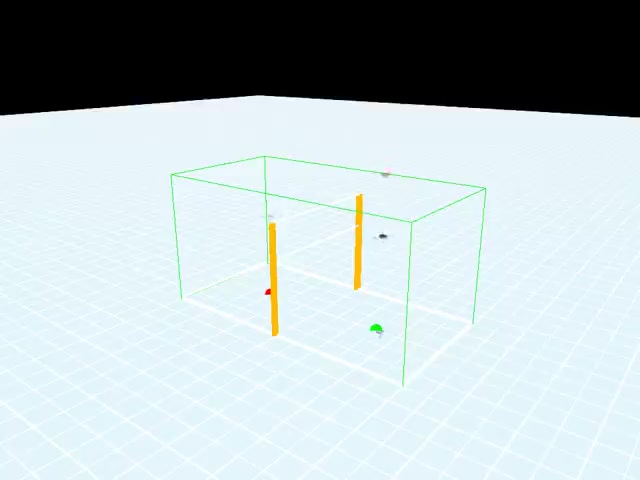}\par{\scriptsize Video 0.10\,s}\end{minipage}\hfill
\begin{minipage}[b]{0.319\linewidth}\centering\includegraphics[width=\linewidth,height=112pt,keepaspectratio]{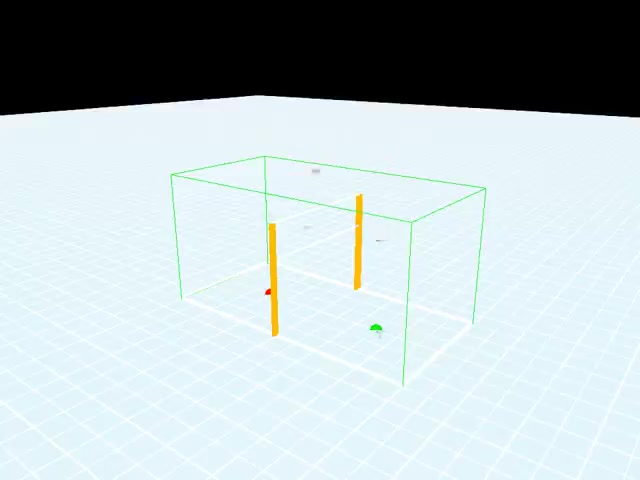}\par{\scriptsize Video 1.93\,s}\end{minipage}\hfill
\begin{minipage}[b]{0.319\linewidth}\centering\includegraphics[width=\linewidth,height=112pt,keepaspectratio]{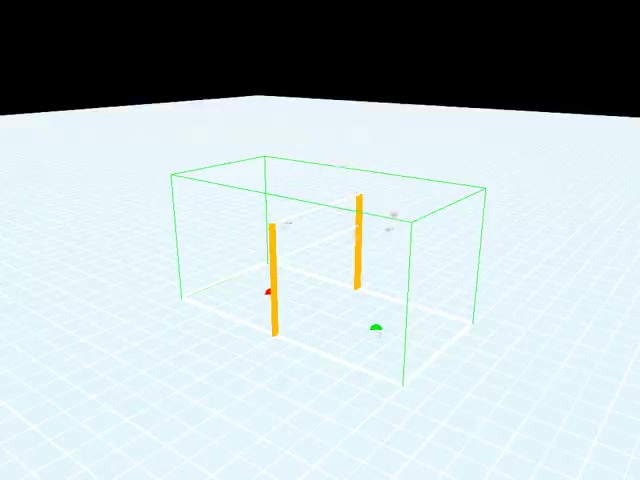}\par{\scriptsize Video 4.69\,s}\end{minipage}
\caption{\textbf{DeepSeek volleyball 1v1.} Reviewer video from \texttt{task-350}. The result is the native terminal win for the controlled actor; surviving or touching the ball alone would not suffice.}\label{fig:app-task-350}\end{figure}
\appstage{1}{After an initial collective command, the agent explicitly revises its hover calculation in response to a larger-than-expected rise. Its explanation is an observable message; it is not treated as a direct measurement.}
\begin{appexcerpt}{task-350 | event 17}\appsmall \textbf{Agent message:} ``The vehicle rose faster than expected'' [excerpt]\end{appexcerpt}
\appstage{2}{The transcript records an attempted return followed by recovery away from the net. The later message below identifies the next incoming ball and a new interception target. These messages document the intended strategy, while the state trace records actual motion.}
\begin{appexcerpt}{task-350 | event 199}\appsmall \textbf{Agent message:} ``The opponent has returned the ball --- it's now heading to my side (arriving \textasciitilde{}x=1.24, z=2.0 around step 192). Repositioning to intercept.''\end{appexcerpt}
\appstage{3}{The native result records four total contacts, two by each side, and a win for the controlled actor. The retained success occurs at step 241 of a 1,000-step horizon. This is one winning episode, not an estimate of match win probability.}

\clearpage
\section{Observable Behavioural Patterns}
\label{app:behaviour}
The cases support the following descriptive categories. Labels attach to linked local sequences of observations, agent outputs, tool requests and receipts. One episode may exhibit several categories. We do not turn this purposive case selection into a corpus-wide frequency estimate or claim inter-annotator agreement that was not measured.
\par\begingroup\small\rwtablemono\renewcommand{\arraystretch}{1.1}
\begin{longtable}{>{\raggedright\arraybackslash}p{.23\linewidth}>{\raggedright\arraybackslash}p{.32\linewidth}>{\raggedright\arraybackslash}p{.35\linewidth}}
\toprule\textbf{Pattern} & \textbf{Trace evidence} & \textbf{Interpretation boundary}\\\midrule\endhead
Sequential decomposition & Astra stacking names the placement order and reuses a stack centre after release/retraction. & An explicit plan is evidence of intent; the final checker confirms the completed arrangement.\\
Feedback-driven correction & Gemini revises a rejected 0.95 m lift to 0.85 m; the next receipt advances from step 501 to 508. & The consecutive request/receipt pair supports a local correction, not a causal effect of a general recovery policy.\\
Frequent dynamic feedback & Opus 5.5 juggling uses 274 robot requests over 800 control steps with repeated strike/descent phases. & This successful case does not establish that more calls universally improve success.\\
Calibration revision & DeepSeek changes its hover computation after reporting excessive rise. & A visible self-correction is distinct from independently verified correctness of its entire controller.\\
Contact/grasp reassessment & Kimi pouring repeatedly reports missed closure and changes approach or wrist orientation. & The agent notices difficulty; final failure should not be described as an unobserved or hallucinated success.\\
Pre-action analysis saturation & DeepSeek computer-use episode reaches a non-action boundary at step zero. & No robot motion is available for diagnosing physical control ability in that episode.\\
Post-boundary continuation & Opus 5.5 stacking has a historical success after an earlier scored interaction cutoff. & Late footage cannot be counted as success under the retained budget.\\
\bottomrule\end{longtable}\endgroup

\subsection{What counts as using feedback}
Receiving a fresh observation, explicitly viewing an image file, and changing an action after feedback are different events. The present archive directly counts tool and image-view events; deciding whether visual information changed an action requires a local evidence chain. For instance, the Gemini case links an unreachable-pose response to a smaller requested lift and subsequent confirmed execution. A low image-view count alone does not show that an agent ignored images already supplied in robot feedback.

\subsection{How to read agent claims}
Tool notes and messages record the agent's intentions or interpretation; they do not independently prove retention, contact or completion. In Kimi pouring, the notes acknowledge a tipped bottle and missed grasps, so the failure should not be described as a false claim of final success. In Astra stacking, the native result reports success although the final release request lacks terminal client feedback. We preserve that distinction between missing feedback and failed execution.

\clearpage
\section{Matched-Task Comparisons}
\label{app:controlled}
The current five-model snapshot supports matched-task observational comparisons, not controlled ablations of history, code control or reasoning budget. All five models encounter the same named 84-task inventory with matching per-task seed metadata. Differences between providers, collection times and incomplete build metadata remain possible confounders. We therefore report outcome overlap and costs for matched successful tasks without attributing them to an isolated interaction strategy.

\subsection{Success overlap}
Each off-diagonal cell below counts tasks solved by both models; diagonal cells are each model's success count. Across the five sets, 21 unique tasks are solved by at least one model. The matrix demonstrates overlap without assuming independent model errors or an executable model-selection oracle.

\begin{center}\par\begingroup\small\rwtablemono\renewcommand{\arraystretch}{1.1}\begin{tabular}{lrrrrr}
\toprule \textbf{Model} & \textbf{Astra} & \textbf{Opus 5.5} & \textbf{Kimi} & \textbf{DeepSeek} & \textbf{Gemini} \\
\midrule
Astra & 16 & 8 & 2 & 1 & 1 \\
Opus 5.5 & 8 & 13 & 1 & 1 & 1 \\
Kimi & 2 & 1 & 2 & 0 & 0 \\
DeepSeek & 1 & 1 & 0 & 1 & 0 \\
Gemini & 1 & 1 & 0 & 0 & 1 \\
\bottomrule\end{tabular}\endgroup\end{center}
\subsection{Astra and Opus 5.5 on shared successes}
Eight tasks are successful for both Astra and Opus 5.5 in this retained snapshot. Table~\ref{tab:app-shared} separates robot requests, all recorded calls, and retained result steps. These are full retained successful episodes; they are not a common intermediate-state alignment or a time-to-first-success experiment. The resource fields are audited from each task's own event stream.
\begin{table}[H]\centering\fontsize{8.6}{10.5}\selectfont\rwtablemono
\caption{Matched successful tasks. Every numeric pair is Astra / Opus 5.5.}\label{tab:app-shared}
\begin{tabularx}{\linewidth}{Yrrr}\toprule
\textbf{Task} & \textbf{Robot requests} & \textbf{All calls} & \textbf{Result steps}\\\midrule
Conveyor matching & 45 / 35 & 45 / 35 & 380 / 347 \\
Plastic clutter sorting & 41 / 104 & 46 / 107 & 784 / 1497 \\
Three fruits on a plate & 21 / 31 & 22 / 58 & 284 / 419 \\
Cube left of bowl & 17 / 18 & 24 / 31 & 286 / 239 \\
White mug centring & 18 / 22 & 20 / 42 & 280 / 341 \\
Volleyball 1v1 & 30 / 152 & 30 / 157 & 135 / 498 \\
Courtyard driving & 16 / 37 & 16 / 42 & 703 / 1006 \\
Hairpin driving & 33 / 88 & 33 / 88 & 1383 / 1829 \\
\bottomrule\end{tabularx}\end{table}
Fewer tool requests need not imply fewer physical steps: a request can hold an action for a longer segment, and auxiliary analysis changes the total-call count without advancing physics. Likewise, a shorter recorded elapsed time can reflect earlier failure. Controlled claims about history retention, segment length or coding control require paired reruns in which those variables are explicitly changed while the task, state, budget and checker are held fixed; this archive does not supply such an experiment.

\clearpage
\section{Failure Cases and Outcome Adjudication}
\label{app:diagnostics}
We distinguish failure after physical interaction, exhaustion before robot execution, and a mismatch between historical terminal results and the scored budget boundary. These examples are selected to clarify the evidence chain; they are not an exhaustive taxonomy or a measured distribution of failure causes.
\subsection{Kimi: repeated grasp revision without a completed pour}
\noindent\textbf{Record:} \nolinkurl{task-25}. \textbf{Outcome:} failure. \textbf{Robot requests:} 27. \textbf{Scored steps:} 400/400.\par
\begin{figure}[H]\centering
\begin{minipage}[b]{0.319\linewidth}\centering\includegraphics[width=\linewidth,height=112pt,keepaspectratio]{figures/appendix/task-25-0.jpg}\par{\scriptsize Video 0.10\,s}\end{minipage}\hfill
\begin{minipage}[b]{0.319\linewidth}\centering\includegraphics[width=\linewidth,height=112pt,keepaspectratio]{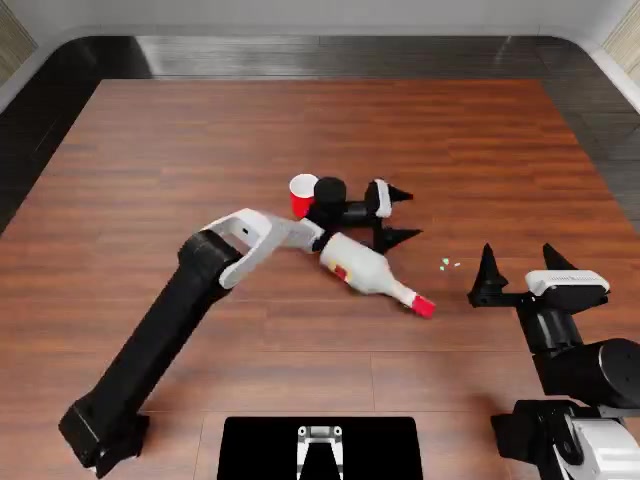}\par{\scriptsize Video 6.40\,s}\end{minipage}\hfill
\begin{minipage}[b]{0.319\linewidth}\centering\includegraphics[width=\linewidth,height=112pt,keepaspectratio]{figures/appendix/task-25-3.jpg}\par{\scriptsize Video 15.52\,s}\end{minipage}
\caption{\textbf{Kimi pouring failure.} Head-camera frames from \texttt{task-25}. The bottle is upright initially and later lies on the table. Agent notes acknowledge failed acquisition; the retained episode ends unsuccessfully at step 400.}\label{fig:app-task-25}\end{figure}
\appstage{1}{The initial approach advances 42 steps. The following tool note reports that contact knocked the bottle onto its side, then requests closure. This is evidence that the agent noticed the changed object state.}
\begin{appexcerpt}{task-25 | event 47 | returned step 48}\appsmall \textbf{Agent tool note:} ``The approach knocked the bottle onto its side'' [excerpt]\end{appexcerpt}
\appstage{2}{The next revision acknowledges that closure missed. The agent subsequently rotates the wrist, raises for clearance, shifts laterally and retries. These actions consume time without establishing a stable bottle grasp.}
\begin{appexcerpt}{task-25 | event 54 | returned step 72}\appsmall \textbf{Agent tool note:} ``The first closure missed because the gripper was above the fallen bottle.'' [excerpt]\end{appexcerpt}
\appstage{3}{Control later switches to the right arm, but the final request still repositions above the bottle. The record ends at the 400-step horizon with failure. The trace supports unsuccessful grasp acquisition and budget exhaustion; it does not support a claim that the model declared the pour complete.}
\begin{appexcerpt}{task-25 | event 327 | returned step 400}\appsmall \textbf{Agent tool note:} ``The right gripper is now beyond the bottle mouth'' [excerpt]\end{appexcerpt}
\clearpage\subsection{DeepSeek: analysis consumes the budget before any action}
\noindent\textbf{Record:} \nolinkurl{task-232}. \textbf{Outcome:} failure. \textbf{Robot requests:} 0. \textbf{Scored steps:} 0/593.\par

The Bench2Dex computer-use record contains no robot-tool request and no confirmed physical step. Its terminal non-action ledger reaches 15 consecutive auxiliary units. The visible merged trace contains ten shell executions and four explicit image views; the ledger total is 15 and flags incomplete accounting. We report both quantities rather than inventing a missing fifteenth invocation.

\appstage{1}{The agent constructs forward-kinematics scripts and inspects supplied observations. These are permitted workspace activities, but they do not advance the robot.}
\begin{appexcerpt}{task-232 | event 4}\appsmall \textbf{Shell command excerpt (Python source):}\par{\ttfamily\footnotesize\raggedright \hspace*{0ex}def T(xyz,rpy):\\\hspace*{4ex}M=np.eye(4)\\\hspace*{4ex}M[:3,:3]=rpy2R(*rpy)\\\hspace*{4ex}M[:3,3]=xyz\\\hspace*{4ex}return M\par}\end{appexcerpt}

\appstage{2}{Subsequent calls crop and inspect images of the hands, keyboard and mouse, and revise the kinematic calculation. This documents preparatory analysis, with no recorded attempt to actuate the robot.}
\begin{appexcerpt}{task-232 | visible auxiliary events | reviewer summary}\appsmall
\textbf{Shell executions:} 10. \textbf{Explicit image views:} 4.\par
\textbf{Robot requests:} 0. \textbf{Confirmed control steps:} 0.\par
\textbf{Terminal ledger:} total 15; consecutive peak 15; stop reason \texttt{nonaction\_consecutive\_limit}.
\end{appexcerpt}
\appstage{3}{The run is unsuccessful under the retained interaction protocol. Since there is no executed robot action, this case cannot establish whether a proposed joint command would have succeeded or failed physically. It identifies a failure to convert available analysis into execution before the interaction limit.}

\subsection{Attribution limits}
A malformed request, a planner rejection, a dropped host event, a simulator failure and an unmet task predicate are different mechanisms. The scored label alone does not identify which one occurred. In particular, the discrepancy between 14 visible auxiliary events and 15 charged units is an accounting limitation; it is not evidence of an unlogged physical action. The raw response, converted request, validation receipt and environment state must be traced before assigning a more specific cause.
\clearpage\subsection{Opus 5.5: late success beyond the scored boundary}
\noindent\textbf{Record:} \nolinkurl{task-82}. \textbf{Outcome:} failure. \textbf{Robot requests:} 40. \textbf{Scored steps:} 313/1350.\par
\begin{figure}[H]\centering
\begin{minipage}[b]{0.319\linewidth}\centering\includegraphics[width=\linewidth,height=112pt,keepaspectratio]{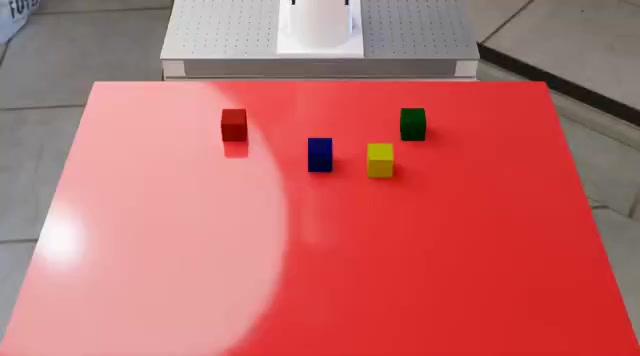}\par{\scriptsize Video 0.10\,s}\end{minipage}\hfill
\begin{minipage}[b]{0.319\linewidth}\centering\includegraphics[width=\linewidth,height=112pt,keepaspectratio]{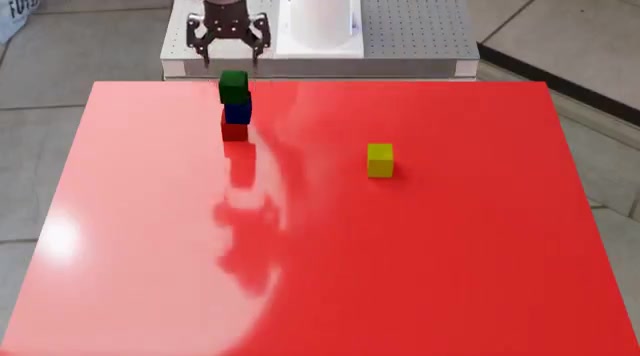}\par{\scriptsize Video 30.06\,s}\end{minipage}\hfill
\begin{minipage}[b]{0.319\linewidth}\centering\includegraphics[width=\linewidth,height=112pt,keepaspectratio]{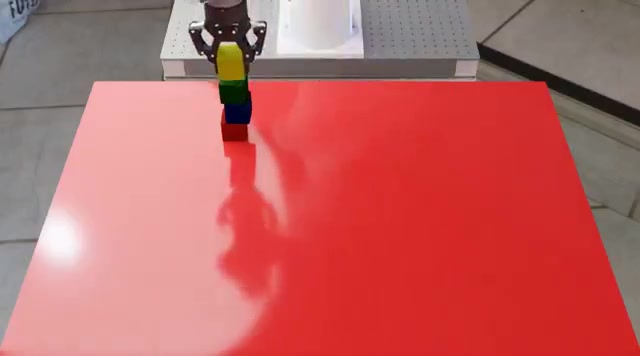}\par{\scriptsize Video 38.88\,s}\end{minipage}
\caption{\textbf{Opus 5.5 stacking: historical continuation.} Frames from the full retained video of \texttt{task-82}. The later two views lie beyond the scored step-313 budget boundary; they illustrate continued work, not valid success within that budget.}\label{fig:app-task-82}\end{figure}
\appstage{1}{The agent calibrates motion, grasps blue and releases it onto red. Its messages describe a stable partial stack. The retrospective non-action ledger reaches the total limit of 30 at control step 313.}
\begin{appexcerpt}{task-82 | event 325}\appsmall \textbf{Agent message:} ``Blue is now on red, slightly offset but stable. Next I'll locate the green block.''\end{appexcerpt}

\appstage{2}{The historical log continues: the agent grasps green, places it, and proceeds to yellow. These later events are available for diagnosing strategy, but are outside the retained scoring window.}
\begin{appexcerpt}{task-82 | event 527}\appsmall \textbf{Agent message:} ``Green is released on blue, and the side view shows red--blue--green stacked. Lifting straight up, then going for yellow, which I estimated at about (0.478, 0.109).''\end{appexcerpt}

\appstage{3}{The original run result preserves success after 602 executed control steps, while completed merged environment events reach step 601. The retained manifest instead reports failure at step 313 with reason \texttt{nonaction\_total\_limit}. Its scoring source explicitly identifies retrospective budget adjudication.}
\begin{appexcerpt}{task-82 | two outcome records, different evaluation boundaries}\appsmall
\textbf{Retained score:} failure at step 313; total non-action limit 30.\par
\textbf{Historical run:} success; 602 executed steps.\par
\textbf{Merged execution stream:} completed events through step 601.\par
\textbf{Interpretation:} the late native success does not change the scored budget-limited outcome.
\end{appexcerpt}
\clearpage\subsection{Complete adjudication ledger}
Twenty-one retained records are marked adjudicated: 15 Opus 5.5 and 6 Kimi records. Nine lack a Boolean success value in the original run record and are assigned retained failure; twelve preserve a Boolean historical outcome. Of the latter, three historical successes become failures at an earlier budget boundary. The ledger below records all affected tasks, including adjudications that leave a historical failure unchanged. A missing original Boolean is shown as ``unavailable'', not silently converted to a native failure.
\par\begingroup\fontsize{8.2}{10}\selectfont\rwtablemono
\begin{longtable}{lp{.37\linewidth}lrr}
\toprule\textbf{Trace} & \textbf{Model / task} & \textbf{Original} & \textbf{Scored step} & \textbf{Observed step}\\\midrule\endhead
\nolinkurl{task-1} & Kimi K3 / \nolinkurl{CountertopCleanup} & failure & 489 & 600 \\
\nolinkurl{task-104} & Opus 5.5 / \nolinkurl{visual} & failure & 19 & 143 \\
\nolinkurl{task-13} & Kimi K3 / \nolinkurl{OrganizeMugsByHandle} & unavailable & 335 & 335 \\
\nolinkurl{task-17} & Kimi K3 / \nolinkurl{MicrowaveCorrectMeal} & unavailable & 372 & 372 \\
\nolinkurl{task-19} & Kimi K3 / \nolinkurl{ResetCabinetDoors} & unavailable & 2420 & 2420 \\
\nolinkurl{task-2} & Opus 5.5 / \nolinkurl{SortingCleanup} & unavailable & 778 & 778 \\
\nolinkurl{task-3} & Kimi K3 / \nolinkurl{SortingCleanup} & unavailable & 306 & 306 \\
\nolinkurl{task-5} & Kimi K3 / \nolinkurl{CoffeeSetupMug} & failure & 192 & 600 \\
\nolinkurl{task-50} & Opus 5.5 / \nolinkurl{carrying_in_groceries} & failure & 265 & 1112 \\
\nolinkurl{task-52} & Opus 5.5 / \nolinkurl{clean_up_your_desk} & failure & 290 & 2001 \\
\nolinkurl{task-54} & Opus 5.5 / \nolinkurl{slicing_vegetables} & failure & 240 & 2001 \\
\nolinkurl{task-58} & Opus 5.5 / \nolinkurl{clean_boxing_gloves} & failure & 220 & 2001 \\
\nolinkurl{task-60} & Opus 5.5 / \nolinkurl{putting_up_Christmas_decorations_inside} & unavailable & 2000 & 2000 \\
\nolinkurl{task-62} & Opus 5.5 / \nolinkurl{setting_the_table} & unavailable & 1114 & 1114 \\
\nolinkurl{task-64} & Opus 5.5 / \nolinkurl{putting_dishes_away_after_cleaning} & failure & 787 & 1132 \\
\nolinkurl{task-70} & Opus 5.5 / \nolinkurl{ToolOrganizationTask} & unavailable & 1025 & 1025 \\
\nolinkurl{task-72} & Opus 5.5 / \nolinkurl{NonHammerToolsInRightBinTask} & unavailable & 1392 & 1392 \\
\nolinkurl{task-74} & Opus 5.5 / \nolinkurl{FoodPacking2CansTask} & success & 425 & 531 \\
\nolinkurl{task-80} & Opus 5.5 / \nolinkurl{PutTwoMugsOnShelfTask} & success & 174 & 466 \\
\nolinkurl{task-82} & Opus 5.5 / \nolinkurl{BlockStackingSpecifiedOrderTask} & success & 313 & 601 \\
\nolinkurl{task-98} & Opus 5.5 / \nolinkurl{mushr-drift} & failure & 120 & 205 \\
\bottomrule\end{longtable}\endgroup
All retained outcomes in this ledger are failures. The supplementary inventory preserves the page hashes and both outcome fields. Some entries have incomplete invocation accounting or lack terminal client feedback; those limitations remain attached to the records. The current local task implementation is not substituted for the historical prompt or checker contract when interpreting these runs.

\end{document}